\documentclass[smallcondensed]{svjour3}
\title{Supersparse Linear Integer Models for Optimized Medical Scoring Systems}
\author{Berk Ustun \and 
        Cynthia Rudin}
\institute{Berk Ustun \at
           Department of Electrical Engineering and Computer Science \\ 
           Massachusetts Institute of Technology \\ 
           \email{ustunb@mit.edu} 
           \and
           Cynthia Rudin \at
           Sloan School of Management and CSAIL \\
            Massachusetts Institute of Technology \\ 
           \email{rudin@mit.edu} 
}
\date{\vspace{-3.5em}}
\usepackage{amsfonts,bbm,bm,courier}
\usepackage{xcolor}
\usepackage{hyperref}
\hypersetup{%
  colorlinks=true,
  urlcolor=blue,%
  linkcolor=blue,%
  citecolor=blue,%
  linkbordercolor=blue,
  pdfborderstyle={/S/U/W 1}
}

\usepackage{etex}
\usepackage{relsize}
\usepackage{rotating}

\usepackage[authoryear, comma,round]{natbib}

\usepackage{amsmath,amssymb}
\usepackage{mathtools}
\usepackage{eqnarray}

\usepackage{array}
\usepackage{tabularx}
\usepackage{multirow}
\usepackage{hhline}
\usepackage{booktabs}
\newcommand{\cell}[2]{\setlength{\tabcolsep}{0pt}\begin{tabular}{#1}#2 \end{tabular}}
\newcommand{\bfcell}[2]{\setlength{\tabcolsep}{0pt}\textbf{\begin{tabular}{#1}#2\end{tabular}}}
\newcommand{\ddcell}[1]{\setlength{\tabcolsep}{0pt}\begin{tabular}{l>{\hspace{4pt}}l}#1\end{tabular}}

\usepackage{comment}
\usepackage{verbatim}
\usepackage{fancyvrb}
\usepackage{colortbl}
\usepackage{placeins}

\usepackage{enumitem} 

\usepackage{algorithm}
\usepackage{algorithmic}

\usepackage{graphicx}
\usepackage{epstopdf} 
\usepackage[font=small,labelfont=bf,width=0.9\textwidth]{caption}
\usepackage{tikz,pgfplots}
\usetikzlibrary{shapes,arrows}
\usetikzlibrary{matrix}
\pgfplotsset{compat=newest}
\pgfplotsset{plot coordinates/math parser=false}
\newlength\fheight
\newlength\fwidth
\newcommand{\tabdataname}[1]{\begin{tabular}{c}\texttt{#1}\end{tabular}}

\newcommand{\shortsection}[1]{\vspace{-0.5em}\subsubsection*{#1}\vspace{-0.5em}}

\newtheorem{thm}{Theorem}

\newcommand{\textds}[1]{{\footnotesize\texttt{#1}}}

\DeclareMathOperator*{\argmin}{argmin}
\newcommand{\lzero}{\ell_0}
\newcommand{\lone}{\ell_1}
\newcommand{\ltwo}{\ell_2}

\newcommand{\vnorm}[1]{\left\|#1\right\|}
\newcommand{\indic}[1]{\mathbbm{1}\left[#1\right]}
\newcommand{\ZeroOneLoss}[1]{\frac{1}{N}\sum_{i=1}^N \indic{y_i #1^T \xb_i \leq 0}}

\newcommand{\for}{\textnormal{ for }}
\newcommand{\sign}[1]{\textnormal{sign}\left(#1\right)}
\newcommand{\lambdab}{\bm{\lambda}}
\newcommand{\xb}{\bm{x}}
\newcommand{\rhob}{\bm{\rho}}

\newcommand{\iplus}{\mathcal{I}^{+}}
\newcommand{\iminus}{\mathcal{I}^{-}}
\newcommand{\wplus}{W^{+}} 
\newcommand{\wminus}{W^{-}}
\newcommand{\nplus}{N^{+}}
\newcommand{\nminus}{N^{-}}

\newcommand{\data}{\mathcal{D}}
\newcommand{\Lset}{\mathcal{L}}
\newcommand{\X}{\mathcal{X}}
\newcommand{\Y}{\mathcal{Y}}
\newcommand{\F}{\mathcal{F}}
\newcommand{\I}{\mathcal{I}}

\newcommand{\R}{\mathbb{R}}

\newcommand{\Z}{\mathbb{Z}}
\newcommand{\B}{\{0,1\}}

\newcommand{\IntPen}{\Phi}
\newcommand{\loss}{\psi}
\newcommand{\Loss}[1]{\textnormal{Loss}\left(#1\right)}

\newcommand{\st}{\textnormal{s.t.}}
\newcommand{\mprange}[3]{{#1}={#2}\textnormal{,...,}{#3}}
\newcommand{\mpdes}[1]{\textit{\scriptsize #1}}

\newcommand{\pkg}[1]{{\fontseries{b}\selectfont #1}} 

\begin{document}
\maketitle
\abstract{Scoring systems are linear classification models that only require users to add, subtract and multiply a few small numbers in order to make a prediction. These models are in widespread use by the medical community, but are difficult to learn from data because they need to be accurate and sparse, have coprime integer coefficients, and satisfy multiple operational constraints. We present a new method for creating data-driven scoring systems called a Supersparse Linear Integer Model (SLIM). SLIM scoring systems are built by solving an integer program that directly encodes measures of accuracy (the 0--1 loss) and sparsity (the $\ell_0$-seminorm) while restricting coefficients to coprime integers. SLIM can seamlessly incorporate a wide range of operational constraints related to accuracy and sparsity, and can produce highly tailored models without parameter tuning. We provide bounds on the testing and training accuracy of SLIM scoring systems, and present a new data reduction technique that can improve scalability by eliminating a portion of the training data beforehand. Our paper includes results from a collaboration with the Massachusetts General Hospital Sleep Laboratory, where SLIM was used to create a highly tailored scoring system for sleep apnea screening.} \vspace{-1.5em}
\section{Introduction}\label{Sec::Introduction}\vspace{-0.5em}
\textit{Scoring systems} are linear classification models that only require users to add, subtract and multiply a few small numbers in order to make a prediction. These models are used to assess the risk of numerous serious medical conditions since they allow physicians to make quick predictions, without extensive training, and without the use of a computer \citep[see e.g.,][]{knaus1991apache,bone1992american,moreno2005saps}. Many medical scoring systems that are currently in use were hand-crafted by physicians, whereby a panel of experts simply agreed on a model \citep[see e.g., the CHADS$_2$ score of][]{gage2001validation}. Some medical scoring systems are data-driven in the sense that they were created by rounding logistic regression coefficients \citep[see e.g., the SAPS II score of][]{le1993new}. Despite the widespread use of medical scoring systems in high-stakes applications, there has been little to no work that has focused on a direct method to learn these models from data.

Scoring systems are difficult to create using traditional machine learning methods because they need to be accurate, sparse, and use small coprime integer coefficients. This task is exceptionally challenging in medical applications because models also need to satisfy explicit constraints on operational quantities such as the false positive rate or the number of features before they can be deployed. The sum of these requirements represent serious challenges for machine learning. Current methods for sparse linear classification such as the Lasso \citep{tibshirani1996regression} and Elastic Net \citep{zou2005regularization} control the accuracy and sparsity of models via convex surrogate functions to speed up computation, and require rounding to yield models with coprime integer coefficients. Approximations such as convex surrogate loss functions, $\lone$-regularization, and rounding not only degrade predictive performance but make it difficult to address operational constraints imposed by physicians. To train a model that satisfies a hard constraint on the false positive rate, for instance, we must compute its value explicitly, which is impossible when we control accuracy by means of a surrogate loss function. In practice, traditional methods are therefore only able to address operational constraints through a parameter tuning process that involves high-dimensional grid search. As we show, this approach often fails to produce a model that satisfies operational constraints, let alone a model that is optimized for predictive accuracy.

In this paper, we present a new method to create data-driven scoring systems called a Supersparse Linear Integer Model (SLIM). SLIM is a integer programming problem that optimizes direct measures of accuracy (the 0--1 loss) and sparsity (the $\ell_0$-seminorm) while restricting coefficients to a small set of coprime integers. In comparison to current methods for sparse linear classification, SLIM can produce scoring systems that are fully optimized for accuracy and sparsity, and that satisfy a wide range of complicated operational constraints without any parameter tuning. 

The main contributions of our paper are as follows.
\begin{itemize}[leftmargin=0.35cm,topsep=1pt,parsep=1pt,label=$\bullet$]

\item We present a principled machine learning approach to learn scoring systems from data. This approach can produce tailored scoring systems that satisfy multiple operational constraints without any parameter tuning. Further, it has a unique advantage for imbalanced classification problems, where constraints on class-based accuracy can be explicitly enforced. 

\item We derive new bounds on the accuracy of discrete linear classification models. In particular, we present discretization bounds that guarantee that we will not lose training accuracy when the size of the coefficient set is sufficiently large. In addition, we present generalization bounds that relate the size of the coefficient set to a uniform guarantee on testing accuracy. 

\item We develop a novel data reduction technique that can improve the scalability of supervised classification algorithms by removing a portion of the training data beforehand. Further, we show how data reduction can be applied directly to SLIM. 

\item We present results from a collaboration with the Massachusetts General Hospital (MGH) Sleep Laboratory where SLIM was used to create a highly tailored scoring system for sleep apnea screening. Screening for sleep apnea is important: the condition is difficult to diagnose, has significant costs, and affects over 12 million people in the United States alone \citep{kapur2010obstructive}.
	
\item We provide a detailed experimental comparison between SLIM and eight popular classification methods on publicly available datasets. Our results suggest that SLIM can produce scoring systems that are accurate and sparse in a matter of minutes.

\item We provide software to create \href{https://github.com/ustunb/slim_for_matlab}{SLIM scoring systems using MATLAB and the CPLEX API}.

\end{itemize}

The remainder of our paper is structured as follows. 
In the rest of Section \ref{Sec::Introduction}, we discuss related work.
In Section \ref{Sec::Methodology}, we introduce SLIM and discuss its special properties.
In Section \ref{Sec::OperationalConstraints}, we explain how SLIM can easily enforce operational constraints that are important for medical scoring systems to be used in practice.
In Section \ref{Sec::Theory}, we present theoretical bounds on the accuracy of SLIM scoring systems and other discrete linear classification models.
In Section \ref{Sec::Reduction}, we present a data reduction technique to decrease the computation associated with SLIM and other supervised classification methods.
In Section \ref{Sec::SleepApnea}, we discuss a collaboration with the MGH Sleep Laboratory where we used SLIM to create a highly tailored scoring system for sleep apnea screening.
In Section \ref{Sec::NumericalExperiments}, we present experimental results to show that SLIM can create high-quality scoring systems in minutes.
In Section \ref{Sec::Extensions}, we present specialized extensions of SLIM.
\subsection{Related Work}
In what follows, we briefly discuss related work in medical scoring systems and linear classification. 
\shortsection{Medical Scoring Systems}
Medical scoring systems are sparse linear models with small coprime coefficients. Some popular examples include: SAPS I, II and III \citep{le1993new,moreno2005saps} and APACHE I, II and III to assess ICU mortality risk \citep{knaus1981apache,knaus1985apache,knaus1991apache}; CHADS$_2$ to assess the risk of stroke in patients with atrial fibrillation \citep{gage2001validation}; and TIMI, to assess the risk of death and ischemic events \citep{antman2000timi}.%
%
Most of the scoring systems that are in widespread use today were built without optimizing for predictive accuracy. In some cases, physicians built scoring systems by combining existing methods and heuristics. The SAPS II score, for instance, was built by rounding logistic regression coefficients as \citet{le1993new} write, ``\textit{the general rule was to multiply the $\beta$ for each range by 10 and round off to the nearest integer.}" This approach is at odds with the fact that rounding is known to produce suboptimal solutions in the field of integer programming. In other cases, scoring systems were hand-crafted by a panel of physicians, and not learned from data at all. This was the case for CHADS$_2$ as explained by \citet{gage2001validation}: ``\textit{We calculated CHADS$_2$, by adding 1 point each for each of the following -- recent CHF, hypertension, age 75 years or older, and DM -- and 2 points for a history of stroke or TIA.}" Methods that can learn tailored predictive models from data, such as SLIM, should eliminate the need for physicians to build scoring systems by hand.

To date, SLIM has already been used to create medical scoring systems for the purposes of diagnosing cognitive impairments using features derived from a clock-drawing test \citep[see][]{william2015med}, and for screening sleep apnea from electronic health records \citep[see][]{ustun2015apnea}.
%
%
%
%
\shortsection{Sparse Linear Classification Models}
\noindent In comparison to SLIM, current methods for sparse linear classification are designed to fit models with real coefficients, and need to be paired with a rounding procedure to create the same kinds of scoring systems used by physicians. In practice, rounding the coefficients of a linear model may significantly alter its accuracy and sparsity, and may result in a scoring system that violates operational constraints on these quantities. Current methods are also ill-suited to create scoring systems because they control accuracy and sparsity by means of convex surrogate functions to preserve scalability (see e.g., \cite{tibshirani1996regression,efron2004least}). As we show in Sections  \ref{Sec::SleepApnea} and \ref{Sec::NumericalExperiments}, surrogate functions result in a poor trade-off between accuracy and sparsity. Convex surrogate loss functions, for instance, produce models that are not robust to outliers \citep{nguyen2013algorithms}. Similarly, $\ell_1$-regularization is only guaranteed to recover the correct sparse solution (i.e., the one that minimizes the $\ell_0$-norm) under restrictive conditions that are rarely satisfied in practice \citep{zhao2007model}. In fact, $\ell_1$-regularization may recover a solution with significantly less predictive accuracy relative to the correct sparse solution (see \cite{lin2008defense} for a discussion). 

SLIM is also related to a recent body of work on methods for discrete linear classification. Specifically, \citet{chevaleyre2013rounding} consider training linear classifiers with binary coefficients by rounding the coefficients of linear classifiers. In addition, \citet{carrizosaDILSVM13} consider training linear classifiers with small integer coefficients using a MIP formulation. SLIM can reproduce both of these models. The converse, however, is not true because the methods of \citet{chevaleyre2013rounding} and \citet{carrizosaDILSVM13}: (i) optimize the hinge loss as opposed to the 0--1 loss; and (ii) do not include a mechanism to control sparsity. These differences may result in better scalability compared to SLIM. However, they also prevent these methods to create scoring systems that are sparse, that satisfy operational constraints on accuracy and/or sparsity, and that can be trained without parameter tuning. In addition to these differences, we note that the discretization bounds and generalization bounds in Section \ref{Sec::Theory} are a novel contribution to this body of work and applicable to all linear models with discrete coefficients. 

%
%
%
\clearpage
\section{Methodology}\label{Sec::Methodology}
We start with a dataset of $N$ i.i.d. training examples $\data_N = \{(\xb_i,y_i)\}_{i=1}^N$ where $\xb_i \in \X \subseteq \R^{P+1}$ denotes a vector of features $[1, x_{i,1},\ldots,x_{i,P}]^T$ and $y_i \in \Y = \{-1,1\}$ denotes a class label. We consider linear models of the form $\hat{y}=\text{sign}(\lambdab^T\xb)$, where $\lambdab = [\lambda_0, \lambda_1,\ldots,\lambda_P]^T$ represents a vector of coefficients and $\lambda_0$ represents an intercept term. We learn the coefficients by solving an optimization problem of the form:
\begin{align}
\begin{split}\label{Eq::InterpretabilityFrameworkLinear}
\min_{\lambdab} & \qquad \Loss{\lambdab;\data_N} + C \cdot \IntPen(\lambdab) \\ 
\st & \qquad \lambdab \in \Lset.
\end{split}
\end{align}
Here: %
the \textit{loss function} $\Loss{\lambdab;\data_N}: \R^{P+1} \times (\X\times\Y)^N \to \R$  penalizes misclassifications; %
the \textit{coefficient penalty} $\IntPen(\lambdab): \R^{P+1} \to \R$ induces soft qualities that are desirable but may be sacrificed for greater accuracy; %
the \textit{coefficient set} $\Lset$ encodes hard qualities must be satisfied; %
and the \textit{trade-off parameter} $C$ controls the balance between accuracy and soft qualities. 
We assume:
(i) the coefficient set contains the null vector, $\bf{0} \in \Lset$;
(ii) the penalty is additively separable, $\IntPen(\lambdab) = \sum_{j=0}^P\IntPen_j(\lambda_j)$;
(iii) the intercept is never penalized, $\IntPen_0(\lambda_0) = 0$.

A \textit{Supersparse Linear Integer Model} (SLIM) is a special case of the optimization in \eqref{Eq::InterpretabilityFrameworkLinear}:
\begin{align}
\label{Eq::SLIMFormulation}
\begin{split}
\min_{\lambdab} & \qquad \ZeroOneLoss{\lambdab} + C_0 \vnorm{\lambdab}_0 + \epsilon \vnorm{\lambdab}_1 \\ 
\st &  \qquad \lambdab \in \Lset. 
\end{split}
\end{align}
SLIM directly optimizes accuracy and sparsity by minimizing the 0--1 loss $\ZeroOneLoss{\lambdab}$ and $\lzero$-norm $\vnorm{\lambdab}_0:=\sum_{j=1}^P{\indic{\lambda_j\neq0}}$ respectively. The constraints usually restrict coefficients to a finite set of discrete values such as $\Lset =  \{-10,\ldots,10\}^{P+1}$, and may include additional operational constraints such as $\vnorm{\lambdab}_0 \leq 10$. SLIM includes a \textit{tiny} $\lone$-penalty $\epsilon \vnorm{\lambdab}_1$ in the objective for the sole purpose of restricting coefficients to coprime values.%
\footnote{To illustrate the use of the $\lone$-penalty, consider a classifier such as $\hat{y}=\sign{x_1 + x_2}$. If the objective in \eqref{Eq::SLIMFormulation} only minimized the 0--1 loss and an $\lzero$-penalty, then $\hat{y}=\sign{2 x_1 + 2 x_2}$ would have the same objective value as $\hat{y}=\sign{x_1 + x_2}$ because it makes the same predictions and has the same number of non-zero coefficients. Since coefficients are restricted to a finite discrete set, we add a \textit{tiny} $\lone$-penalty in the objective of \eqref{Eq::SLIMFormulation} so that SLIM chooses the classifier with the smallest (i.e. coprime) coefficients, $\hat{y} = \sign{x_1+x_2}$.}
To be clear, the $\lone$-penalty parameter $\epsilon$ is always set to a value that is small enough to avoid $\lone$-regularization (that is, $\epsilon$ is small enough to guarantee that SLIM never sacrifices accuracy or sparsity to attain a smaller $\lone$-penalty).

SLIM is designed to produce scoring systems that attain a pareto-optimal trade-off between accuracy and sparsity: when we minimize 0--1 loss and the $\ell_0$-penalty, we only sacrifice classification accuracy to attain higher sparsity, and vice versa. Minimizing the 0--1 loss produces scoring systems that are completely robust to outliers and attain the best learning-theoretic guarantee on predictive accuracy \citep[see e.g.][]{brooks2011support,nguyen2013algorithms}. Similarly, controlling for sparsity via $\lzero$-regularization prevents the additional loss in accuracy due to $\ell_1$-regularization \citep[see][for a discussion]{lin2008defense}. In addition to these performance benefits, minimizing an approximation-free object function over a finite set of discrete coefficients means that the free parameters in SLIM's object have special properties.
\begin{remark}\label{Rem::CoprimeGuarantee}
If $\epsilon <\frac{\min{(1/N,C_0})}{\max_{\lambdab\in\Lset}\vnorm{\lambdab}_1}$ and $\Lset$ is a finite subset of $\Z^{P+1}$ then the optimization of \eqref{Eq::SLIMFormulation} will produce a scoring system with coprime coefficients without affecting accuracy or sparsity:\vspace{-1em} 
\begin{align*}
\argmin_{\lambdab\in\Lset} \ZeroOneLoss{\lambdab} + C_0 \vnorm{\lambdab}_0 + \epsilon \vnorm{\lambdab}_1 \subseteq \argmin_{\lambdab\in\Lset} \ZeroOneLoss{\lambdab} + C_0 \vnorm{\lambdab}_0 \\ 
\text{and \textnormal{gcd}}(\{\lambda_j^*\}_{j=0}^P)=1 \text{ for all } \lambdab^* \in \argmin_{\lambdab\in\Lset} \ZeroOneLoss{\lambdab} + C_0 \vnorm{\lambdab}_0 + \epsilon \vnorm{\lambdab}_1.
\end{align*}
\end{remark}
\begin{remark}\label{Rem::TradeOffParameterMeaning}
The trade-off parameter $C_0$ represents the maximum accuracy that SLIM will sacrifice to remove a feature from the optimal scoring system.
\end{remark}
\begin{remark}\label{Rem::TradeOffParamaterMin}
If $C_0 < \frac{1}{NP}$ and $\epsilon < \frac{\min{(1/N,C_0})}{\max_{\lambdab\in\Lset}\vnorm{\lambdab}_1} = \frac{C_0}{\max_{\lambdab\in\Lset}\vnorm{\lambdab}_1} $ then the optimization of \eqref{Eq::SLIMFormulation} will produce a scoring system with coefficients $\lambdab \in \Lset$ with the highest possible training accuracy: $$\argmin_{\lambdab\in\Lset} \ZeroOneLoss{\lambdab} + C_0 \vnorm{\lambdab}_0 + \epsilon \vnorm{\lambdab}_1 \subseteq \argmin_{\lambdab\in\Lset} \ZeroOneLoss{\lambdab}.$$
\end{remark}
\begin{remark}\label{Rem::TradeOffParamaterMax}
If $C_0 > 1 - \frac{1}{N}$ and $\epsilon <\frac{\min{(1/N,C_0)}}{\max_{\lambdab\in\Lset}\vnorm{\lambdab}_1} = \frac{1/N}{\max_{\lambdab\in\Lset}\vnorm{\lambdab}_1} $ then the optimization of \eqref{Eq::SLIMFormulation} will produce a scoring system with coefficients $\lambdab \in \Lset$ with the highest possible sparsity: $$\argmin_{\lambdab\in\Lset} \ZeroOneLoss{\lambdab} + C_0 \vnorm{\lambdab}_0 + \epsilon \vnorm{\lambdab}_1 \subseteq \argmin_{\lambdab\in\Lset} C_0 \vnorm{\lambdab}_0.$$
\end{remark}
Note that these properties are only possible using the formulation in \eqref{Eq::SLIMFormulation}. In particular, Remarks \ref{Rem::TradeOffParameterMeaning}-\ref{Rem::TradeOffParamaterMax} require that we control accuracy using the 0--1 loss and control sparsity using an $\ell_0$-penalty, and Remark \ref{Rem::CoprimeGuarantee} requires that we restrict coefficients to a finite discrete set.
\subsection{SLIM IP Formulation}\label{Sec::SLIMIPFormulation}
We train SLIM scoring systems using the following IP formulation:\vspace{-1em}
\begin{subequations}
\begin{equationarray}{crcl>{\qquad}l>{\qquad}r}
\min_{\lambdab,\bf{\psi},\bf{\Phi},\bf{\alpha},\bf{\beta}} &\frac{1}{N}\sum_{i=1}^{N} \loss_i & + & \sum_{j=1}^{P} \IntPen_j  \notag \\
\st          & M_i \loss_i                  & \geq & \gamma -\sum_{j=0}^P y_i \lambda_j x_{i,j}       &\mprange{i}{1}{N} & \mpdes{0--1 loss} \label{Con::SLIMLoss} \\
& \IntPen_j & = & C_0\alpha_j + \epsilon\beta_j &\mprange{j}{1}{P}& \mpdes{int. penalty} \label{Con::SLIMIntPenalty} \\
& -\Lambda_j\alpha_j  & \leq & \lambda_j \leq \Lambda_j\alpha_j  &\mprange{j}{1}{P} & \mpdes{$\lzero$-norm} \label{Con::SLIML0Norm} \\
& -\beta_j &  \leq & \lambda_j  \leq \beta_j &\mprange{j}{1}{P} & \mpdes{$\lone$-norm} \label{Con::SLIML1Norm} \\
& \lambda_j & \in & \Lset_j &  \mprange{j}{0}{P} & \mpdes{coefficient set} \notag \\ 
& \loss_i & \in & \B &  \mprange{i}{1}{N} & \mpdes{loss variables} \notag  \\
& \IntPen_j  & \in & \R_+  & \mprange{j}{1}{P} & \mpdes{penalty variables} \notag \\
& \alpha_j  & \in & \B  & \mprange{j}{1}{P} & \mpdes{$\lzero$ variables} \notag \\
& \beta_j    & \in & \R_+ & \mprange{j}{1}{P} & \mpdes{$\lone$ variables} \notag
\end{equationarray}
\end{subequations}
Here, the constraints in \eqref{Con::SLIMLoss} set the loss variables $\loss_i = \indic{y_i \lambdab^T\xb_i \leq 0}$ to $1$ if a linear classifier with coefficients $\lambdab$ misclassifies example $i$. This is a Big-M constraint for the 0--1 loss that depends on scalar parameters $\gamma$ and $M_i$ \cite[see e.g.,][]{rubin2009mixed}. The value of $M_i$ represents the maximum score when example $i$ is misclassified, and can be set as $M_i = \max_{\lambdab \in \Lset} (\gamma - y_i\lambdab^T\xb_i)$ which is easy to compute since $\Lset$ is finite. The value of $\gamma$ represents the ``margin" and should be set as a lower bound on $y_i\lambdab^T\xb_i$. When the features are binary, $\gamma$ can be set to any value between 0 and 1. In other cases, the lower bound is difficult to calculate exactly so we set $\gamma=0.1$, which makes an implicit assumption on the values of the features. The constraints in \eqref{Con::SLIMIntPenalty} set the total penalty for each coefficient to $\IntPen_j = C_0 \alpha_j + \epsilon \beta_j$, where $\alpha_j := \indic{\lambda_j\neq 0}$ is defined by Big-M constraints in \eqref{Con::SLIML0Norm}, and $\beta_j := |\lambda_j|$ is defined by the constraints in \eqref{Con::SLIML1Norm}. We denote the largest absolute value of each coefficient as $\Lambda_j := \max_{\lambda_j\in\Lset_j} |\lambda_j|$.

Restricting coefficients to a finite set results in significant practical benefits for the SLIM IP formulation, especially in comparison to other IP formulations that minimize the 0--1-loss and/or penalize the $\ell_0$-norm. Many IP formulations compute the 0--1 loss and $\ell_0$-norm by means of Big-M constraints that use require users to specify Big-M constants \citep[see e.g.,][]{wolsey1998integer}. Restricting the coefficients to a finite set allows us to bound Big-M constants in the SLIM IP formulation. Specifically, the Big-M constant for computing the 0--1 loss in constraints \eqref{Con::SLIMLoss} is bounded as $M_i \leq \max_{\lambdab \in \Lset} (\gamma - y_i\lambdab^T\xb_i)$ and the Big-M constant used to compute the $\ell_0$-norm in constraints \eqref{Con::SLIML0Norm} is bounded as $\Lambda_j \leq \max_{\lambda_j\in\Lset_j} |\lambda_j|$ (compare with \cite{brooks2011support,guan2009mixed} where the same parameters have to be approximated by a ``sufficiently large" constants). Bounding these constants lead to a tighter LP relaxation, which narrows the integrality gap, and improves the ability of commercial IP solvers to quickly obtain a proof of optimality.
\subsection{Operational Constraints}\label{Sec::OperationalConstraints}
SLIM provides users with an unprecedented amount of flexibility over their models by allowing them to directly encode a wide range of operational constraints into its IP formulation. In what follows, we provide a few examples to illustrate this process. We note that these techniques are possible because: (i) the variables used to encode the 0--1 loss and $\lzero$-penalty in the SLIM IP formulation can also encode operational constraints related to accuracy and sparsity; (i) the free parameters in the SLIM objective can be set without tuning (see Remarks \ref{Rem::TradeOffParameterMeaning}--\ref{Rem::TradeOffParamaterMax}).
\shortsection{Loss Constraints for Imbalanced Data}\label{Sec::HandlingImbalancedData}
The majority of classification problems in the medical domain are imbalanced. Handling imbalanced data is incredibly difficult for most classification methods since maximizing classification accuracy often produces a trivial model (i.e., if the probability of heart attack is 1\%, a model that never predicts a heart attack is still 99\% accurate). SLIM has a unique advantage on such problems as it not only avoid producing a trivial model, but can produce a model at any user-specified point on the ROC curve without parameter tuning. That is, when physicians specify hard constraints on sensitivity (or specificity), we can encode these as \textit{loss constraints} into the IP formulation, and solve a single IP to obtain the least specific (or most sensitive) model. To train the most sensitive scoring system with a maximum error of $\gamma \in [0,1]$ on negatively-labeled examples we solve an IP with the form:
\begin{align}
\min_{\lambdab} & \qquad {\frac{W}{N}}^+ \sum_{i\in\iplus} \indic{y_i \lambdab^T\xb_i \leq 0} + \frac{W}{N}^{-} \sum_{i\in\iminus} \indic{y_i \lambdab^T\xb_i \leq 0} + C_0 \vnorm{\lambdab}_0 + \epsilon\vnorm{\lambdab}_1 \notag \\ 
\st & \qquad \frac{1}{N^-} \sum_{i\in\iminus} \indic{y_i \lambdab^T\xb_i \geq 0} \leq \gamma \label{Con::MaxFPRConstraint}\\
& \qquad \lambdab\in\Lset. \notag
\end{align}
This formulation optimizes a \textit{weighted} 0--1 loss function where $\wplus$ and $\wminus$ are user-defined weights that control the accuracy on the $\nplus$ positive examples from the set $\iplus = \{i:y_i = +1\}$, and $\nminus$ negative examples from the set $\iminus = \{i:y_i = -1\}$, respectively. Assuming that $\wplus + \wminus = 1$, we set $\wplus > \frac{\nminus}{1+\nminus}$ so that SLIM weighs the accuracy on each positive example as much as all of the negative examples. In a typical setting, this would return a scoring system that classifies all positive examples correctly at the expense of misclassify all of the negative examples in order to classify an additional positive example correctly. In this case, however, the loss constraint \eqref{Con::MaxFPRConstraint} explicitly limits the error on negative examples to $\gamma$. Thus, SLIM returns a scoring system that attains the highest sensitivity among models with a maximum error of $\gamma$ on negative examples.
%
\shortsection{Feature-Based Constraints for Input Variables}\label{Sec::FeatureBasedConstraints}
SLIM provides fine-grained control over the composition of input variables in a scoring system by formulating feature-based constraints. Specifically, we can use the indicator variables that encode the $\ell_0$-norm $\alpha_j := \indic{\lambda_j \neq 0}$ to formulate many logical constraint between features such as ``either-or" conditions and ``if-then" conditions (see \citep[][]{wolsey1998integer} for an overview). This presents a practical alternative to create classification models that obey structured sparsity constraints \cite[][]{jenatton2011structured} or hierarchical constraints \citep[][]{bien2013lasso}.

The indicator variables $\alpha_j$ can be used to limit the number of input variables to at most $\Theta$ by adding the constraint, $\sum_{j = 1}^P{\alpha_j} \leq \Theta.$ More complicated feature-based constraints include ``if-then" constraints to ensure that a scoring system will only include $hypertension$ and $heart\_attack$ if it also includes $stroke$: $\alpha_{heart\_attack} + \alpha_{hypertension} \leq 2 \alpha_{stroke},$ or hierarchical constraints to ensure that an input variable in the leaves can only be used when all features above it in the hierarchy are also used: $\alpha_{leaf} \leq \alpha_{node} \textrm{ for all nodes above the leaf}.$
\subsection{Feature-Based Preferences}

Physicians often have soft preferences between different input variables. SLIM allows practitioners to encode these preferences by specifying a distinct trade-off parameter for each coefficient $C_{0,j}$.

Explicitly, when our model should use feature $j$ instead of feature $k$, we set $C_{0,k} = C_{0,j} + \delta$, where $\delta > 0$ represents the maximum additional training accuracy that we are willing to sacrifice in order to use feature $j$ instead of feature $k$. Thus, setting $C_{0,k} = C_{0,j} + 0.02$ would ensure that we would only be willing to use feature $k$ instead of feature $j$ if it yields an additional 2\% gain in training accuracy over feature $k$. 

This approach can also be used to handle problems with missing data. Consider training a model where feature $j$ contains $M<N$ missing points. Instead of dropping these points, we can impute the values of the $M$ missing examples, and adjust the trade-off parameter $C_{0,j}$ so that our model only uses feature $j$ if it yields an additional gain in accuracy of more than $M$ examples:
\begin{align*}
C_{0,j} = C_0 + \frac{M}{N} .
\end{align*}
The adjustment factor is chosen so that: if $M=0$ then $C_{0,j} = C_0$ and if $M=N$ then $C_{0,j} = 1$ and the coefficient is dropped entirely (see Remark \ref{Rem::TradeOffParamaterMax}). This ensures that features with lots of imputed values are more heavily penalized than features with fewer imputed values.
\section{Bounds on Training and Testing Accuracy}\label{Sec::Theory}
In this section, we present bounds on the training and testing accuracy of SLIM scoring systems.
\subsection{Discretization Bounds on Training Accuracy}
\label{Sec::BoundsOnTrainingAccuracy}
Our first result shows that we can always craft a finite discrete set of coefficients $\Lset$ so that the training accuracy of a linear classifier with discrete coefficients $\lambdab \in \Lset$ (e.g. SLIM) is no worse than the training accuracy of a baseline linear classifier with real-valued coefficients $\rhob \in \R^P$ (e.g. SVM).
\clearpage
\vspace{0.5em}
\begin{thm}[Minimum Margin Resolution Bound]\label{Thm::MinMarginBound}
Let $\rhob = [\rho_1,\ldots,\rho_P]^T \in \R^P$ denote the coefficients of a baseline linear classifier trained using data $\data_N = (\xb_i,y_i)_{i=1}^N$. Let $X_{\max} = \max_i \|\xb_i\|_2$ and $\gamma_{\min} = \min_i \frac{|\rhob^T\xb_i|}{\vnorm{\rhob}_2}$ denote the largest magnitude and minimum margin achieved by any training example, respectively. 

Consider training a linear classifier with coefficients $\lambdab = [\lambda_1,\ldots,\lambda_P]^T$ from the set $\Lset = \{-\Lambda,\ldots,\Lambda\}^P$. If we choose a resolution parameter $\Lambda$ such that:
\begin{align}
\Lambda &> \frac{X_{\max}\sqrt{P}}{2 \gamma_{\min}}, \label{Eq::MinMarginLambda}
\end{align}
then there exists $\lambdab\in \Lset$ such that the 0--1 loss of $\lambdab$ is less than or equal to the 0--1 loss of $\rhob$: 
\begin{align*}
\sum_{i=1}^N \indic{y_i\lambdab^T\xb_i\leq 0} \leq \sum_{i=1}^N \indic{y_i \rhob^T \xb_i \leq 0}. 
\end{align*}
\end{thm}
\proof See Appendix \ref{Appendix::Proofs}. \endproof

\noindent The proof of Theorem \ref{Thm::MinMarginBound} uses a rounding procedure to choose a resolution parameter $\Lambda$ so that the coefficient set $\Lset$ contains a classifier with discrete coefficients $\lambdab$ that attains the same the 0--1 loss as the baseline classifier with real coefficients $\rhob$. If the baseline classifier $\rhob$ is obtained by minimizing a convex surrogate loss, then the optimal SLIM classifier trained with the coefficient set from Theorem \ref{Thm::MinMarginBound} may attain a lower 0--1 loss than $\indic{y_i\rhob^T\xb_i\leq 0}$ because SLIM directly minimizes the 0--1 loss.

The next corollary yields additional bounds on the training accuracy by considering progressively larger values of the margin. These bounds can be used to relate the resolution parameter $\Lambda$ to a worst-case guarantee on training accuracy.
\begin{corollary}[$k^\text{th}$ Margin Resolution Bound]\label{Thm::MinMarginBoundCorollary}
Let $\rhob = [\rho_1,\ldots,\rho_P]^T \in \R^P$ denote the coefficients of a linear classifier trained with data $\data_N = (\xb_i,y_i)_{i=1}^N$. Let $\gamma_{(k)}$ denote the value of the $k^\text{th}$ smallest margin, $\I_{(k)}$ denote the set of training examples with $\frac{|\rhob^T\xb_i|}{\vnorm{\rhob}_2} \leq \gamma_{(k)}$, and $X_{(k)} = \max_{i \not\in \I_{(k)}} \|\xb_i\|_2$ denote the largest magnitude of any training example $\xb_i \in \data_N$ for $i \not\in \I_{(k)}$.

Consider training a linear classifier with coefficients $\lambdab = [\lambda_1,\ldots,\lambda_P]^T$ from the set $\Lset = \{-\Lambda,\ldots,\Lambda\}^P$. If we choose a resolution parameter $\Lambda$ such that:
\begin{align*}
\Lambda > \frac{X_{(k)}\sqrt{P}}{2 \gamma_{(k)}}, \hspace{8em}\\
\intertext{then there exists $\lambdab \in \Lset$ such that the 0--1 loss of $\lambdab$ and the 0--1 loss of $\rhob$ differ by at most $k-1$:}
\sum_{i=1}^N \indic{y_i\lambdab^T\xb_i\leq 0}  - \sum_{i=1}^N \indic{y_i \rhob^T \xb_i \leq 0} \leq  k - 1. 
\end{align*}
\end{corollary}
\proof The proof follows by applying Theorem \ref{Thm::MinMarginBound} to the reduced dataset $\data_N \backslash \I_{(k)}$. \endproof
We have now shown that good discretized solutions exist and can be constructed easily. This motivates that optimal discretized solutions, which by definition are better than rounded solutions, will also be good relative to the best non-discretized solution.
\subsection{Generalization Bounds on Testing Accuracy}\label{Sec::BoundsOnTestingAccuracy}
According to the principle of structural risk minimization \citep{vapnik1998statistical}, fitting a classifier from a simpler class of models may lead to an improved guarantee on predictive accuracy. Consider training a classifier $f:\X\rightarrow\Y$ with data $\data_N = (\xb_i,y_i)_{i=1}^N$, where $\xb_i \in \X \subseteq \R^P$ and $y_i \in \Y = \{-1,1\}$. In what follows, we provide uniform generalization guarantees on the predictive accuracy of all functions, $f \in \F$. These guarantees bound the true risk $R^\text{true}(f) = \mathbb{E}_{\X,\Y} \indic{f(\xb) \neq y}$ by the empirical risk $R^\text{emp}(f) = \frac{1}{N}\sum_{i=1}^N \indic{f(\xb_i) \neq y_i}$ and other quantities important to the learning process.
\begin{thm}[Occam's Razor Bound for Discrete Linear Classifiers]\label{Thm::NormalBound}
Let $\F$ denote the set of linear classifiers with coefficients $\lambdab \in \Lset$:
\begin{align*}
\F = \left\{f :\X\to\Y \;\big|\; f(\xb)=\sign{\lambdab^T\xb} \textnormal{ and } \lambdab \in \Lset \right\}.
\end{align*}
For every $\delta > 0,$ with probability at least $1-\delta$, every classifier $f\in\F$ obeys:
\begin{align*} 
R^\textnormal{true}(f) \leq R^\textnormal{emp}(f) + \sqrt{\frac{\log(|\Lset|) - \log(\delta)}{2N}}.
\end{align*}
\end{thm}
A proof of Theorem \ref{Thm::NormalBound} can be found in Section 3.4 of \citealt{bousquet2004introduction}. The result that more restrictive hypothesis spaces can lead to better generalization provides motivation for using discrete models without necessarily expecting a loss in predictive accuracy. The bound indicates that we include more coefficients in the set $\Lset$ as the amount of data $N$ increases.

In Theorem \ref{Thm::L0Bound}, we improve the generalization bound from Theorem  \ref{Thm::NormalBound} by excluding models that are provably suboptimal from the hypothesis space. Here, we exploit the fact that we can bound the number of non-zero coefficients in a SLIM scoring system based on the value of $C_0$.
\begin{thm}[Generalization of Sparse Discrete Linear Classifiers]\label{Thm::L0Bound} Let $\F$ denote the set of linear classifiers with coefficients $\lambdab$ from a finite set $\Lset$ such that:
\begin{align*}
\F &= \left\{f :\X\to\Y \;\big|\; f(\xb)=\sign{\lambdab^T\xb} \right\} \\ 
\lambdab &\in \argmin_{\lambda \in \Lset} \frac{1}{N} \sum_{i=1}^N \indic{y_i \lambdab^T \xb_i \leq 0} + C_0\vnorm{\lambdab}_0
\end{align*}
For every $\delta > 0,$ with probability at least $1-\delta$, every classifier $f\in\F$ obeys:
\begin{align*} 
R^{\textnormal{true}}(f) &\leq R^{\textnormal{emp}}(f) + \sqrt{\frac{\log(|\mathcal{H}_{P,C_0}|) - \log(\delta)}{2N}}.
\intertext{where}
\mathcal{H}_{P,C_0} &= \bigg\{\lambdab \in \Lset \;\Big|\; \vnorm{\lambdab}_0 \leq \left\lfloor \frac{1}{C_0} \right\rfloor \bigg\}.
\end{align*}
\end{thm}
\proof See Appendix \ref{Appendix::Proofs}.\endproof

\noindent This theorem relates the trade-off parameter $C_0$ in the SLIM objective to the generalization of SLIM scoring systems. It indicates that increasing the value of the $C_0$ parameter will produce a model with better generalization properties.

In Theorem \ref{Thm::CoprimeBound}, we produce a better generalization bound by exploiting the fact that SLIM scoring systems use coprime integer coefficients (see Remark \ref{Rem::CoprimeGuarantee}). In particular, we express the generalization bound from Theorem \ref{Thm::NormalBound} using the $P$-dimensional Farey points of level $\Lambda$ \citep[see][for a definition]{2012arXiv1207.0954M}.
\begin{thm}[Generalization of Discrete Linear Classifiers with Coprime Coefficients]\label{Thm::CoprimeBound}
Let $\F$ denote the set of linear classifiers with coprime integer coefficients, $\lambdab$, bounded by $\Lambda$:
\begin{align*}
\F &= \Big\{f :\X\to\Y \;\big|\; f(\xb)=\sign{\lambdab^T\xb} \textnormal{ and } \lambdab \in \Lset \Big\},\\
\Lset &= \Big\{\lambdab \in \mathbb{\hat{Z}}^P \;\big|\; |\lambda_j| \leq \Lambda \textnormal{ for } j=1,\ldots,P\Big\},\\
\hat{\Z}^P &= \Big\{\bm{z}\in\Z^P \;\big|\; \textnormal{gcd}(\bm{z}) = 1\Big\}.
\end{align*}
For every $\delta > 0,$ with probability at least $1-\delta$, every classifier $f\in\F$ obeys:
\begin{align*} 
R^{\textnormal{true}}(f) &\leq R^{\textnormal{emp}}(f) + \sqrt{\frac{\log(|\mathcal{C}_{P,\Lambda}|) - \log(\delta)}{2N}}, \\
\intertext{where $\mathcal{C}_{P,\Lambda}$ denotes the set of Farey points of level $\Lambda$:}
\mathcal{C}_{P,\Lambda} &= \left\{ \frac{\lambdab}{q} \in [0,1)^P: (\lambdab,q) \in \mathbb{\hat{Z}}^{P+1} \text{ and } 1 \leq q \leq \Lambda \right\}.
\end{align*}
\end{thm}
The proof involves a counting argument over coprime integer vectors, using the definition of Farey points from number theory.

\begin{figure}[htbp]
\centering
\includegraphics[width=0.4\textwidth,keepaspectratio=true,trim=0mm 0mm 0mm 14mm,clip=true]{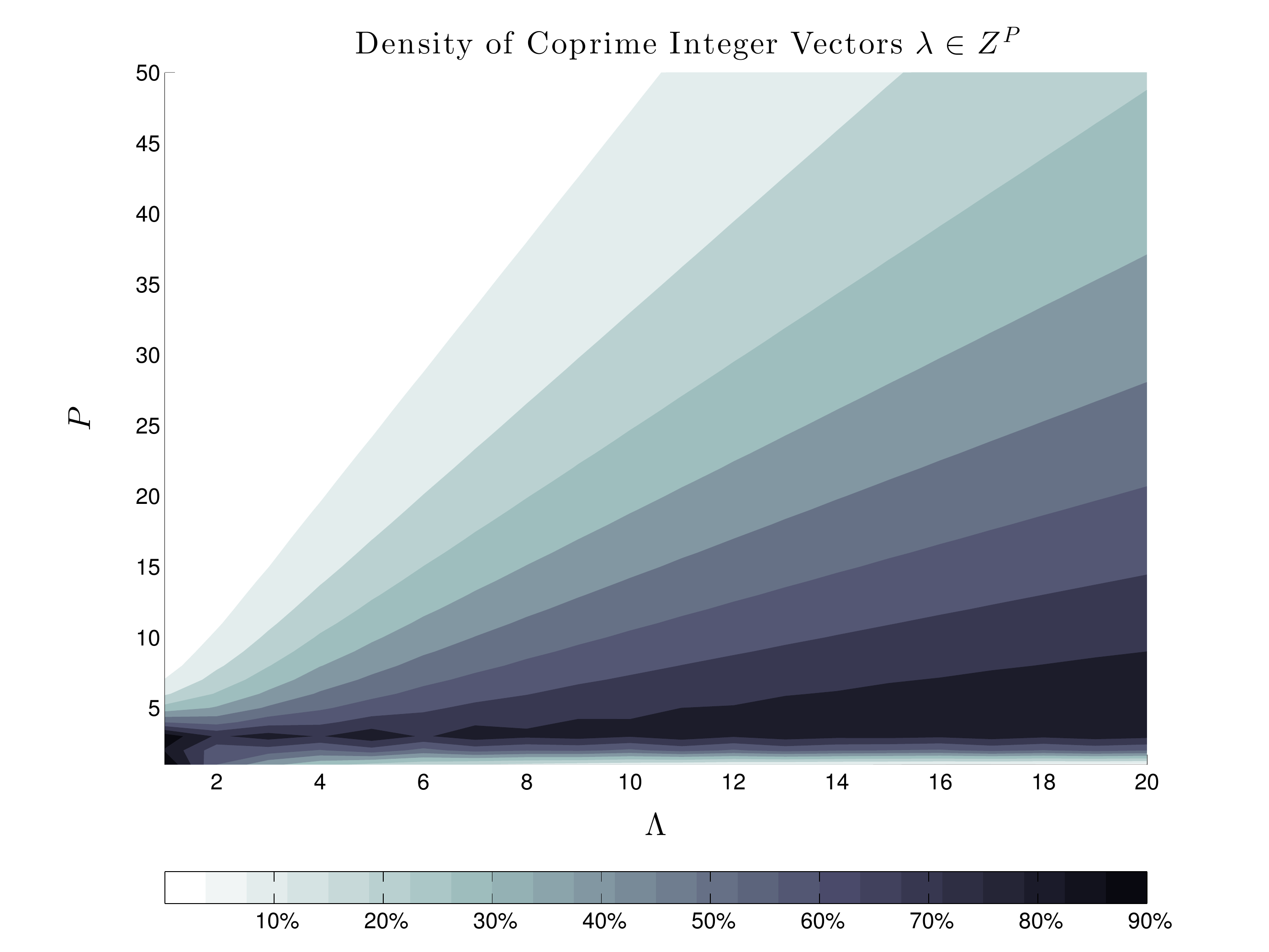}
\includegraphics[width=0.4\textwidth,keepaspectratio=true,trim=0mm 0mm 0mm 20mm,clip=true]{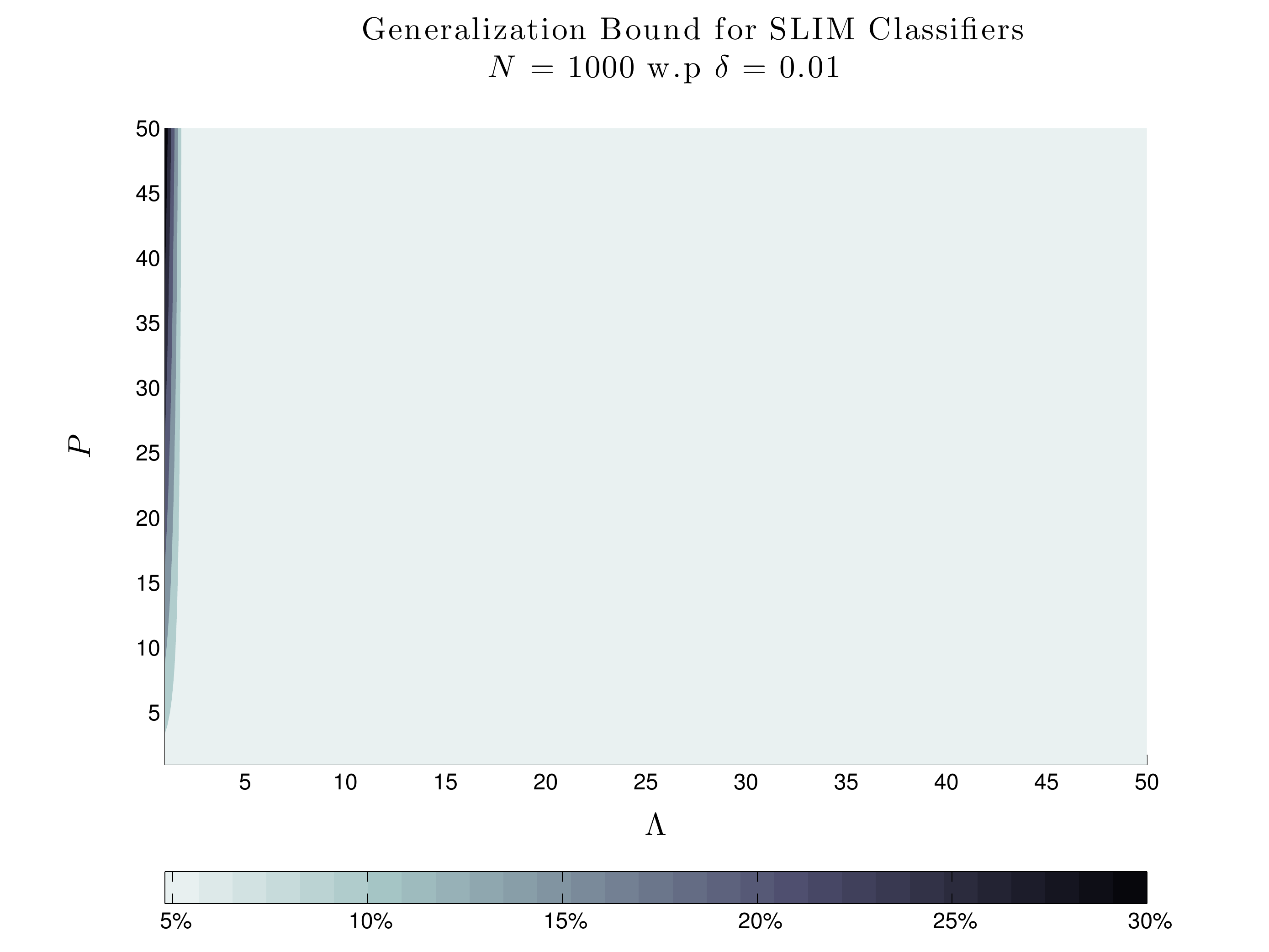}
\caption{Relative density of coprime integer vectors in $\Z^P$ (left), and the relative improvement in the generalization bound due to the use of coprime coefficients for $\delta=0.01$ (right).}
\label{Fig::CoprimeFigure}
\end{figure}

In Figure \ref{Fig::CoprimeFigure}, we plot the relative density of coprime integer vectors  bounded by $\Lambda$ (i.e., $|\mathcal{C}_{P,\Lambda}|/(2\Lambda+1)^P$), and the relative improvement in the generalization bound due to the use of coprime coefficients. We see that the use of coprime coefficients can significantly reduce the number of classifiers based on the dimensionality of the data and the value of $\Lambda$. The corresponding improvement in the generalization bound may be significant when the data are high dimensional and $\Lambda$ is small.%
\FloatBarrier
\section{Data Reduction}\label{Sec::Reduction}
Data reduction is a technique that can decrease the computation associated with training a supervised classification model by discarding redundant training data. This technique can be applied to any supervised classification method where the training procedure is carried out by solving an optimization problem. However, it is best suited for methods such as SLIM, where the underlying optimization problem may be difficult to solve for large instances. In this section, we first describe how data reduction works in a general setting, and then show how it can be applied to SLIM. 
\subsection{Data Reduction for Optimization-Based Supervised Classification}
Consider training a classifier $f: \X \to \Y$ by solving a computationally challenging optimization problem,
\begin{align}
\label{Eq::ReductionOriginal}
\min_{f} ~ Z(f;\data_N) ~\text{s.t.}~ f\in\F.
\end{align}
We refer to the optimization problem in \eqref{Eq::ReductionOriginal} as the \textit{original problem}. Here, $\F$ represents the set of feasible classifiers and $Z:\F  \times (\X\times\Y)^N \rightarrow\R$ represents its objective function.

Data reduction aims to decrease the computation associated with solving the original problem by removing redundant examples from $\data_N = (\xb_i,y_i)_{i=1}^N$ (i.e., data points that can be safely discarded without changing the optimal solution to \eqref{Eq::ReductionOriginal}). The technique requires users to specify a \textit{surrogate problem} that is considerably easier to solve. Given the initial training data $\data_N = (\xb_i,y_i)_{i=1}^N$, and the surrogate problem, data reduction solves $N+1$ variants of the surrogate problem to identify redundant examples. These examples are then removed from the initial training data to leave behind a subset of reduced training data $\data_M \subseteq \data_N$ that is guaranteed to yield the same optimal classifier as $\data_N$. Thus, the computational gain from data reduction comes from training a model with $\data_M$ (i.e., solving an instance of the original problem with $N-M$ fewer examples).

We provide an overview of data reduction in Algorithm \ref{Alg::DataReduction}. To explain how the algorithm works, let us denote the surrogate problem as:
\begin{align}
\label{Eq::ReductionProxy}
\min_{f} ~ \tilde{Z}(f;\data_N) ~\text{s.t}~ f\in\tilde{\F}.
\end{align}
Here $\tilde{Z}: \tilde{\F} \times (\X\times\Y)^N \rightarrow\R$ denotes the objective function of the surrogate problem, and $\tilde{\F}$ denotes its set of feasible classifiers. Data reduction can be used with any surrogate problem so long as the $\varepsilon$-level set of the surrogate problem contains all optimizers to the original problem. That is, we can use any feasible set $\tilde{\F}$ and any objective function $\tilde{Z}(.)$ as long as we can specify a value of $\varepsilon$ such that 
\begin{align}
\hspace{0.3\textwidth} & \tilde{Z}(f^*) \leq  \tilde{Z}(\tilde{f}^*) + \varepsilon & \forall f^* \in \F^* \text{ and } \tilde{f}^* \in \tilde{\F}^*. \label{Eq::ReductionLevelSet}
\end{align}
Here, $f^*$ denotes an optimal classifier to the original problem from the set $\F^* = \argmin_{f\in\F} Z(f)$, and $\tilde{f}^*$ denotes an optimal classifier to the surrogate problem from the set $\tilde{\F}^* = \argmin_{f\in\tilde{\F}} \tilde{Z}(f)$. The width of the the surrogate level set $\varepsilon$ is related to the amount of data that will be removed. If $\varepsilon$ is too large, the method will not remove very many examples and will be less helpful for reducing computation (see Figure \ref{Fig::BankruptcyDataReduction}). 

\begin{figure}[htbp]
\centering
\includegraphics[width=0.5\textwidth,trim=5mm 22.5mm 5mm 22.5mm,clip=true]{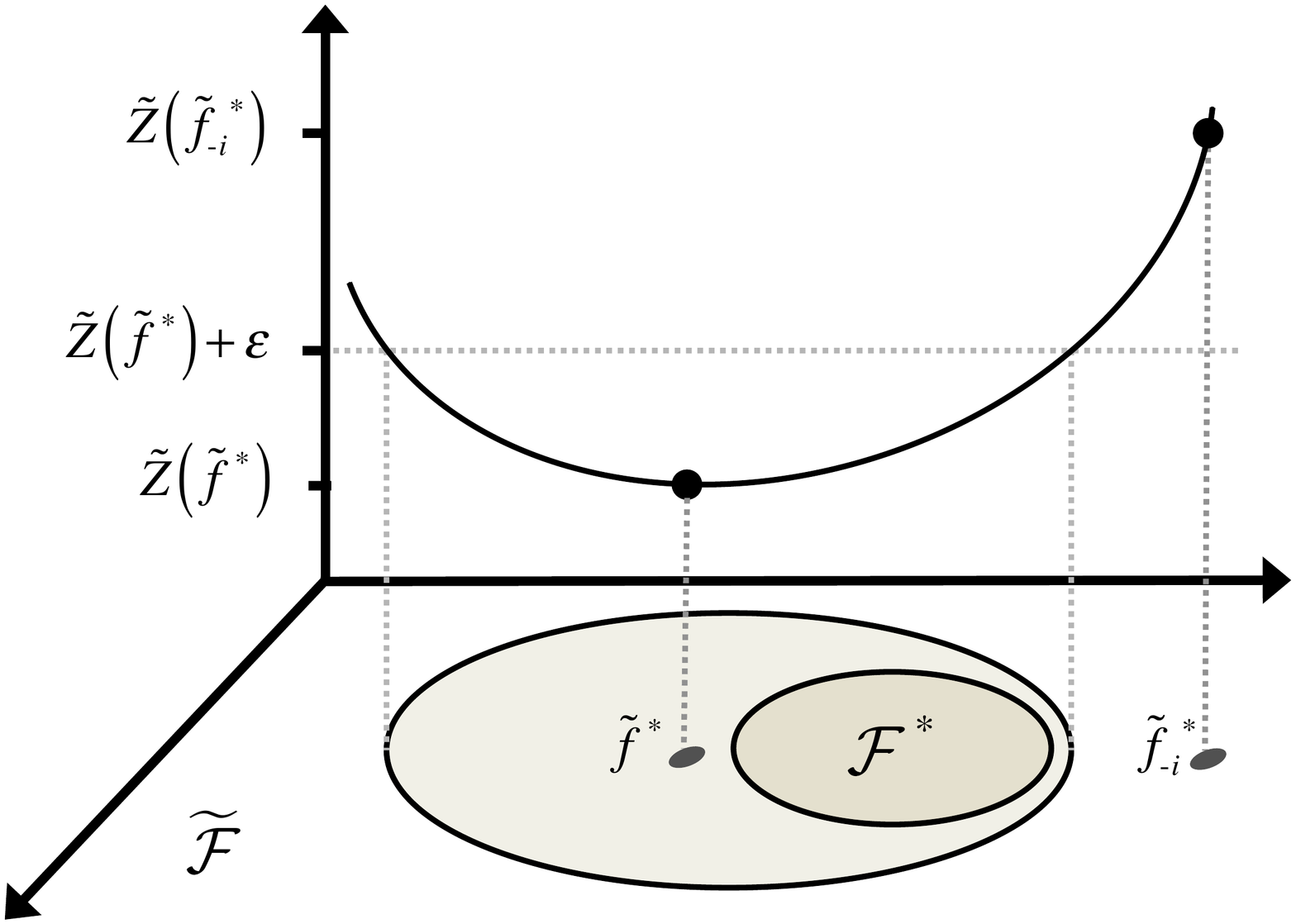}
\caption{We initialize data reduction with $\varepsilon$ large enough so that $\tilde{Z}(f^*)<\tilde{Z}(\tilde{f}^*)+\varepsilon$ for all $f^* \in \F^*$ and all $\tilde{f}^* \in \tilde{\F}^*$. Here, $f^*$ is the optimal classifer to the original problem from the set of optimal classifiers $\F^*$, and $\tilde{f}^*$ is the optimal classifier to the surrogate problem from the set of optimal classifiers $\tilde{\F}^*$. Data reduction fits a classifier $\tilde{f}^*_{\text{-}i}$ for each example in the initial training data $\data_N$ by solving a variant of the surrogate problem with an additional constraint that forces $\tilde{f}^*_{\text{-}i}$ to classify $i$ in a different way than $\tilde{f}^*$. If $\tilde{Z}(\tilde{f}^*_{\text{-}i})>\tilde{Z}(\tilde{f}^*)+\varepsilon$, then we know the predicted class of example $i$ under $f^*$ and can remove it from the reduced training data $\data_M$.}
\label{Fig::Reduction}
\end{figure}

In the first stage of data reduction, we solve the surrogate problem to: (i) compute the upper bound on the objective value of classifiers in the surrogate level set $\tilde{Z}(\tilde{f}^*)+\varepsilon$; and (ii) to identify a baseline label $\tilde{y}_i := \sign{\tilde{f}^*(\xb_i)}$ for each example $i = 1,\ldots,N$. In the second stage of data reduction, we solve a variant of the surrogate problem for each example $i = 1,\ldots,N$. The $i^\text{th}$ variant of the surrogate problem includes an additional constraint that forces example $i$ to be classified as $-\tilde{y}_i$:
\begin{align}
\label{Eq::DataReductionConvexVariant}
\min_{f} ~ \tilde{Z}(f;\data_N) ~\text{s.t}~ f\in\tilde{\F} \text{ and } \tilde{y}_i f(\xb_i) < 0
\end{align}
We denote the optimal classifier to the $i^\text{th}$ variant as $\tilde{f}^*_{\text{-}i}$. If $\tilde{f}^*_{\text{-}i}$ lies outside of the surrogate level set (i.e., $\tilde{Z}(\tilde{f}^*_{\text{-}i})>\tilde{Z}(\tilde{f}^*)+\varepsilon$) then no classifier in the surrogate level set will label point $i$ as $-\tilde{y}_i$. In other words, all classifiers in the surrogate level set must label this point as $\tilde{y}_i$. Since the surrogate level set contains the optimal classifiers to the original problem by the assumption in \eqref{Eq::ReductionLevelSet}, we can therefore remove example $i$ from the reduced dataset $\data_M$ because we know that an optimal classifier to the original problem will label this point as $\tilde{y}_i$. We illustrate this situation in Figure \ref{Fig::Reduction}.

In Theorem \ref{Thm::DataReduction}, we prove that we obtain the same set of optimal classifiers if we train a model with the initial data $\data_N$ or the reduced data $\data_M$. In Theorem \ref{Thm::ReductionSufficientConditions}, we provide sufficient conditions for a surrogate loss function to satisfy the level set condition in \eqref{Eq::ReductionLevelSet}. 
\vspace{-1em}
\begin{algorithm}
\caption{Data Reduction from $\data_N$ to $\data_M$\label{Alg::DataReduction}}
\begin{algorithmic}
  \REQUIRE initial training data, $\data_N = (\xb_i,y_i)_{i=1}^N$
  \REQUIRE surrogate problem, $\min \, \tilde{Z}(f;\data_N) \textrm{ s.t. }  f \in \tilde{\F}$
  \REQUIRE width of the surrogate level set, $\varepsilon$ 
  \STATE $\data_M \longleftarrow \emptyset$
  \STATE $\tilde{f}^* \longleftarrow \argmin_f \tilde{Z}(f;\data_N)$ 
  \FOR {$i = 1,\ldots,N$}
    \STATE $\tilde{y}_i \longleftarrow \sign{\tilde{f}^*(\xb_i)}$
    \STATE $\tilde{f}^*_{\text{-}i} \longleftarrow \argmin \tilde{Z}(f;\data_N) \; \st \; f \in \tilde{\F} \text{ and } \tilde{y}_i f(\xb_i) < 0$ 
           \IF {$\tilde{Z}(\tilde{f}^*_{\text{-}i};\data_N) \leq \tilde{Z}(\tilde{f}^*;\data_N) + \varepsilon$}
               \STATE $\data_M \longleftarrow \data_M \cup (\xb_i,y_i)$
            \ENDIF
            \ENDFOR
  \ENSURE $\data_M$, reduced training data
\end{algorithmic}
\end{algorithm}
\begin{thm}[Equivalence of the Reduced Data]\label{Thm::DataReduction}
Consider an optimization problem to train a classifier $f \in \F$ with data $\data_N$, $$\min_{f} ~ Z(f;\data_N) ~\textit{s.t}~ f\in\F,$$ as well as a surrogate optimization problem to train a classifier $f \in \tilde{\F}$ with data $\data_N$, $$\min_{f} ~ \tilde{Z}(f;\data_N) ~\textit{s.t.}~ f\in\tilde{\F}.$$
Let $f^* \in \F^* := \argmin_{f\in\F} Z(f;\data_N)$ and $\tilde{f} \in \tilde{\F}^* := \argmin_{f\in\tilde{\F}}  \tilde{Z}(f;\data_N)$. If we choose a value of $\varepsilon$ so that
\begin{align}
\label{Eq::ReductionAssumption}
\centering
\tilde{Z}(f^*;\data_N) &\leq \tilde{Z}(\tilde{f}^*;\data_N) + \varepsilon \quad \forall f^* \in \F^* \text{ and } \tilde{f}^* \in \tilde{\F}^*, 
\end{align}
then Algorithm \ref{Alg::DataReduction} will output a reduced dataset $\data_M \subseteq \data_N$ such that
\begin{align}\label{Eq::ReductionWTS}
\argmin_{f\in\F} Z(f;\data_N) = \argmin_{f\in\F} Z(f;\data_M).
\end{align}
\end{thm}
\proof See Appendix \ref{Appendix::Proofs}. \endproof
\newcommand{\z}[2]{Z_{#1}\left(#2\right)}
\newcommand{\Zmip}[1]{\z{01}{#1}}
\newcommand{\Zcvx}[1]{\z{\psi}{#1}}
\newcommand{\lmip}[0]{\lambdab^*_{01}}
\newcommand{\lcvx}[0]{\lambdab^*_{\psi}}
\begin{thm}[Sufficient Conditions to Satisfy the Level Set Condition]\label{Thm::ReductionSufficientConditions}
Consider an optimization problem where the objective minimizes the 0--1 loss function $Z_{01}: \R^P\rightarrow\R$, $$\min_{\lambdab \in \R^P} \Zmip{\lambdab},$$  as well as a surrogate optimization problem where the objective minimizes a surrogate loss function $\psi:\R^P\rightarrow\R$, $$\min_{\lambdab \in \R^P} \Zcvx{\lambdab}.$$  If the surrogate loss function $\psi$ satisfies the following properties for all $\lambdab \in \R^P$, $\lmip \in \argmin_{\lambdab \in \R^P} \Zmip{\lambdab}$, and $\lcvx \in \argmin_{\lambdab \in \R^P} \Zcvx{\lambdab}$:
\begin{enumerate}[leftmargin=2em,itemsep=3pt,topsep=3pt,label=\textnormal\emph{\Roman*.}]
   \item \textit{Upper bound on the 0--1 loss:} $\Zmip{\lambdab} \leq \Zcvx{\lambdab}$
   \item \textit{Lipschitz near $\lmip$:} $\| \lambdab - \lcvx \| < A \implies \Zcvx{\lambdab} - \Zcvx{\lcvx} < L \| \lambdab - \lcvx \| $
   \item \textit{Curvature near $\lcvx$:} $\| \lambdab - \lcvx \| > C_{\lambdab} \implies \Zcvx{\lambdab} - \Zcvx{\lcvx} > C_{\psi}$
   \item \textit{Closeness of loss near $\lmip$:} $| \Zcvx{\lmip} - \Zmip{\lmip}| < \varepsilon $
  \end{enumerate}
then it will also satisfy a level-set condition required for data reduction,
\begin{align*}
\hspace{0.25\textwidth} & \Zcvx{\lmip} \leq \Zcvx{\lcvx}+\varepsilon & \forall \lmip \text{ and } \lcvx, 
\end{align*}
whenever $\varepsilon = LC_{\lambdab}$ obeys $C_{\psi} > 2\varepsilon$. 
\end{thm}
\proof See Appendix \ref{Appendix::Proofs}. \endproof
\subsection{Off-The-Shelf Data Reduction for SLIM}\label{Sec::ReductionDemo}
Data reduction can easily be applied to SLIM by using an off-the-shelf approach where we use the LP relaxation of the SLIM IP as the surrogate problem. The off-the-shelf approach may be used as a preliminary procedure before the training process, or as an iterative procedure that is called by the IP solver during the training process as feasible solutions are found. 

When we use the LP relaxation to the SLIM IP as the surrogate problem, we can determine a suitable width for the surrogate level set $\varepsilon$ by using a feasible solution to the SLIM IP. To see this, let us denote the SLIM IP as $\min_f Z(f) \;\st\; f \in \F$, and denote its LP relaxation as $\min_f Z(f) \;\st\; f \in \tilde\F$. In addition, let us denote the optimal solution to the SLIM IP as $f^*$ and the optimal solution to the LP relaxation as $\tilde{f}^*$. Since $\F \subseteq \tilde{\F}$, we have that $Z(\tilde{f}^*) \leq Z(f^*)$. For any feasible solution to the SLIM IP $\hat{f} \in \F$, we also have that $Z(f^*) \leq  Z(\hat{f})$. Combining both inequalities, we see that, $$Z(\tilde{f}^*) \leq Z(f^*) \leq Z(\hat{f}).$$ Thus, we can satisfy the level set condition \eqref{Eq::ReductionLevelSet} using a feasible solution to the SLIM IP $\hat{f} \in \F$ by setting the width of the surrogate level set as $$\varepsilon(\hat{f}) := Z(\hat{f}) - Z(\tilde{f}^*).$$

\begin{figure}[htbp]
\centering
\includegraphics[width=0.425\textwidth,trim=0mm 2.5mm 0mm 2.5mm,clip=true]{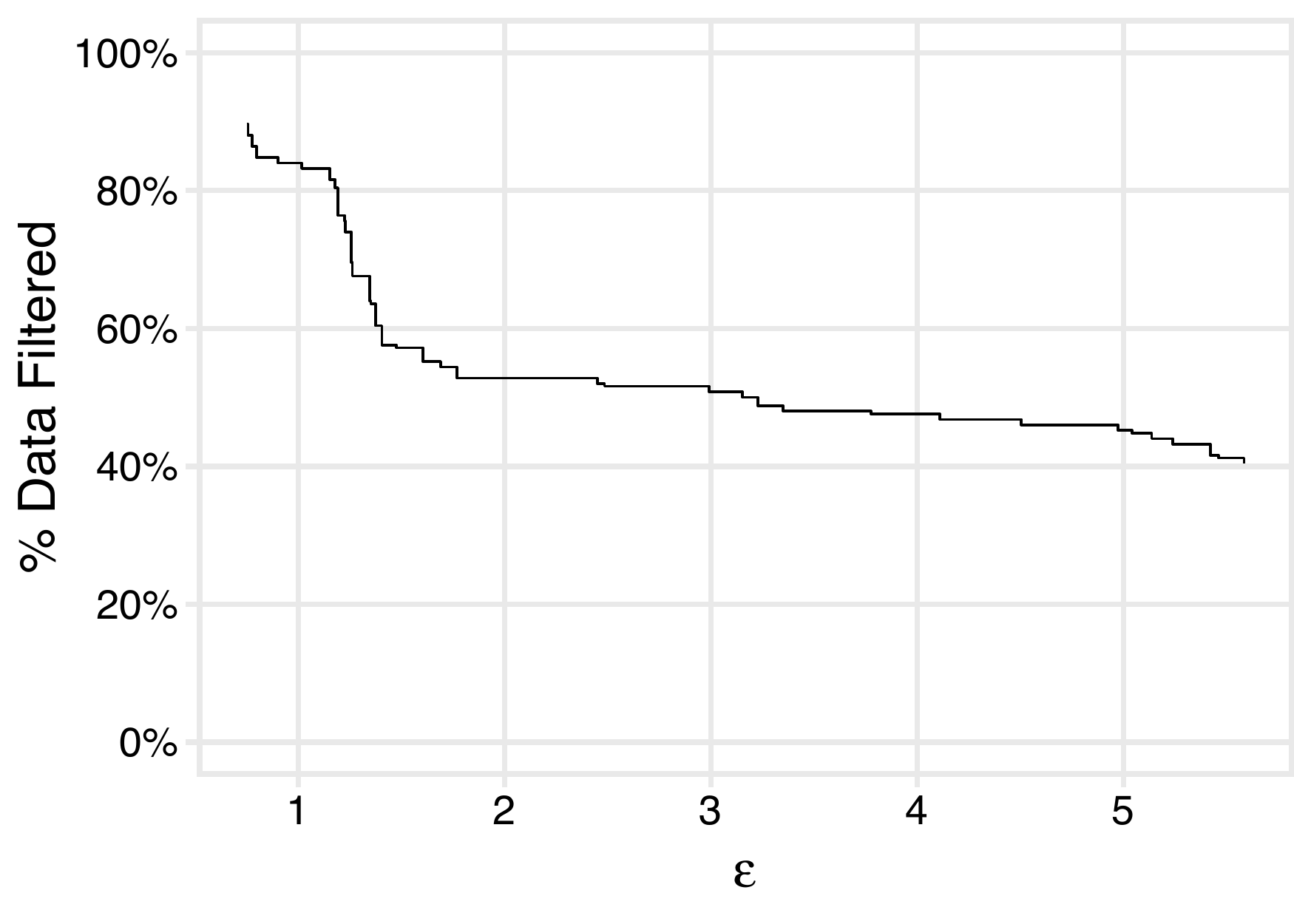}
\caption{Proportion of training data filtered as a function of the width of the level set, $\varepsilon$ for the \textds{bankruptcy} dataset. Here, the original problem is an instance of the SLIM IP with $C_0 = 0.01$ and $\Lset = \{-10,\ldots,10\}^{P+1}.$}
\label{Fig::BankruptcyDataReduction}
\end{figure}

In Figure \ref{Fig::BankruptcyDataReduction}, we show much training data can be discarded using off-the-shelf data reduction when we train a SLIM scoring system on the \textds{bankruptcy} dataset (see Table \ref{Table::ExperimentalDatasets}). Specifically, we plot the percentage of data removed by Algorithm 1 for values of $\varepsilon \in [\varepsilon_{\min},\varepsilon_{\max}]$ where $\varepsilon_{\min}$ and $\varepsilon_{\max}$ represent the smallest and largest widths of the surrogate level set that could be used in practice. In particular, $\varepsilon_{\min}$ is computed using the optimal solution to the IP as: $$\varepsilon_{\min} := Z(f^*) -  Z(\tilde{f}^*),$$ and $\varepsilon_{\max}$ is computed using a feasible solution to the IP that can be guessed without any computation (i.e., a linear classifier with coefficients $\lambdab = \textbf{0}$): $$\varepsilon_{\max} := Z(\textbf{0}) -  Z(\tilde{f}^*).$$
In this case, we can discard over 40\% of the training data by using the trivial solution $\lambdab=0$, and discard over 80\% of the training data by using a higher quality feasible solution. 
%
%
\FloatBarrier
\section{Application to Sleep Apnea Screening}\label{Sec::SleepApnea}
In this section, we discuss a collaboration with the MGH Sleep Laboratory where we used SLIM to create a scoring system for sleep apnea screening \citep[see also][for a far more detailed treatment]{ustun2015apnea}. Our goal is to highlight the flexibility and performance of our approach on a real-world problem that requires a tailored prediction model.
\subsection{Data and Operational Constraints}
The dataset for this application contains $N = 1922$ records of patients and $P = 33$ binary features related to their health and sleep habits. Here, $y_i=+1$ if patient $i$ has obstructive sleep apnea (OSA) and $y_i=-1$ otherwise. There is significant class imbalance as Pr$(y_i=+1) = 76.9\%$. 

To ensure that the scoring system we produced would be used and accepted by physicians, our collaborators specified three simple operational constraints:
\begin{enumerate}[leftmargin=0.40cm,topsep=2pt,parsep=2pt]

\item \textit{Limited FPR:} The model had to achieve the highest possible true positive rate (TPR) while maintaining a maximum false positive rate (FPR) of 20\%. This would ensure that the model could diagnose as many cases of sleep apnea as possible but limit the number of faulty diagnoses. 

\item \textit{Limited Model Size:} The model had to be transparent and use at most 5 features. This would ensure that the model was could be explained and understood by other physicians in a short period of time.

\item \textit{Sign Constraints:} The model had to obey established relationships between well-known risk factors and the incidence of sleep apnea (e.g. it could not suggest that a patient with hypertension had a lower risk of sleep apnea since hypertension is a positive risk factor for sleep apnea).

\end{enumerate}
\subsection{Training Setup and Model Selection}
We trained a SLIM scoring system with integer coefficients between $-10$ and $10$. We addressed all three operational constraints without parameter tuning or model selection, as follows:
\begin{enumerate}[leftmargin=0.40cm,topsep=2pt,parsep=2pt,label=$\bullet$]

\item We added a loss constraint using the loss variables to limit the maximum FPR at 20\%. We then set $\wplus = \nminus/(1+\nminus)$ to guarantee that the optimization would yield a classifier with the highest possible TPR with a maximum FPR less than 20\% (see Section \ref{Sec::HandlingImbalancedData}).

\item We added a feature-based constraint using the loss variables to limit the maximum number of features to 5 (see Section \ref{Sec::FeatureBasedConstraints}). We then set $C_0 = 0.9\wminus/NP$ so that the optimization would yield a classifier that did not sacrifice accuracy for sparsity (see Remark \ref{Rem::TradeOffParamaterMin}).

\item We added sign constraints to the coefficients to ensure that our model would not violate established relationships between features and the predicted outcome (i.e., we set $\lambda_j \geq 0$ if there had to be a positive relationship, and $\lambda_j \leq 0$ if there had to be a negative relationship).

\end{enumerate}
With this setup, we trained 10 models with subsets of the data to assess predictive accuracy via 10-fold cross validation (10-CV), and 1 final model with all of data to hand over to our collaborators. We set up each IP using the slim\_for\_matlab toolbox \citep{ustun2015slimformatlab} and solved each IP for 1 hour, in parallel, on 12-core 2.7GHZ machine with 48GB RAM. Thus, the training process for SLIM required 1 hour of computing time.

As a comparison, we trained models with 8 baseline classification methods shown in Table \ref{Table::SleepApneaSetup}. We dealt with the class imbalance by using a cost-sensitive approach, where we used a weighted loss function and varied its sensitivity parameter $\wplus$ across a large range. When possible, we addressed the remaining operational constraints by searching over a fine grid of free parameters. Model selection was difficult for baseline methods because they could not accomodate operational constraints in the same way as SLIM. For each baseline method, we chose the best model that satisfied all operational constraints by: (i) dropping any instance of the free parameters where operational constraints were violated; (ii) choosing the instance that maximized the 10-CV mean test TPR. We ruled that an instance of the free parameters violated an operational constraint if any of the following conditions were met: (1) the 10-CV mean test FPR of the model produced with the instance was greater than the 10-CV mean test FPR of the SLIM model (20.9\%); (2) the model size%
\footnote{Model size represents the number of coefficients for linear models (Lasso, Ridge, Elastic Net, SLIM, SVM Lin.), the number of leaves for decision tree models (C5.0T, CART), and the number of rules for rule-based models (C5.0R). For completeness, we set the model size for black-box models (SVM RBF) to the number of features in each dataset.}
of the final model produced with the instance was greater than 5; (3) the final model produced did not obey sign constraints. This model selection procedure may have biased the results in favor of the baseline methods because we mixed testing and training data by looking at the final model to ensure that operational constraints were satisfied.
%
\begin{table}[htbp]
\scriptsize
\centering
\begin{tabular}{lccl}
\toprule 

\textbf{Method} & \bfcell{c}{Controls} & \bfcell{c}{\# Instances} & \bfcell{l}{Settings and Free Parameters} \\ 

\midrule

CART & \cell{c}{Max FPR\\Model Size} & 39 &  

\cell{l}{
39 values of $\wplus \in \{0.025,0.05,\ldots,0.975\}$
} \\

\midrule

C5.0T & \cell{l}{Max FPR} & 39 &

\cell{l}{
39 values of $\wplus \in \{0.025,0.05,\ldots,0.975\}$
} \\ 

\midrule

C5.0R & \cell{c}{Max FPR\\Model Size} & 39 &

\cell{l}{
39 values of $\wplus \in \{0.025,0.05,\ldots,0.975\}$
} \\ 

\midrule

Lasso & \cell{c}{Max FPR\\Model Size\\Signs} & 39000 & 

\cell{l}{
39 values of $\wplus \in \{0.025,0.05,\ldots,0.975\}$ \\
$\times$ 1000 values of $\lambda$ chosen by \pkg{glmnet}
} \\ 

\midrule

Ridge & \cell{c}{Max FPR\\Signs} & 39000 & 

\cell{l}{
39 values of $\wplus \in \{0.025,0.05,\ldots,0.975\}$ \\
$\times$ 1000 values of $\lambda$ chosen by \pkg{glmnet}
} \\ 

\midrule

Elastic Net & \cell{c}{Max FPR\\Model Size\\Signs}  & 975000 &

\cell{l}{
39 values of $\wplus \in \{0.025,0.05,\ldots,0.975\}$ \\
$\times$ 1000 values of $\lambda$ chosen by \pkg{glmnet}\\
$\times$ 19 values of $\alpha \in \{0.05,0.10,\ldots,0.95\}$
} \\ 

\midrule

SVM Lin.  & \cell{c}{Max FPR} &  975 & 

\cell{l}{
39 values of $\wplus \in \{0.025,0.05,\ldots,0.975\}$ \\
$\times$ 25 values of  $C \in \{10^{-3},10^{-2.75},\ldots,10^3\}$
} \\ 

\midrule

SVM RBF & \cell{c}{Max FPR} & 975 & 

\cell{l}{
39 values of $\wplus \in \{0.025,0.05,\ldots,0.975\}$ \\
$\times$ 25 values of $C \in \{10^{-3},10^{-2.75},\ldots,10^3\}$
} \\ 

\midrule

SLIM & \cell{c}{Max FPR\\Model Size\\Signs} & 1 &

\cell{l}{
$\wplus = \nminus/(1+\nminus)$,  $C_0 = 0.9\wminus/NP$, \\
$\lambda_0 \in \{-100,\ldots,100\}$, $\lambda_j \in \{-10,\ldots,10\}$} \\ 

\bottomrule 
\end{tabular}
\caption{Training setup for all methods. An instance is a unique combination of free parameters. Controls refer to operational constraints that we expect each method to handle. We include further details on methods and software packages in Table \ref{Table::ExperimentalMethods}.}
\label{Table::SleepApneaSetup}
\end{table}
\subsection{Results and Observations}
\begin{figure}[htbp]
\centering
\includegraphics[height=0.225\textheight,trim=0mm 2.5mm 0mm 0mm,clip]{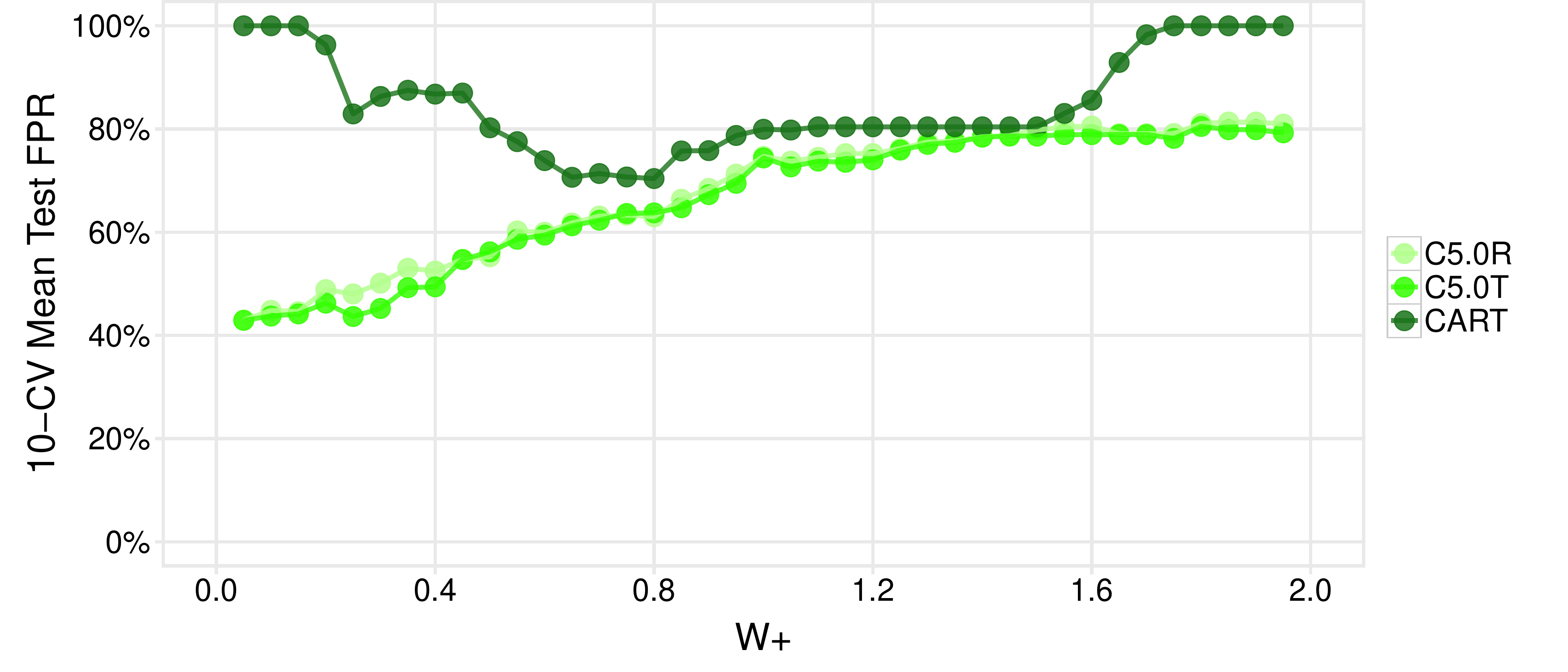} 
\caption{10-CV mean test FPR for models trained with CART, C5.0, C5.0T across the full range of $\wplus$. These methods cannot produce a model that satisfies the max FPR $\leq 20 \%$ constraint.}
\label{Fig::WRangeGraph}
\end{figure}

\begin{figure}
\centering
\includegraphics[width=0.425\textwidth,trim=0mm 2.5mm 0mm 5mm,clip]{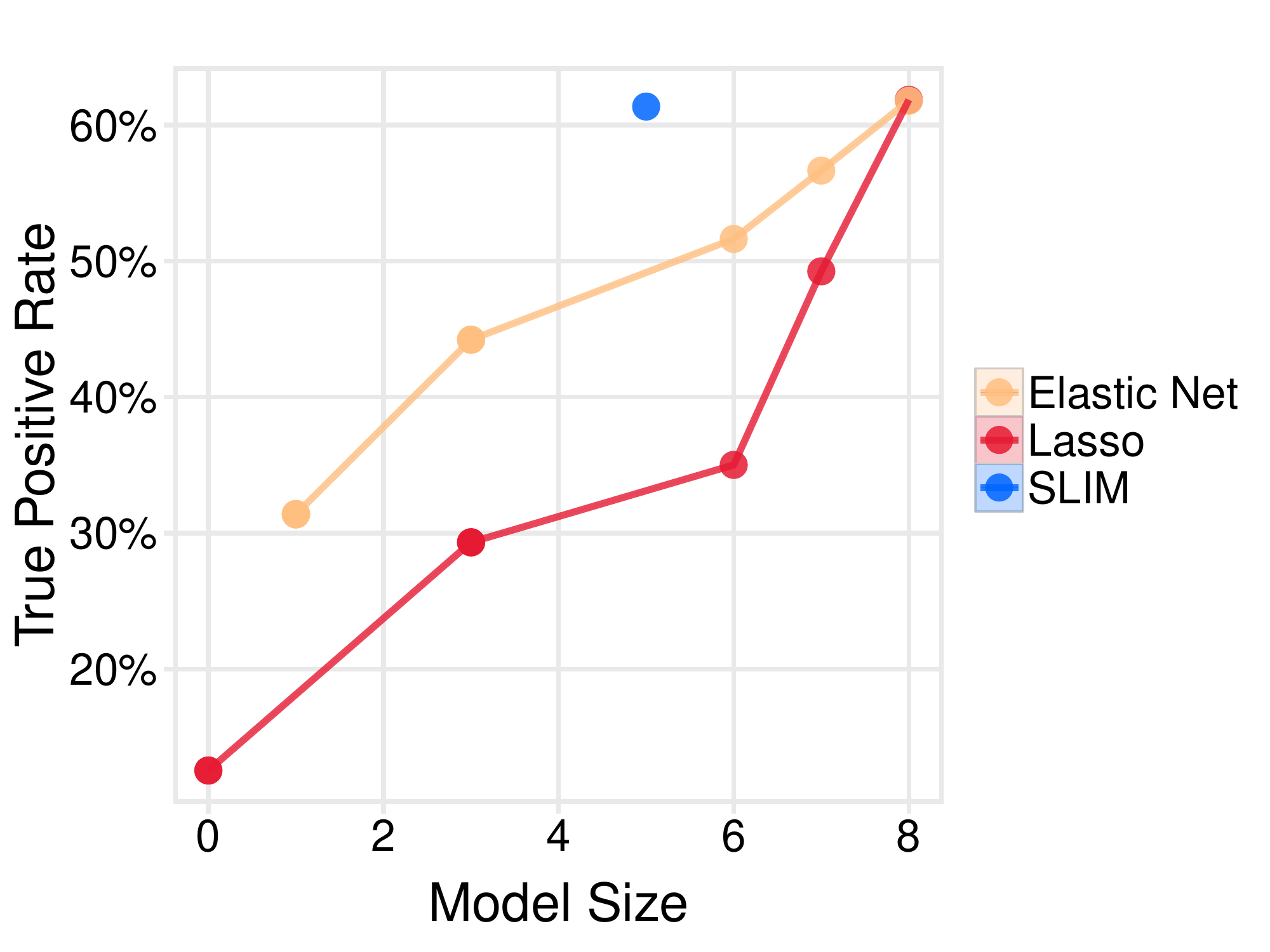}
\caption{Sensitivity and model size of Lasso and Elastic Net models that satisfy the sign and FPR constraints. For each method, we plot the instance that attains the highest 10-CV mean test TPR at model sizes between 0 and 8. Lasso and Elastic Net need at least 8 coefficients to produce a model with the same sensitivity as SLIM.}
\label{Figure::SleepApneaRegPath}
\end{figure}

\begin{figure}
\centering
\includegraphics[width=0.425\textwidth,trim=0mm 7.5mm 0mm 5mm,clip]{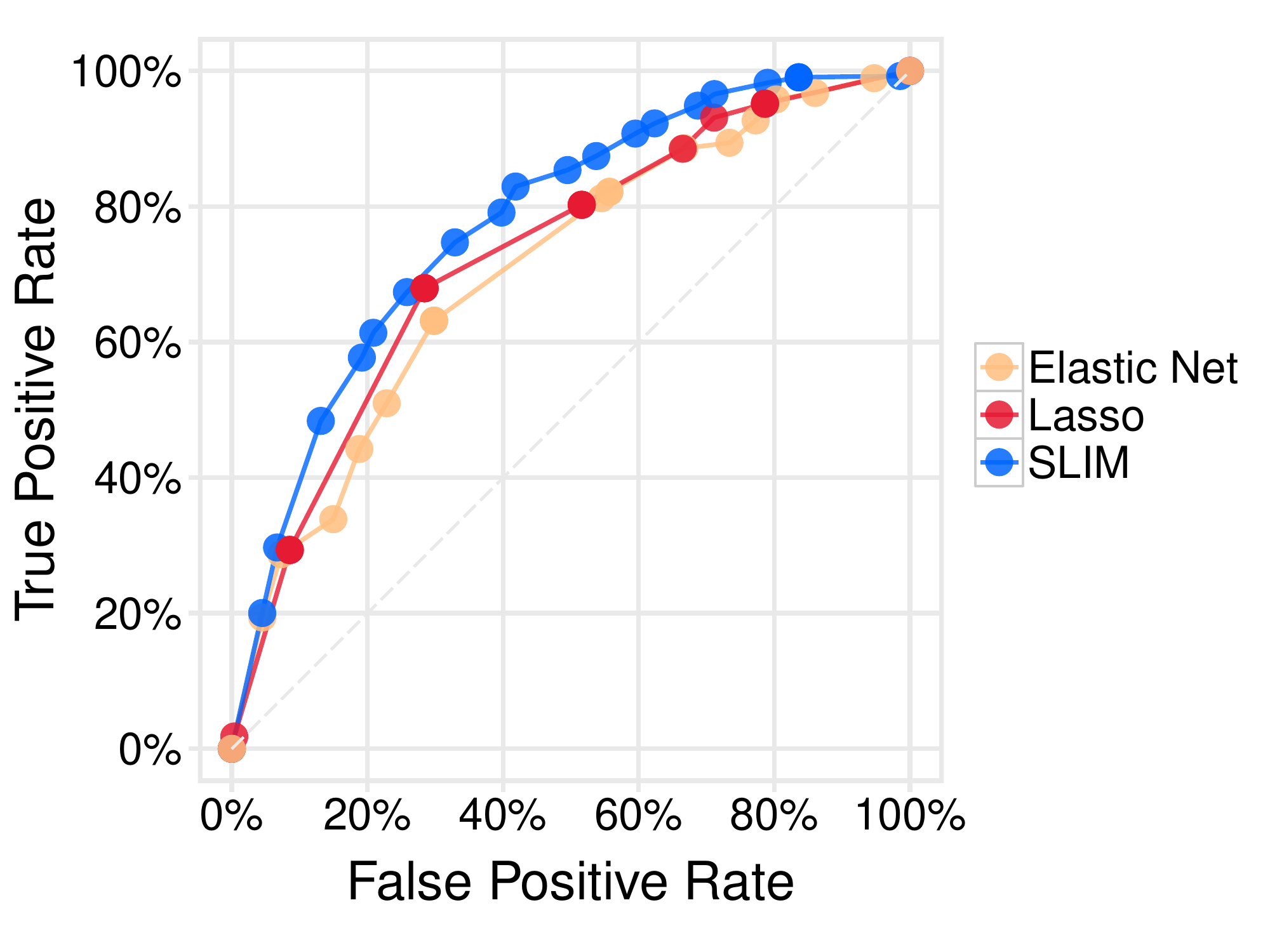}
\caption{ROC curve for SLIM, Lasso and Elastic Net instances that satisfy the sign and model size constraints. For each method, we plot the instance that attains the highest 10-CV mean test TPR for 10-CV mean FPR values of $5\%,10\%,\ldots,95\%$. Note that we had to train 19 additional instances of SLIM to create this plot.}
\label{Figure::SleepApneaROC}
\end{figure}

In what follows, we report our observations related to operational constraints, predictive performance and interpretability. We show the performance of the best model we trained using each method in Table \ref{Table::DemoTableMaxFPR}, and summarize the operational constraints they were able to satisfy in Table \ref{Table::DemoFlexibilityTable}.
\begin{table}[ht]
\centering
{\scriptsize
\begin{tabular}{l@{ }c@{ }c@{ }ccccccc}
  \toprule
 & & \multicolumn{1}{c}{\textbf{OBJECTIVE}} & \multicolumn{2}{c}{\textbf{CONSTRAINTS}}& \multicolumn{5}{c}{\textbf{OTHER INFORMATION}} \\ \cmidrule(lr){3-3}\cmidrule(lr){4-5}\cmidrule(lr){6-10}
\bfcell{c}{Method} & \bfcell{c}{Constraints\\Satisfied} & \bfcell{c}{Test\\TPR} & \bfcell{c}{Test\\FPR} & \bfcell{c}{Final\\Model\\Size} & \bfcell{c}{Model\\Size} & \bfcell{c}{Train\\TPR} & \bfcell{c}{Train\\FPR} & \bfcell{c}{Final\\Train\\TPR} & \bfcell{c}{Final\\Train\\FPR} \\ 
  
\toprule

SLIM & \scriptsize{\cell{c}{All} }
& \cell{c}{\scriptsize{61.4$\%$}\\\tiny{55.5 - 68.8$\%$}} 
& \cell{c}{\scriptsize{20.9$\%$}\\\tiny{15.0 - 30.4$\%$}} 
& \cell{c}{\scriptsize{5}\\\tiny{-}} 
& \cell{c}{\scriptsize{5}\\\tiny{5 - 5}}
& \cell{c}{\scriptsize{62.4$\%$}\\\tiny{61.0 - 64.2$\%$}} 
& \cell{c}{\scriptsize{19.7$\%$}\\\tiny{19.3 - 20.0$\%$}} 
& \cell{c}{\scriptsize{62.0$\%$}\\\tiny{-}} 
& \cell{c}{\scriptsize{19.6$\%$}\\\tiny{-}} \\

\midrule

Lasso & \scriptsize{\cell{c}{All}}
& \cell{c}{\scriptsize{29.3$\%$}\\\tiny{19.2 - 60.0$\%$}} 
& \cell{c}{\scriptsize{8.6$\%$}\\\tiny{0.0 - 33.3$\%$}} 
& \cell{c}{\scriptsize{3}\\\tiny{-}}
& \cell{c}{\scriptsize{3}\\\tiny{3 - 6}}
& \cell{c}{\scriptsize{28.7$\%$}\\\tiny{21.4 - 54.6$\%$}} 
& \cell{c}{\scriptsize{7.2$\%$}\\\tiny{3.5 - 20.5$\%$}} 
& \cell{c}{\scriptsize{22.1$\%$}\\\tiny{-}} 
& \cell{c}{\scriptsize{3.8$\%$}\\\tiny{-}} \\

\midrule

Elastic Net & \scriptsize{\cell{c}{All}}
& \cell{c}{\scriptsize{44.2$\%$}\\\tiny{0.0 - 64.1$\%$}} 
& \cell{c}{\scriptsize{18.8$\%$}\\\tiny{0.0 - 37.0$\%$}} 
& \cell{c}{\scriptsize{3}\\\tiny{-}} 
& \cell{c}{\scriptsize{3}\\\tiny{3 - 6}} 
& \cell{c}{\scriptsize{45.6$\%$}\\\tiny{0.0 - 66.5$\%$}} 
& \cell{c}{\scriptsize{17.4$\%$}\\\tiny{0.0 - 36.4$\%$}} 
& \cell{c}{\scriptsize{54.3$\%$}\\\tiny{-}} 
& \cell{c}{\scriptsize{20.7$\%$}\\\tiny{-}} \\

\midrule
   
Ridge & \scriptsize{\cell{c}{Max FPR}} 
& \cell{c}{\scriptsize{66.0$\%$}\\\tiny{60.5 - 68.5$\%$}} 
& \cell{c}{\scriptsize{20.6$\%$}\\\tiny{8.6 - 32.6$\%$}} 
& \cell{c}{\scriptsize{30}\\\tiny{-}}
& \cell{c}{\scriptsize{30}\\\tiny{30 - 30}} 
& \cell{c}{\scriptsize{66.4$\%$}\\\tiny{64.0 - 68.9$\%$}} 
& \cell{c}{\scriptsize{18.9$\%$}\\\tiny{17.3 - 21.5$\%$}} 
& \cell{c}{\scriptsize{66.0$\%$}\\\tiny{-}} 
& \cell{c}{\scriptsize{18.9$\%$}\\\tiny{-}}  \\

\midrule

SVM RBF & \scriptsize{\cell{c}{Max FPR}} 
& \cell{c}{\scriptsize{64.3$\%$}\\\tiny{59.2 - 71.1$\%$}} 
& \cell{c}{\scriptsize{20.8$\%$}\\\tiny{10.0 - 30.4$\%$}} 
& \cell{c}{\scriptsize{33}\\\tiny{-}}
& \cell{c}{\scriptsize{33}\\\tiny{33 - 33}} 
& \cell{c}{\scriptsize{67.9$\%$}\\\tiny{66.5 - 70.0$\%$}} 
& \cell{c}{\scriptsize{12.2$\%$}\\\tiny{11.1 - 13.3$\%$}} 
& \cell{c}{\scriptsize{67.8$\%$}\\\tiny{-}} 
& \cell{c}{\scriptsize{12.4$\%$}\\\tiny{-}} \\ 
   
\midrule

SVM Lin. & \scriptsize{\cell{c}{Max FPR}} 
& \cell{c}{\scriptsize{62.7$\%$}\\\tiny{57.9 - 69.0$\%$}} 
& \cell{c}{\scriptsize{19.8$\%$}\\\tiny{7.5 - 28.6$\%$}} 
& \cell{c}{\scriptsize{31}\\\tiny{-}}
& \cell{c}{\scriptsize{31}\\\tiny{31 - 31}}
& \cell{c}{\scriptsize{63.7$\%$}\\\tiny{61.5 - 66.1$\%$}} 
& \cell{c}{\scriptsize{17.0$\%$}\\\tiny{15.6 - 18.5$\%$}} 
& \cell{c}{\scriptsize{63.1$\%$}\\\tiny{-}} 
& \cell{c}{\scriptsize{17.1$\%$}\\\tiny{-}} \\

\midrule

C5.0R & None
& \cell{c}{\scriptsize{84.0$\%$}\\\tiny{78.9 - 87.7$\%$}} 
& \cell{c}{\scriptsize{43.0$\%$}\\\tiny{32.6 - 54.2$\%$}} 
& \cell{c}{\scriptsize{26}\\\tiny{-}} 
& \cell{c}{\scriptsize{23}\\\tiny{18 - 30}}
& \cell{c}{\scriptsize{86.1$\%$}\\\tiny{84.2 - 88.5$\%$}} 
& \cell{c}{\scriptsize{33.8$\%$}\\\tiny{30.9 - 38.2$\%$}} 
& \cell{c}{\scriptsize{85.5$\%$}\\\tiny{-}} 
& \cell{c}{\scriptsize{32.9$\%$}\\\tiny{-}} \\
   
\midrule

C5.0T & None & \cell{c}{\scriptsize{81.3$\%$}\\\tiny{77.4 - 84.8$\%$}} 
& \cell{c}{\scriptsize{42.9$\%$}\\\tiny{29.6 - 62.5$\%$}} 
& \cell{c}{\scriptsize{39}\\\tiny{-}}
& \cell{c}{\scriptsize{42}\\\tiny{39 - 50}} 
& \cell{c}{\scriptsize{85.3$\%$}\\\tiny{82.6 - 88.6$\%$}} 
& \cell{c}{\scriptsize{29.5$\%$}\\\tiny{24.6 - 33.7$\%$}} 
& \cell{c}{\scriptsize{84.5$\%$}\\\tiny{-}} 
& \cell{c}{\scriptsize{28.4$\%$}\\\tiny{-}} \\
   
\midrule

CART & None 
& \cell{c}{\scriptsize{93.0$\%$}\\\tiny{88.8 - 96.1$\%$}} 
& \cell{c}{\scriptsize{70.4$\%$}\\\tiny{61.1 - 83.3$\%$}} 
& \cell{c}{\scriptsize{8}\\\tiny{-}}
& \cell{c}{\scriptsize{9}\\\tiny{4 - 12}}
& \cell{c}{\scriptsize{95.2$\%$}\\\tiny{93.1 - 97.2$\%$}} 
& \cell{c}{\scriptsize{66.8$\%$}\\\tiny{55.0 - 76.0$\%$}} 
& \cell{c}{\scriptsize{95.9$\%$}\\\tiny{-}} 
& \cell{c}{\scriptsize{73.9$\%$}\\\tiny{-}} \\
   
\bottomrule
\end{tabular}
}
\caption{TPR, FPR and model size for all methods. We report the 10-CV mean TPR and FPR, and the 10-CV median for the model size. The ranges in each cell represent the 10-CV minimum and maximum.}
\label{Table::DemoTableMaxFPR}
\end{table}
\shortsection{On the Difficulties of Handling Operational Constraints}
Among the 9 classification methods that we used, only SLIM, Lasso and Elastic Net could produce a model that satisfied all of operational constraints given to us by physicians. Tree and rule-based methods such as CART, C5.0 Tree and C5.0 Rule were unable to produce a model with a maximum FPR of 20\% (see Figure \ref{Fig::WRangeGraph}). Methods that used $\ltwo$-regularization such as Ridge, SVM Lin. and SVM RBF were unable to produce a model with the required level of sparsity. While we did not expect all methods to satisfy all of the operational constraints, we included them to emphasize the following important points. Namely, state-of-the-art methods for applied predictive modeling do not:
\begin{itemize}[leftmargin=0.40cm,topsep=2pt,parsep=2pt,label=$\bullet$]

\item Handle simple operational constraints that are crucial for models to be used and accepted. Implementations of popular classification methods do not have a mechanism to adjust important model qualities. That is, there is no mechanism to control sparsity in C5.0T (\citealt{kuhn2012c50}) and no mechanism to incorporate sign constraints in SVM (\citealt{meyer2012e1071}). Finding a method with suitable controls is especially difficult when a model has to satisfy multiple operational constraints.

\item Have controls that are easy-to-use and/or that work correctly. When a method has suitable controls to handle operational constraints, producing a model often requires a tuning process over a high-dimensional free parameter grid. Even after extensive tuning, however, it is possible to never find a model that satisfies all operational constraints (e.g. CART, C5.0R, C5.0T for the Max FPR constraint in Figure \ref{Fig::WRangeGraph}).

\item  Allow tuning to be portable when the training set changes. Consider a standard model selection procedure where we choose free parameters to maximize predictive accuracy. In this case, we would train models on several folds for each instance of the free parameters, choose an instance of the free parameters that maximized our estimate of predictive accuracy among the instances that met all operational constraints, and then train a final model using these values of the free parameters. Unfortunately, there is no guarantee that the final model will obey all operational constraints. 

\end{itemize}
%
\begin{table}[htbp]
\centering
{\scriptsize
\begin{tabular}{lccc}
  \toprule 
  & \multicolumn{3}{c}{\textbf{\% of Instances that Satisfied} }\\ 
  \cmidrule(lr){2-4}  
  \bfcell{c}{Method} & \bfcell{c}{Max FPR} & \bfcell{c}{Max FPR \& Model Size} & \bfcell{c}{Max FPR, Model Size \& Signs} \\ 
   \toprule SLIM & 100.0$\%$ & 100.0$\%$ & 100.0$\%$ \\ 
   \midrule Lasso & 19.6$\%$ & 4.8$\%$ & 4.8$\%$ \\ 
   \midrule Elastic Net & 18.3$\%$ & 1.0$\%$ & 1.0$\%$ \\ 
   \midrule  Ridge & 20.9$\%$ & 0.0$\%$ & 0.0$\%$ \\ 
   \midrule SVM Lin & 18.7$\%$ & 0.0$\%$ & 0.0$\%$ \\ 
   \midrule SVM RBF & 15.8$\%$ & 0.0$\%$ & 0.0$\%$ \\ 
   \midrule C5.0R  & 0.0$\%$ & 0.0$\%$ & 0.0$\%$ \\ 
   \midrule C5.0T & 0.0$\%$ & 0.0$\%$ & 0.0$\%$ \\ 
   \midrule CART & 0.0$\%$ & 0.0$\%$ & 0.0$\%$ \\ 
   \bottomrule \end{tabular}
}
\caption{Percentage of instances that fulfilled operational constraints. Each instance is a unique combination of free parameters for a given method.}
\label{Table::DemoFlexibilityTable}
\end{table}
\shortsection{On the Sensitivity of Acceptable Models}
Among the three methods that produced acceptable models, the scoring system produced by SLIM had significantly higher sensitivity than the models produced by Lasso and Elastic Net -- a result that we expected given that SLIM minimizes the 0--1 loss and an $\lzero$-penalty while Lasso and Elastic Net minimize convex surrogates of these quantities. This result held true even when we relaxed various operational constraints. In Figure \ref{Figure::SleepApneaRegPath}, for instance, we plot the sensitivity and sparsity of models that satisfied the max FPR and sign constraints. Here, we see that Lasso and Elastic Net need at least 8 coefficients to produce a model with the same degree of sensitivity as SLIM. In Figure \ref{Figure::SleepApneaROC}, we plot the TPR and FPR of models that satisfied the sign and model size constraints. As shown, SLIM scoring systems dominate Lasso and Elastic Net models across the entire ROC curve. These sensitivity advantages are also evident in Table \ref{Table::DemoTableMaxFPR}: in particular, SLIM yields a model with a similar level of sensitivity and specificity as Ridge and SVM Lin. even as it is fitting models from a far smaller hypothesis space (i.e. linear classifiers with 5 features, sign constraints and integer coefficients vs. linear classifiers with real coefficients).
\shortsection{On the Usability and Interpretability of Acceptable Models}
To discuss interpretability, we compare the best models that satisfied all operational constraints in Figure \ref{Fig::SleepApneaModels}, and present the SLIM model as a scoring system in Figure \ref{Fig::SleepApneaScoringSystem}.
\begin{figure}[htbp]
\centering
\renewcommand{\arraystretch}{1.05}
\begin{tabular}{llp{1mm}lp{1mm}lp{1mm}lp{1mm}lp{1mm}lp{1mm}}
SLIM 			& $4 ~age\geq60$ 	& $\scriptsize{+}$ & $4 ~{hypertension}$ & $\scriptsize{+}$ & $2 ~{bmi\geq30}$ & $\scriptsize{+}$ & $2 ~{bmi\geq40}$ & $\scriptsize{-}$ & $6 ~{female}$ & $\scriptsize{-}$ & $1$ \\ 
\midrule
Lasso 			& $0.13 ~{snoring}$ & $\scriptsize{+}$ & $0.12 ~{hypertension}$ & $\scriptsize{-}$ & $0.26 ~{female}$ & $\scriptsize{-}$ & $0.17$ 				&  						&  					   &  							&  \\ 
\midrule
Elastic Net 	& $0.03 ~{snoring}$ & $\scriptsize{+}$ & $0.02 ~{hypertension}$ & $\scriptsize{-}$ & $0.09 ~{female}$ & $\scriptsize{-}$ & $0.02$ 				&  						&  					   &  							&  \\ 
\end{tabular}
\caption{Score functions of the most sensitive predictive models that satisfied all three operational constraints. The baseline models have very poor sensitivity as shown in Table \ref{Table::DemoTableMaxFPR}.}
\label{Fig::SleepApneaModels}
\end{figure}
\begin{figure}[htbp]
\centering
\renewcommand{\arraystretch}{1.05}
\small{\textbf{PREDICT PATIENT HAS OBSTRUCTIVE SLEEP APNEA IF SCORE $> 1$}\vspace{0.25em}}\\
\begin{tabular}{|l l  r | c |}
   \hline
  1. & $age$ $\geq$ 60 & 4 points & $\quad\cdots\cdots$ \\ 
  2. & $hypertension$ & 4 points & $+\quad\cdots\cdots$ \\ 
  3. & $body~mass~index$ $\geq$ 30 & 2 points & $+\quad\cdots\cdots$ \\ 
  4. & $body~mass~index$ $\geq$ 40 & 2 points & $+\quad\cdots\cdots$ \\ 
  5. & $female$ & -6 points & $+\quad\cdots\cdots$ \\ 
   \hline
 & \small{\textbf{ADD POINTS FROM ROWS 1 -- 5}} & \small{\textbf{SCORE}} & $=\quad\cdots\cdots$ \\ 
   \hline
\end{tabular}
\caption{SLIM scoring system for sleep apnea screening. This model achieves a 10-CV mean test TPR/FPR of 61.4/20.9\%, obeys all operational constraints, and was trained without parameter tuning. It also generalizes well due to the simplicity of the hypothesis space: here the training TPR/FPR of the final model is 62.0/19.6\%.}
\label{Fig::SleepApneaScoringSystem}
\end{figure}
 
In this case, our collaborators found that all three models were aligned with domain knowledge as they obeyed sign constraints and had large coefficients for well-known risk factors such as $bmi$, $female$, $age$, $snoring$ and/or $hypertension$. Unfortunately, the Lasso and Elastic Net models could not be deployed as screening tools due to their poor sensitivity (29.3\% for Lasso and 44.2\% for Elastic Net). This was not the case for the SLIM model, which had a much higher sensitivity (61.4\%). 

Our results highlight some of the unique \textit{interpretability} benefits of SLIM scoring systems -- that is, their ability to provide ``a \textit{qualitative understanding} of the relationship between \textit{joint} values of the input variables and the resulting predicted response value" \citep{hastie2009elements}. SLIM scoring systems are well-suited to provide this kind of qualitative understanding due to their high level of sparsity and small integer coefficients. These qualities help users gauge the influence of each input variable with respect to the others, which is especially important because humans can only handle a few cognitive entities at once ($7\pm 2$ according to \citealt[][]{miller1984selection}), and are seriously limited in estimating the association between three or more variables \citep{jennings1982informal}. Sparsity and small integer coefficients also allow users to make quick predictions without a computer or a calculator, which may help them understand how the model works by actively using it to classify prototypical examples. Here, this process helped our collaborators come up with the following simple rule-based explanation for our model predicted that a patient has OSA (i.e., when SCORE $>$ 1): ``\textit{if the patient is male, predict OSA if age $\geq$ 60 OR hypertension OR bmi $\geq$ 30; if the patient is female, predict OSA if bmi $\geq$ 40 AND (age $\geq$ 60 OR hypertension).}"
\FloatBarrier
\section{Numerical Experiments}\label{Sec::NumericalExperiments}
In this section, we present numerical experiments to compare the accuracy and sparsity of SLIM scoring systems to other popular classification models. Our goal is to illustrate the off-the-shelf performance of SLIM and show that we can train accurate scoring systems for real-sized datasets in minutes.
\subsection{Experimental Setup}\label{Sec::ExperimentalSetup}
\textit{Datasets}: We ran numerical experiments on 8 datasets from the UCI Machine Learning Repository \citep{Bache+Lichman:2013} summarized in Table \ref{Table::ExperimentalDatasets}. We chose these datasets to explore the performance of each method as we varied the size and nature of the training data. We processed each dataset by binarizing all categorical features and some real-valued features. For the purposes of reproducibility, we include all processed datasets in Online Resource 1.
\begin{table}[htbp]
\scriptsize
\centering
\begin{tabular}{llrrl}
\toprule
\textbf{Dataset}  & \textbf{Source} & $N$ & $P$ & \textbf{Classification Task} \\
\toprule \texttt{adult} & \citet{kohavi1996scaling} & 32561 & 36 & predict if a U.S. resident earns more than $\$50\,000$ \\
\midrule \texttt{breastcancer} & \citet{mangasarian1995breast}& 683 & 9 & detect breast cancer using a biopsy \\ 
\midrule \texttt{bankruptcy} & \citet{kim2003discovery} & 250 & 6 & predict if a firm will go bankrupt \\ 
\midrule \texttt{haberman} & \citet{haberman1976generalized} & 306 & 3 & predict 5-year survival after breast cancer surgery \\ 
\midrule \texttt{heart} & \citet{detrano1989international} & 303 & 32 & identify patients a high risk of heart disease \\
\midrule \texttt{mammo} & \citet{elter2007prediction} & 961 & 12 & detect breast cancer using a mammogram \\
\midrule \texttt{mushroom} & \citet{schlimmer1987concept}& 8124 & 113 & predict if a mushroom is poisonous \\ 
\midrule \texttt{spambase} & \citet{cranor1998spam} & 4601 & 57 & predict if an e-mail is spam \\ 
\bottomrule
\end{tabular}
\caption{Datasets used in the numerical experiments.}
\label{Table::ExperimentalDatasets}
\end{table}
\FloatBarrier
\noindent \textit{Methods}: We summarize the training setup for each method in Table \ref{Table::ExperimentalMethods}. We trained SLIM scoring systems using slim\_for\_matlab toolbox paired with the CPLEX 12.6.0 API and models with baseline methods using publicly available packages in R 3.1.1 \citep{Rcitation}. For each method, each dataset, and each unique combination of free parameters, we trained 10 models using subsets of the data to estimate predictive accuracy via 10-fold cross-validation (10-CV), and 1 final model using all of the data to assess sparsity and interpretability. We ran all baseline methods without time constraints over a large grid of free parameters. We produced an $\ell_0$-regularization path for SLIM by solving $6 \times 11$ IPs for each dataset (6 values of $C_0$ $\times$ 11 training runs per $C_0$). We allocated at most 10 minutes to solve each IP, and solved 12 IPs in parallel on a 12-core 2.7 GHZ machine with 48 GB RAM. Thus, it took at most 1 hour to train SLIM scoring systems for each dataset. Since the \texttt{adult} and \texttt{haberman} datasets were imbalanced, we trained all methods on these datasets with a weighted loss function where we set $\wplus = \nminus/N$ and $\wminus = \nplus/N$.
\begin{table}[htbp]
\scriptsize
\centering
\begin{tabular}{@{ }l@{\hskip 3pt}l@{ }l@{\hskip 3pt}l@{ }}

\toprule 

\textbf{Method}&\textbf{Acronym}&\textbf{Software} & \textbf{Settings and Free Parameters}\\ 

\toprule

CART Decision Trees & CART & \pkg{rpart} \citep{the2012rpart} & default settings\\ 

\midrule

C5.0 Decision Trees & C5.0T & \pkg{c50} \citep{kuhn2012c50} & default settings\\ 

\midrule

C5.0 Rule List & C5.0R &  \pkg{c50} \citep{kuhn2012c50} & default settings\\ 

\midrule

Log. Reg. + $\ell_1$ penalty & Lasso & \pkg{glmnet} \citep{friedman2010glmnet} & 1000 values of $\lambda$ chosen by \pkg{glmnet}\\

\midrule

Log. Reg. + $\ell_2$ penalty & Ridge &  \pkg{glmnet} \citep{friedman2010glmnet} & 1000 values of $\lambda$ chosen by \pkg{glmnet}\\ 

\midrule

Log. Reg. + $\ell_1$/$\ell_2$ penalty & Elastic Net &  \pkg{glmnet} \citep{friedman2010glmnet} & 

\cell{l}{1000 values of $\lambda$ chosen by \pkg{glmnet}\\$\times$ 19 values of $\alpha \in \{0.05,0.10,\ldots,0.95\}$} \\

\midrule

SVM + Linear Kernel  & SVM Lin. &  \pkg{e1071} \citep{meyer2012e1071} & 25 values of $C \in \{10^{-3},10^{-2.75},\ldots,10^3\}$\\ 

\midrule

SVM + RBF Kernel & SVM RBF & \pkg{e1071} \citep{meyer2012e1071} & 25 values of $C \in \{10^{-3},10^{-2.75},\ldots,10^3\}$\\ 

\midrule

SLIM Scoring Systems & SLIM & \pkg{slim\_for\_matlab} \citep{ustun2015slimformatlab} & 

\cell{l}{
6 values of $C_0 \in \{0.01, 0.075, 0.05, 0.025, 0.001, 0.9/NP\}$\\
with $\lambda_j \in \{-10,\ldots,10\}$; $\lambda_0 \in \{-100,\ldots,100\}$ 
} \\ 
\bottomrule 
\end{tabular}
\caption{Training setup for classification methods used for the numerical experiments.}
\label{Table::ExperimentalMethods}
\end{table}
\subsection{Results and Observations}\label{Sec::NumericalExperimentsResults}
We summarize the results of our experiments in Table \ref{Table::ExpResultsTable} and Figures \ref{Fig::ExpPlots1}--\ref{Fig::ExpPlots2}. We report the sparsity of models using a metric that we call \textit{model size}. Model size represents the number of coefficients for linear models (Lasso, Ridge, Elastic Net, SLIM, SVM Lin.), the number of leaves for decision tree models (C5.0T, CART), and the number of rules for rule-based models (C5.0R). For completeness, we set the model size for black-box models (SVM RBF) to the number of features in each dataset.

We show the accuracy and sparsity of all methods on all dataset in Figures \ref{Fig::ExpPlots1}--\ref{Fig::ExpPlots2}. For each dataset, and each method, we plot a point at the 10-CV mean test error and final model size, and surround this point with an error bar whose height corresponds to the 10-CV standard deviation in test error. In addition, we include $\ell_0$-regularization paths for SLIM and Lasso on the right side of Figures \ref{Fig::ExpPlots1}--\ref{Fig::ExpPlots2} to show how the test error varies at different levels of sparsity for sparse linear models.
\begin{table}[htbp]
\scriptsize
\centering
\resizebox{\textwidth}{!} {
\setlength{\tabcolsep}{1pt}
\begin{tabular}{l@{   }l@{  }lccccccccc}
   \toprule \bf{Dataset} & 
   \bf{Details} & 
   \bf{Metric} & \bf{SLIM} & \bf{Lasso} & \bf{Ridge} & \bf{Elastic Net} & \bf{C5.0R} & \bf{C5.0T} & \bf{CART} & \bf{SVM Lin.} & \bf{SVM RBF} \\ 

   \midrule \texttt{adult} 
   & \ddcell{$N$&32561\\$P$&37\\Pr($y$=$+1$)&24\%\\Pr($y$=$-1$)&76\%} 
   & \bfcell{l}{test error\\train error\\model size\\model range}  
   & \cell{c}{17.4 $\pm$ 1.4$\%$\\17.5 $\pm$ 1.2$\%$\\18\\7 - 26} 
   & \cell{c}{17.3 $\pm$ 0.9$\%$\\17.2 $\pm$ 0.1$\%$\\14\\13 - 14} & \cell{c}{17.6 $\pm$ 0.9$\%$\\17.6 $\pm$ 0.1$\%$\\36\\36 - 36} & \cell{c}{17.4 $\pm$ 0.9$\%$\\17.4 $\pm$ 0.1$\%$\\17\\16 - 18} & \cell{c}{26.4 $\pm$ 1.8$\%$\\25.3 $\pm$ 0.4$\%$\\41\\38 - 46} & \cell{c}{26.3 $\pm$ 1.4$\%$\\24.9 $\pm$ 0.4$\%$\\87\\78 - 99} & \cell{c}{75.9 $\pm$ 0.0$\%$\\75.9 $\pm$ 0.0$\%$\\ 4\\4 - 4} & \cell{c}{16.8 $\pm$ 0.8$\%$\\16.7 $\pm$ 0.1$\%$\\36\\36 - 36} & \cell{c}{16.3 $\pm$ 0.5$\%$\\16.3 $\pm$ 0.1$\%$\\36\\36 - 36} \\ 

   \midrule \texttt{breastcancer} 
   & \ddcell{$N$&683\\$P$&10\\Pr($y$=$+1$)&35\%\\Pr($y$=$-1$)&65\%} 
   & \bfcell{l}{test error\\train error\\model size\\model range} 
   & \cell{c}{3.4 $\pm$ 2.0$\%$\\3.2 $\pm$ 0.2$\%$\\ 2\\2 - 2}
   & \cell{c}{3.4 $\pm$ 2.2$\%$\\2.9 $\pm$ 0.3$\%$\\ 9\\8 - 9} & \cell{c}{3.4 $\pm$ 2.0$\%$\\3.0 $\pm$ 0.3$\%$\\ 9\\9 - 9} & \cell{c}{3.1 $\pm$ 2.1$\%$\\2.8 $\pm$ 0.3$\%$\\ 9\\9 - 9} & \cell{c}{4.3 $\pm$ 3.3$\%$\\2.1 $\pm$ 0.3$\%$\\ 8\\6 - 9} & \cell{c}{5.3 $\pm$ 3.4$\%$\\1.6 $\pm$ 0.4$\%$\\13\\7 - 16} & \cell{c}{5.6 $\pm$ 1.9$\%$\\3.6 $\pm$ 0.3$\%$\\ 7\\3 - 7} & \cell{c}{3.1 $\pm$ 2.0$\%$\\2.7 $\pm$ 0.2$\%$\\ 9\\9 - 9} & \cell{c}{3.5 $\pm$ 2.5$\%$\\0.3 $\pm$ 0.1$\%$\\ 9\\9 - 9}  \\ 

   \midrule \texttt{bankruptcy} 
   & \ddcell{$N$&250\\$P$&7\\Pr($y$=$+1$)&57\%\\Pr($y$=$-1$)&43\%} 
   & \bfcell{l}{test error\\train error\\model size\\model range} 
   & \cell{c}{0.8 $\pm$ 1.7$\%$\\0.0 $\pm$ 0.0$\%$\\3\\2 - 3} 
   & \cell{c}{0.0 $\pm$ 0.0$\%$\\0.0 $\pm$ 0.0$\%$\\3\\3 - 3} & \cell{c}{0.4 $\pm$ 1.3$\%$\\0.4 $\pm$ 0.1$\%$\\6\\6 - 6} & \cell{c}{0.0 $\pm$ 0.0$\%$\\0.4 $\pm$ 0.7$\%$\\3\\3 - 3} & \cell{c}{0.8 $\pm$ 1.7$\%$\\0.4 $\pm$ 0.2$\%$\\4\\4 - 4} & \cell{c}{0.8 $\pm$ 1.7$\%$\\0.4 $\pm$ 0.2$\%$\\4\\4 - 4} & \cell{c}{1.6 $\pm$ 2.8$\%$\\1.6 $\pm$ 0.3$\%$\\2\\2 - 2} & \cell{c}{0.4 $\pm$ 1.3$\%$\\0.4 $\pm$ 0.1$\%$\\6\\6 - 6} & \cell{c}{0.4 $\pm$ 1.3$\%$\\0.4 $\pm$ 0.1$\%$\\6\\6 - 6} \\ 

   \midrule \texttt{haberman} 
   & \ddcell{$N$&306\\$P$&4\\Pr($y$=$+1$)&74\%\\Pr($y$=$-1$)&26\%} 
   & \bfcell{l}{test error\\train error\\model size\\model range} 
   & \cell{c}{29.2 $\pm$ 14.0$\%$\\29.7 $\pm$ \phantom{0}1.5$\%$\\3\\2 - 3}
   & \cell{c}{42.5 $\pm$ 11.3$\%$\\40.6 $\pm$ \phantom{0}1.9$\%$\\2\\2 - 2} & \cell{c}{36.9 $\pm$ 15.0$\%$\\41.0 $\pm$ \phantom{0}9.7$\%$\\3\\3 - 3} & \cell{c}{40.9 $\pm$ 14.0$\%$\\45.1 $\pm$ 12.0$\%$\\1\\1 - 1} & \cell{c}{42.7 $\pm$ 9.4$\%$\\40.4 $\pm$ 8.5$\%$\\3\\0 - 3} & \cell{c}{42.7 $\pm$ 9.4$\%$\\40.4 $\pm$ 8.5$\%$\\3\\1 - 3} & \cell{c}{43.1 $\pm$ 8.0$\%$\\34.3 $\pm$ 2.8$\%$\\9\\4 - 9} & \cell{c}{45.3 $\pm$ 14.7$\%$\\46.0 $\pm$ 3.6$\%$\\3\\3 - 3} & \cell{c}{47.5 $\pm$ 6.2$\%$\\5.4 $\pm$ 1.5$\%$\\4\\4 - 4}  \\ 

   \midrule \texttt{mammo} 
   & \ddcell{$N$&961\\$P$&15\\Pr($y$=$+1$)&46\%\\Pr($y$=$-1$)&54\%} 
   & \bfcell{l}{test error\\train error\\model size\\model range} 
   & \cell{c}{19.5 $\pm$ 3.0$\%$\\18.3 $\pm$ 0.3$\%$\\ 9\\9 - 11} 
   & \cell{c}{19.0 $\pm$ 3.1$\%$\\19.3 $\pm$ 0.3$\%$\\13\\12 - 13} & \cell{c}{19.2 $\pm$ 3.0$\%$\\19.2 $\pm$ 0.4$\%$\\14\\14 - 14} & \cell{c}{19.0 $\pm$ 3.1$\%$\\19.2 $\pm$ 0.3$\%$\\14\\13 - 14} & \cell{c}{20.5 $\pm$ 3.3$\%$\\19.8 $\pm$ 0.3$\%$\\ 5\\3 - 5} & \cell{c}{20.3 $\pm$ 3.5$\%$\\19.9 $\pm$ 0.3$\%$\\ 5\\4 - 6} & \cell{c}{20.7 $\pm$ 3.9$\%$\\20.0 $\pm$ 0.6$\%$\\ 5\\3 - 5} & \cell{c}{20.3 $\pm$ 3.0$\%$\\20.3 $\pm$ 0.4$\%$\\14\\14 - 14} & \cell{c}{19.1 $\pm$ 3.1$\%$\\18.2 $\pm$ 0.4$\%$\\14\\14 - 14} \\ 

   \midrule \texttt{heart} 
   & \ddcell{$N$&303\\$P$&33\\Pr($y$=$+1$)&46\%\\Pr($y$=$-1$)&54\%} 
   & \bfcell{l}{test error\\train error\\model size\\model range} 
   & \cell{c}{16.5 $\pm$ 7.8$\%$\\14.9 $\pm$ 1.1$\%$\\ 3\\3 - 3}
   & \cell{c}{15.2 $\pm$ 6.3$\%$\\14.0 $\pm$ 1.0$\%$\\12\\10 - 13} & \cell{c}{14.9 $\pm$ 5.9$\%$\\13.1 $\pm$ 0.8$\%$\\32\\30 - 32} & \cell{c}{14.5 $\pm$ 5.9$\%$\\13.2 $\pm$ 0.6$\%$\\24\\22 - 27} & \cell{c}{21.2 $\pm$ 7.5$\%$\\10.0 $\pm$ 1.8$\%$\\ 7\\9 - 17} & \cell{c}{23.2 $\pm$ 6.8$\%$\\8.5 $\pm$ 2.0$\%$\\16\\12 - 27} & \cell{c}{19.8 $\pm$ 6.5$\%$\\14.3 $\pm$ 0.9$\%$\\ 6\\6 - 8} & \cell{c}{15.5 $\pm$ 6.5$\%$\\13.6 $\pm$ 0.5$\%$\\31\\28 - 32} & \cell{c}{15.2 $\pm$ 6.0$\%$\\10.4 $\pm$ 0.8$\%$\\32\\32 - 32} \\ 

   \midrule \texttt{mushroom} 
   & \ddcell{$N$&8124\\$P$&114\\Pr($y$=$+1$)&48\%\\Pr($y$=$-1$)&52\%} 
   & \bfcell{l}{test error\\train error\\model size\\model range} 
   & \cell{c}{0.0 $\pm$ 0.0$\%$\\0.0 $\pm$ 0.0$\%$\\  7\\7 - 7}
   & \cell{c}{0.0 $\pm$ 0.0$\%$\\0.0 $\pm$ 0.0$\%$\\ 21\\19 - 23} & \cell{c}{1.7 $\pm$ 0.3$\%$\\1.7 $\pm$ 0.0$\%$\\113\\113 - 113} & \cell{c}{0.0 $\pm$ 0.0$\%$\\0.0 $\pm$ 0.0$\%$\\108\\106 - 108} & \cell{c}{0.0 $\pm$ 0.0$\%$\\0.0 $\pm$ 0.0$\%$\\  8\\8 - 8} & \cell{c}{0.0 $\pm$ 0.0$\%$\\0.0 $\pm$ 0.0$\%$\\  9\\9 - 9} & \cell{c}{1.2 $\pm$ 0.6$\%$\\1.1 $\pm$ 0.3$\%$\\  7\\6 - 8} & \cell{c}{0.0 $\pm$ 0.0$\%$\\0.0 $\pm$ 0.0$\%$\\ 98\\98 - 108} & \cell{c}{0.0 $\pm$ 0.0$\%$\\0.0 $\pm$ 0.0$\%$\\113\\113 - 113}  \\ 

   \midrule \texttt{spambase} 
   & \ddcell{$N$&4601\\$P$&58\\Pr($y$=$+1$)&39\%\\Pr($y$=$-1$)&61\%} 
   & \bfcell{l}{test error\\train error\\model size\\model range} 
   & \cell{c}{6.3 $\pm$ 1.2$\%$\\5.7 $\pm$ 0.3$\%$\\34\\28 - 40}
   & \cell{c}{10.0 $\pm$ 1.7$\%$\\9.5 $\pm$ 0.3$\%$\\28\\28 - 29} & \cell{c}{26.3 $\pm$ 1.7$\%$\\26.1 $\pm$ 0.2$\%$\\57\\57 - 57} & \cell{c}{10.0 $\pm$ 1.7$\%$\\9.6 $\pm$ 0.2$\%$\\28\\28 - 29} & \cell{c}{6.6 $\pm$ 1.3$\%$\\4.2 $\pm$ 0.3$\%$\\29\\23 - 31} & \cell{c}{7.3 $\pm$ 1.0$\%$\\3.9 $\pm$ 0.3$\%$\\73\\56 - 78} & \cell{c}{11.1 $\pm$ 1.4$\%$\\9.8 $\pm$ 0.3$\%$\\ 7\\6 - 10} & \cell{c}{7.8 $\pm$ 1.5$\%$\\8.1 $\pm$ 0.8$\%$\\57\\57 - 57} & \cell{c}{13.7 $\pm$ 1.4$\%$\\1.3 $\pm$ 0.1$\%$\\57\\57 - 57}  \\ 

   \bottomrule \end{tabular}
}
\caption{Accuracy and sparsity of all methods on all datasets. Here: test error refers to the 10-CV mean test error $\pm$ the 10-CV standard deviation in test error; train error refers to the 10-CV mean training error $\pm$ the 10-CV standard deviation in training error; model size refers to the final model size; and model range refers to the 10-CV minimum and maximum model size. The results reflect the models produced by each method when free parameters are chosen to minimize the 10-CV mean test error. We report the 10-CV weighted test and training error for \textds{adult} and \textds{haberman}.}
\label{Table::ExpResultsTable}
\end{table}
\FloatBarrier
\noindent We wish to make the following observations regarding our results: \vspace{-0.25em}
\shortsection{On the Accuracy, Sparsity and Computation}\vspace{-0.25em}
Our results show that many methods are unable to produce models that attain the same levels of accuracy and sparsity as SLIM. As shown in Figures \ref{Fig::ExpPlots1}--\ref{Fig::ExpPlots2}, SLIM always produces a model that is more accurate than Lasso at some level of sparsity, and sometimes more accurate at all levels of sparsity (e.g., \textds{spambase}, \textds{haberman}, \textds{mushroom}, \textds{breastcancer}). Although optimization problems to train SLIM scoring systems were $\mathcal{NP}$-hard, we did not find any evidence that computational issues hurt the performance of SLIM on any of the datasets. We obtained accurate and sparse models for all datasets in 10 minutes using CPLEX 12.6. Further, the solver provided a proof of optimality (i.e., a MIPGAP of 0.0\%) for all models we trained for \textds{mammo}, \textds{mushroom}, \textds{bankruptcy}, \textds{breastcancer}. We attribute these benefits to SLIM's tighter MIP formulation (see Section \ref{Sec::SLIMIPFormulation}).
\shortsection{On the Regularization Effect of Discrete Coefficients}\vspace{-0.25em}
We expect that methods that directly optimize accuracy and sparsity will achieve the best possible accuracy at every level of sparsity (i.e. the best possible trade-off between accuracy and sparsity). SLIM directly optimizes accuracy and sparsity. However, it may not necessarily achieve the best possible accuracy at each level of sparsity because it restricts coefficients to a finite discrete set $\Lset$. 

By comparing SLIM to Lasso, we can identify a baseline regularization effect due to this $\Lset$ set restriction. In particular, we know that when Lasso's performance dominates that of SLIM, it is very arguably due to the use of a small set of discrete coefficients. Our results show that this tends to happen mainly at large model sizes (see e.g., the regularization path for \textds{breastcancer}, \textds{heart}, \textds{mammo}). This suggests that the $\Lset$ set restriction has a more noticeable impact on accuracy at larger model sizes.

One interesting effect of the $\Lset$ set restriction is that the most accurate SLIM scoring system may not use all of the features in the dataset. In our experiments, we always trained SLIM with $C_0 = 0.9/NP$ to obtain a scoring system with the highest training accuracy among linear models with coefficients in $\lambdab \in \Lset$ (see Remark \ref{Rem::TradeOffParamaterMin}). In the \textds{bankruptcy} dataset, for example, we find that this model only uses 3 out of 6 features. This is due to the $\Lset$ set restriction: if the $\Lset$ restriction were relaxed, then the method would use all features to improve its training accuracy (as is the case with Ridge or SVM Lin.).
\shortsection{On the Interpretability of Models}\vspace{-0.25em}
To discuss interpretability, we focus on the \textds{mushroom} dataset, which provides a nice basis for comparison as many methods produce a model that attains perfect predictive accuracy. In Figures \ref{Fig::MushroomSLIM}--\ref{Fig::MushroomC50R}, we show the sparsest models that achieve perfect predictive accuracy. We omit models from some methods because they do not attain perfect accuracy (CART), or use far more features (Ridge, SVM Lin, SVM RBF).

Here, the SLIM scoring system uses 7 integer coefficients. However, it can be simplified into a 5 line scoring system since \textit{odor=none}, \textit{odor=almond}, and \textit{odor=anise} are mutually exclusive variables with the same coefficient. The model lets users make predictions by hand, and uses a linear form that helps users gauge the influence of each input variable with respect to the others. Note that only some of these qualities are found in the other models. The Lasso model, for instance, has a linear form but uses far more features. In contrast, the C5.0 models let users to make predictions by hand, but have a hierarchical structure that makes it difficult to gauge the influence of each input variable with respect to the others. 

We note that these qualities represent ``baseline" interpretability benefits. In practice, interpretability is a subjective and multifaceted notion (i.e., it depends on who will be using the model, and on many model qualities, as discussed in \cite{kodratoff1994comprehensibility,pazzani2000knowledge,Freitas:2014ic}). In light of this, SLIM has a additional interpretability benefit because it allows practitioners to work closely with their target audience and encode all interpretability-related requirements into their model by means of operational constraints.
\FloatBarrier

\begin{figure}[htbp]
\centering
\renewcommand{\arraystretch}{1.05}
\small{\textbf{PREDICT MUSHROOM IS POISONOUS IF SCORE $> 3$}} \\ 
\vspace{0.25em}
\begin{tabular}{|l l  l | r |}
   \hline
1.   & $spore\_print\_color = green$ & 4 points & $\phantom{+}\quad\cdots\cdots$ \\ 
  2. & $stalk\_surface\_above\_ring = grooves$  & 2 points & $+\quad\cdots\cdots$ \\ 
  3. & $population = clustered$ & 2 points & $+\quad\cdots\cdots$ \\ 
  4. & $gill\_size = broad$ & -2 points & $+\quad\cdots\cdots$ \\ 
  5. & $odor \in \{none, almond, anise\}$ & -4 points & $+\quad\cdots\cdots$ \\ 
   \hline
 & \small{\textbf{ADD POINTS FROM ROWS 1--5}} & \small{\textbf{SCORE}} & $=\quad\cdots\cdots$ \\ 
   \hline
\end{tabular}
\caption{SLIM scoring system for \texttt{mushroom}. This model has a 10-CV mean test error of $0.0\pm0.0\%$.} 
\label{Fig::MushroomSLIM}
\end{figure}

\begin{figure}[htbp]
\centering{
\scriptsize{
\begin{tabularx}{\textwidth}{p{1mm}lp{1mm}lp{1mm}l}
 & $10.86 ~\textit{spore\_print\_color = green}$ & $\scriptsize{+}$ & $4.49 ~\textit{gill\_size = narrow}$ & $\scriptsize{+}$ & $4.29 ~\textit{odor = foul}$ \\ 
 $\scriptsize{+}$ & $2.73 ~\textit{stalk\_surface\_below\_ring = scaly}$ & $\scriptsize{+}$ & $2.60 ~\textit{stalk\_surface\_above\_ring = grooves}$ & $\scriptsize{+}$ & $2.38 ~\textit{population = clustered}$ \\ 
$\scriptsize{+}$ & $0.85 ~\textit{spore\_print\_color = white}$ & $\scriptsize{+}$ & $0.44 ~\textit{stalk\_root = bulbous}$ & $\scriptsize{+}$ & $0.43 ~\textit{gill\_spacing = close}$ \\ 
$\scriptsize{+}$ & $0.38 ~\textit{cap\_color = white}$ & $\scriptsize{+}$ & $0.01 ~\textit{stalk\_color\_below\_ring = yellow}$ & $\scriptsize{-}$ & $8.61 ~\textit{odor = anise}$ \\ 
 $\scriptsize{-}$ & $8.61 ~\textit{odor = almond}$ & $\scriptsize{-}$ & $8.51 ~\textit{odor = none}$ & $\scriptsize{-}$ & $0.53 ~\textit{cap\_surface = fibrous}$ \\ 
 $\scriptsize{-}$ & $0.25 ~\textit{population = solitary}$ & $\scriptsize{-}$ & $0.21 ~\textit{stalk\_surface\_below\_ring = fibrous}$ & $\scriptsize{-}$ & $0.09 ~\textit{spore\_print\_color = brown}$ \\ 
 $\scriptsize{-}$ & $0.00 ~\textit{cap\_shape = convex}$ & $\scriptsize{-}$ & $0.00 ~\textit{gill\_spacing = crowded}$ & $\scriptsize{-}$ & $0.00 ~\textit{gill\_size = broad}$ \\ 
 $\scriptsize{+}$ & $0.25$ &  &  &  &  \\ 
  \end{tabularx}  
  \caption{Lasso score function for \texttt{mushroom}. This model has a 10-CV mean test error of 0.0 $\pm$ 0.0$\%$.}
}
}
\label{Fig::MushroomLasso}
\end{figure}
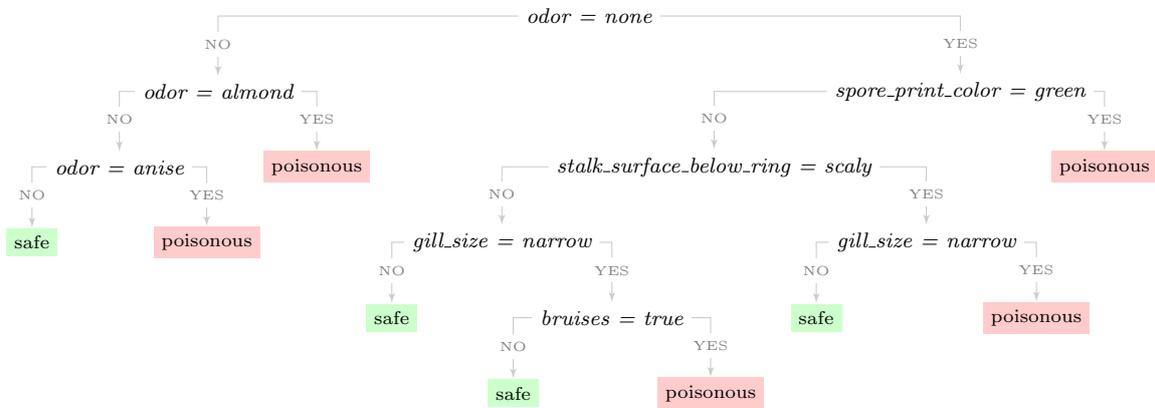
\begin{figure}[htbp]
\label{Fig::MushroomC50T}
\centering
\tikzstyle{block} = [fill=white!20, text centered, minimum height=1em,font=\it]
\tikzstyle{line} = [draw, color=gray!40, -latex']
\tikzstyle{lnode} = [below=-5.5pt, near end, color=gray, fill=white!20, text centered, font=\tiny]
\tikzstyle{rcloud} = [fill=red!20, text centered,font=\scriptsize]
\tikzstyle{gcloud} = [fill=green!20, text centered,font=\scriptsize]
    
    \begin{tikzpicture}

    \footnotesize
    \node [block] (root) {odor = none};
    \node [block, left of=root, left=13em,below of=root] (D1) {odor = almond};
    \node [rcloud, right of=D1, right=1em, below of=D1] (end1yes) {poisonous};
    \node [block, left of=D1, left=1em, below of=D1] (D2) {odor = anise};
    \node [rcloud, right of=D1, right=0.5em, below of=D2] (end12yes) {poisonous};
    \node [gcloud, left of=D1,  left=0.5em, below of=D2] (end12no) {safe};
    \node [block, right of=root, right=13em,below of=root] (D3) {spore\_print\_color = green};
    \node [rcloud,right of=D3, right=3em, below of=D3] (end3yes) {poisonous};
    \node [block, left of=D3, left=7.5em, below of=D3] (D4) {stalk\_surface\_below\_ring = scaly};
    \node [block, right of=D4, right=6em, below of=D4] (D5) {gill\_size = narrow};
    \node [rcloud, right of=D5, right=1.5em, below of=D5] (end5yes) {poisonous};
    \node [gcloud, left of=D5, left=1.5em, below of=D5] (end5no) {safe};
    \node [block, left of=D4, left=6em, below of=D4] (D6) {gill\_size = narrow};
    \node [gcloud, left of=D6, left=1.5em, below of=D6] (end6no) {safe};
    \node [block, right of=D6, right=1.5em, below of=D6] (D7) {bruises = true};
    \node [rcloud, right of=D7, right=1em, below of=D7] (end7yes) {poisonous};
    \node [gcloud, left of=D7, left=1em, below of=D7] (end7no) {safe};

    \path [line] (root) -| node [lnode] {NO} (D1);
    \path [line] (D1) -| node [lnode] {YES} (end1yes);
    \path [line] (D1) -| node [lnode] {NO} (D2);
    \path [line] (D2) -| node [lnode] {YES} (end12yes);
    \path [line] (D2) -| node [lnode] {NO} (end12no);
    \path [line] (root) -| node [lnode] {YES} (D3);
    \path [line] (D3) -| node [lnode] {YES} (end3yes);
    \path [line] (D3) -| node [lnode] {NO} (D4);
    \path [line] (D4) -| node [lnode] {YES} (D5);
    \path [line] (D5) -| node [lnode] {YES} (end5yes);
    \path [line] (D5) -| node [lnode] {NO} (end5no);
    \path [line] (D4) -| node [lnode] {NO} (D6);
    \path [line] (D6) -| node [lnode] {NO} (end6no);
    \path [line] (D6) -| node [lnode] {YES} (D7);
    \path [line] (D7) -| node [lnode] {YES} (end7yes);
    \path [line] (D7) -| node [lnode] {NO} (end7no);
   
\end{tikzpicture}
%
%
\caption{C5.0 decision tree for \textds{mushroom}. This model has a 10-CV mean test error of 0.0$\pm$0.0$\%$.}
\end{figure}
\begin{figure}
%
%
%
%
%
%
%
\scriptsize
\centering
\renewcommand{\arraystretch}{1.05}
\begin{tabular}{@{ }llccc}
\toprule 
\bfcell{c}{Rule} &  & \textbf{Confidence} &  \textbf{Support} & \textbf{Lift} \\ 
\toprule
$odor = none ~\wedge~  gill\_size \neq narrow ~\wedge~ spore\_print\_color \neq green$ & $\implies$ safe & 1.000 & 3216 & 1.9 \\
\midrule
$bruises = false ~\wedge~ odor = none ~\wedge~ stalk\_surface\_below\_ring \neq scaly$ & $\implies$ safe & 0.999 & 1440 & 1.9 \\
\midrule
$odor = almond$ & $\implies$ safe& 0.998 & 400 & 1.9 \\
\midrule
$odor = anise$ & $\implies$ safe & 0.998 & 400 & 1.9 \\
\midrule
$odor \neq almond ~\wedge~ odor \neq anise ~\wedge~ odor \neq none$ & $\implies$ poisonous & 1.000 & 3796 & 2.1 \\
\midrule
$spore\_print\_color = green$ & $\implies$ poisonous & 0.986 & 72 & 2.9 \\
\midrule
$gill\_size = narrow ~\wedge~ stalk\_surface\_below\_ring = scaly$ & $\implies$ poisonous & 0.976 & 40 & 2.0 \\
\bottomrule
\end{tabular}
\caption{C5.0 rule list for \textds{mushroom}. This model has a 10-CV mean test error of 0.0 $\pm$ 0.0$\%$.}
\label{Fig::MushroomC50R}
\end{figure}
%
%
\setlength{\fwidth}{0.425\textwidth}
\begin{figure}[htbp]
\centering
\begin{tabular}{c}
\hspace{0.5in} \includegraphics[trim=1in 1in 1in 1in,clip,scale=0.5]{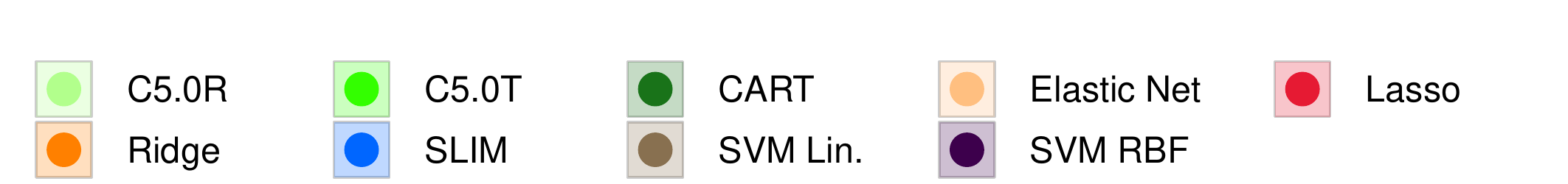} 
\end{tabular}
\begin{tabular}{>{\tiny}m{1cm}c} 
\tabdataname{adult}              & \begin{tabular}{lr}\includegraphics[width=\fwidth]{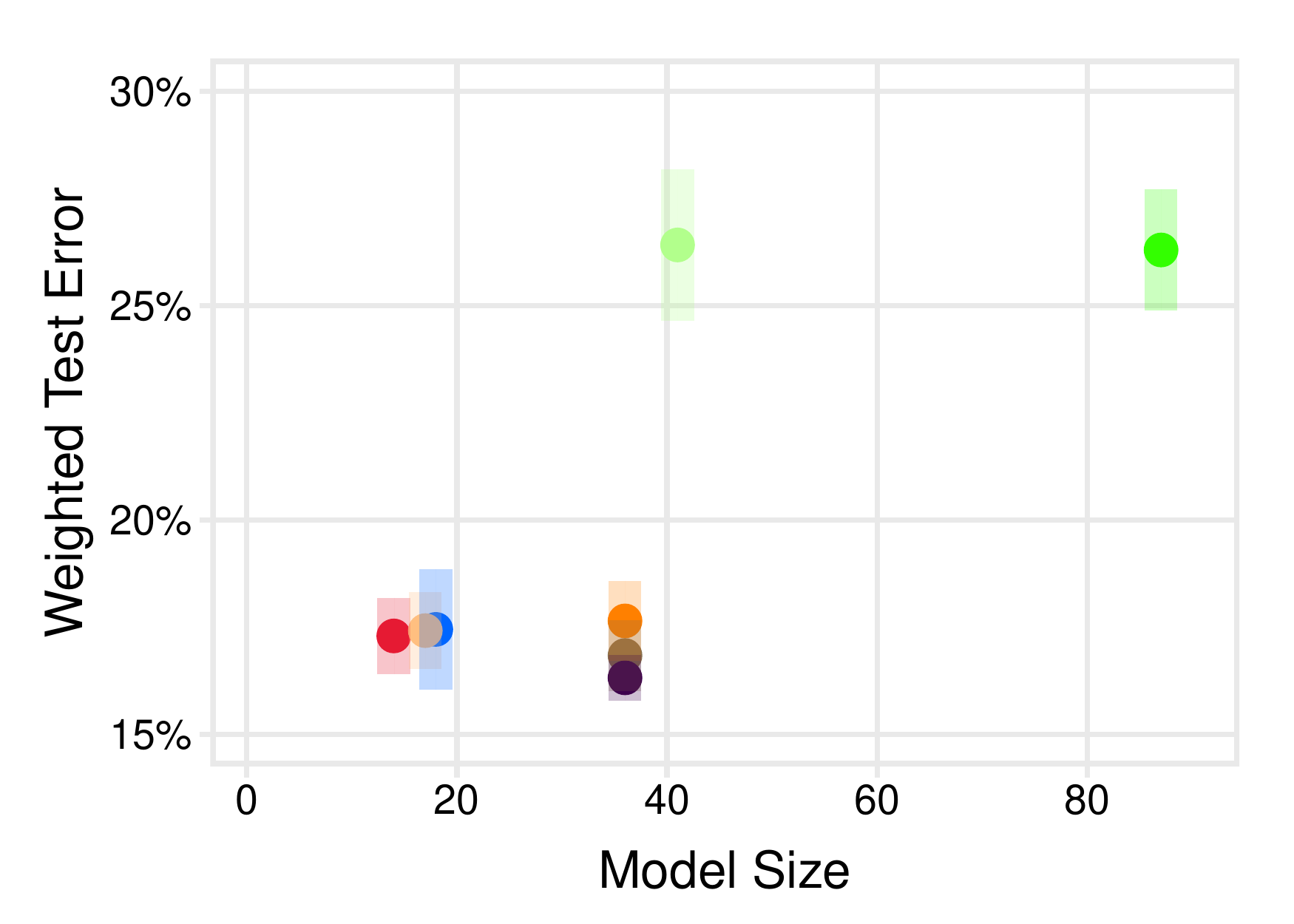}  & \includegraphics[width=\fwidth]{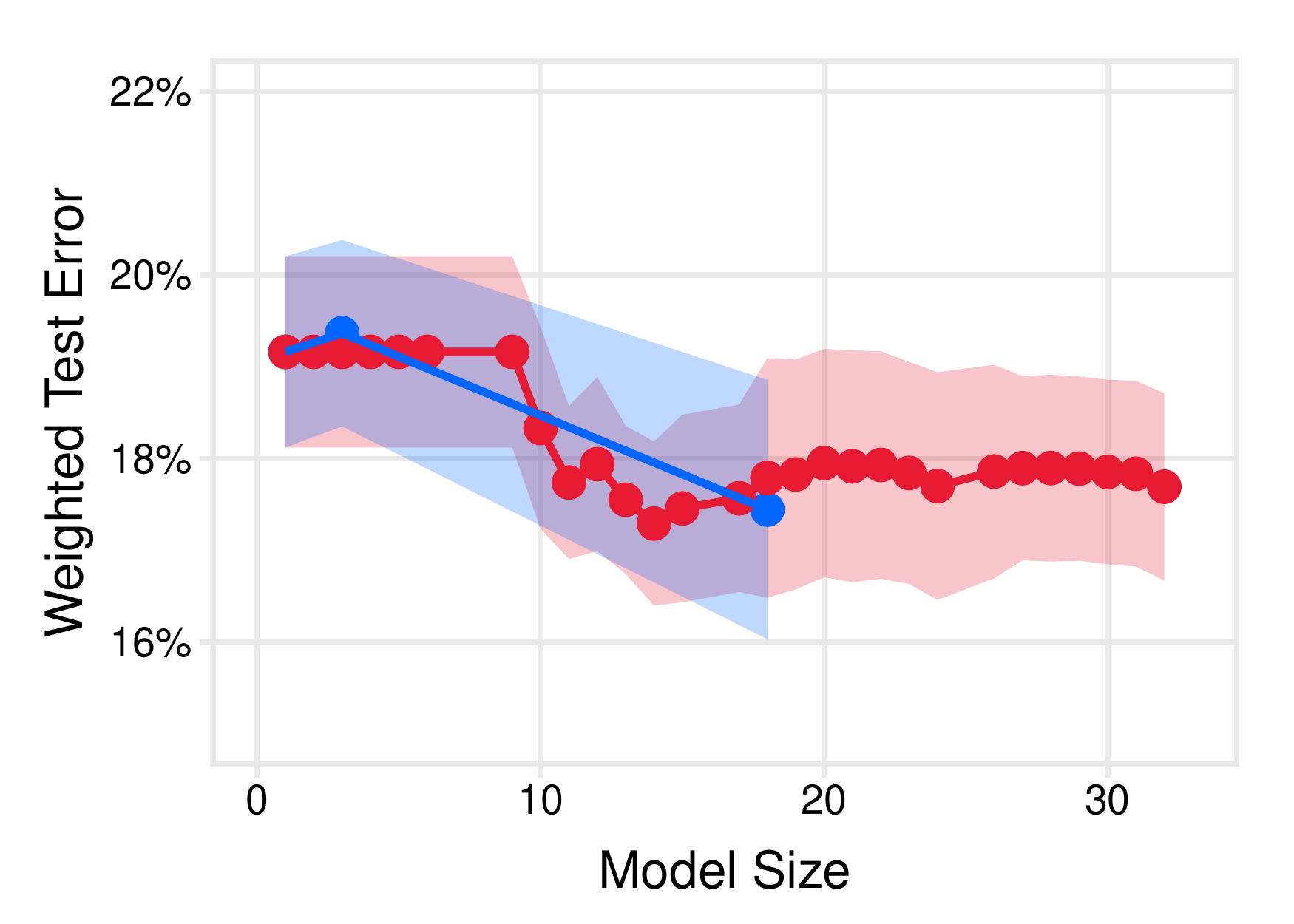} \end{tabular} \\ 
\tabdataname{breastcancer}       & \begin{tabular}{lr}\includegraphics[width=\fwidth]{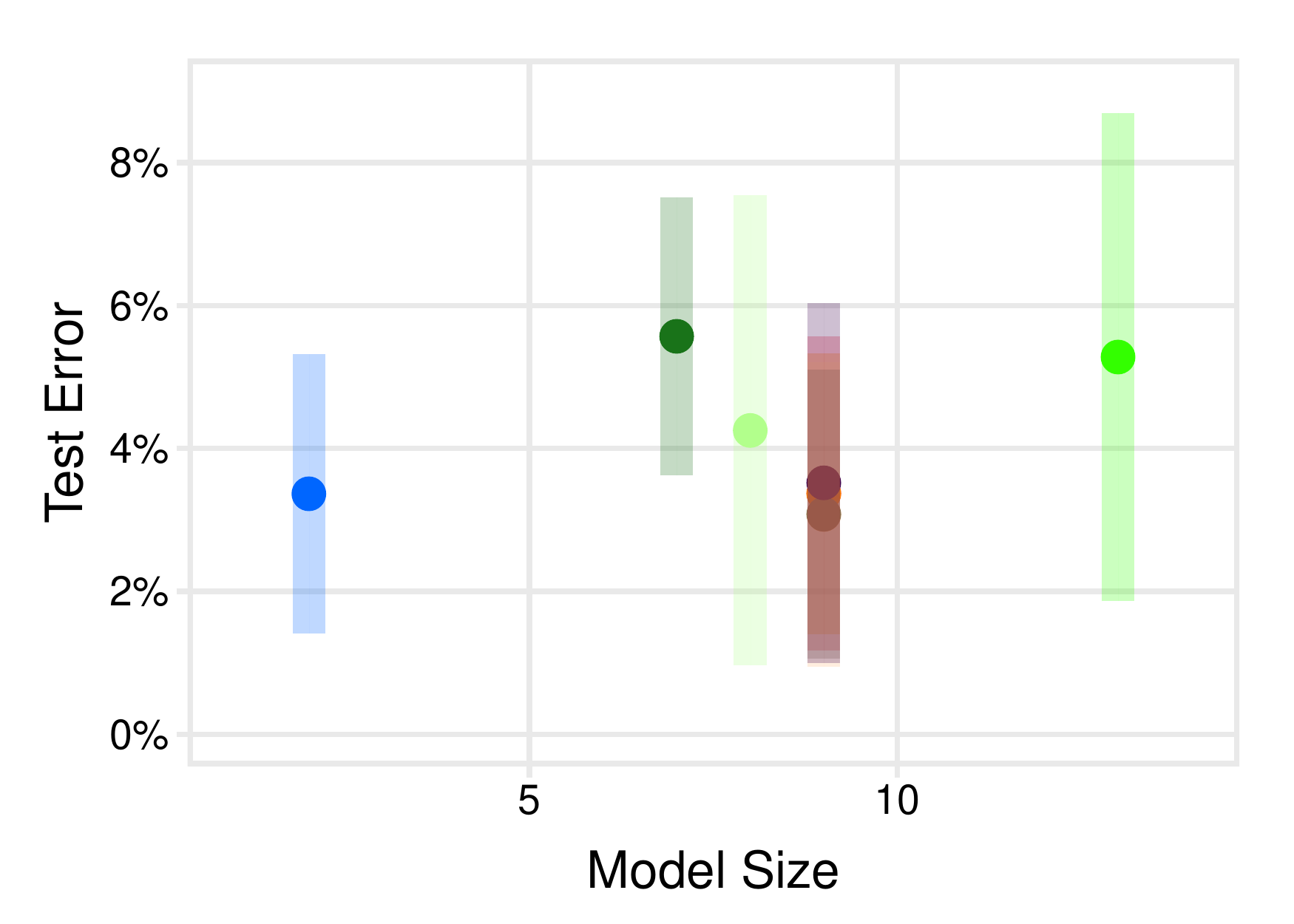} & \includegraphics[width=\fwidth]{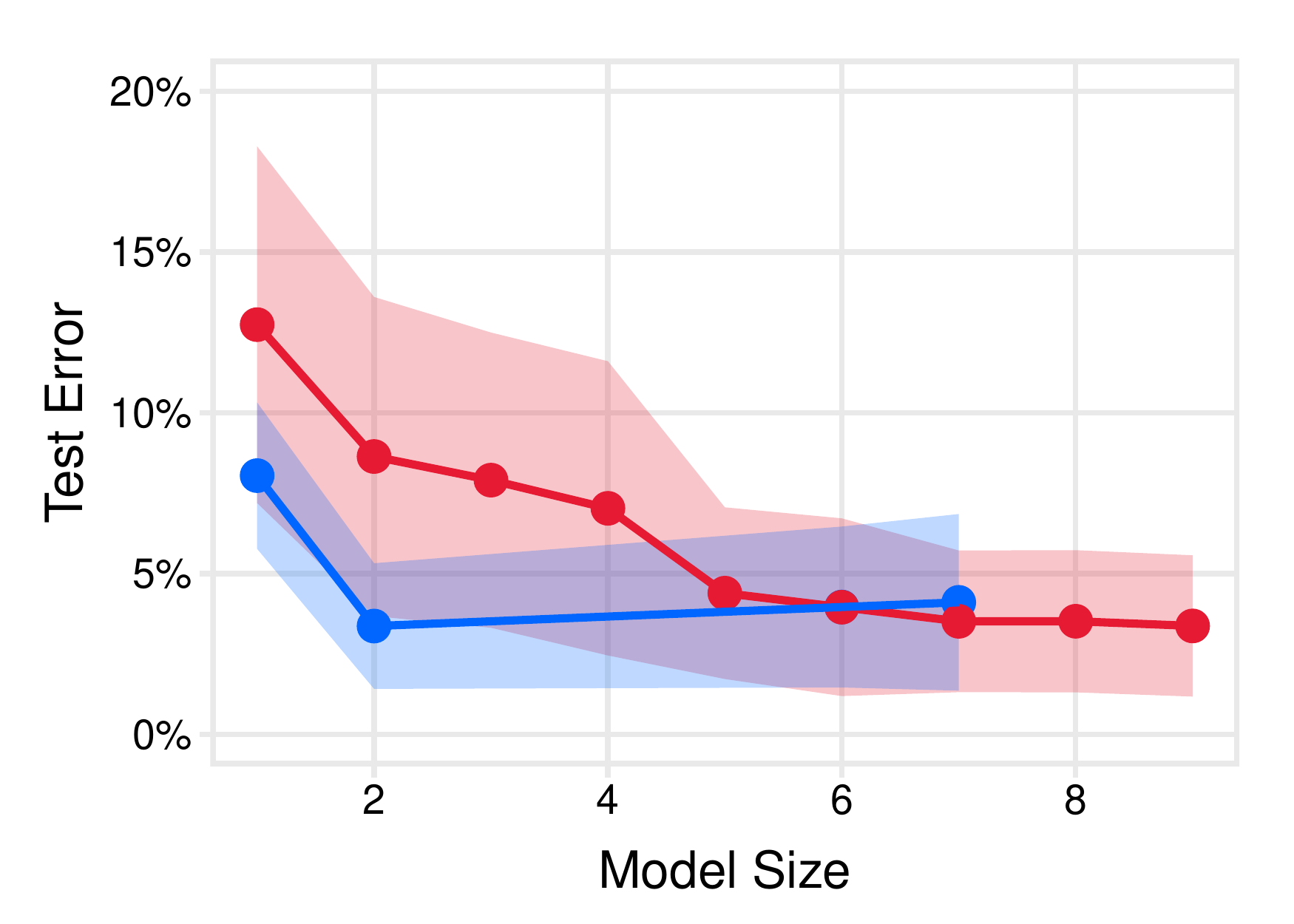} \end{tabular} \\ 
\tabdataname{bankruptcy}         & \begin{tabular}{lr}\includegraphics[width=\fwidth]{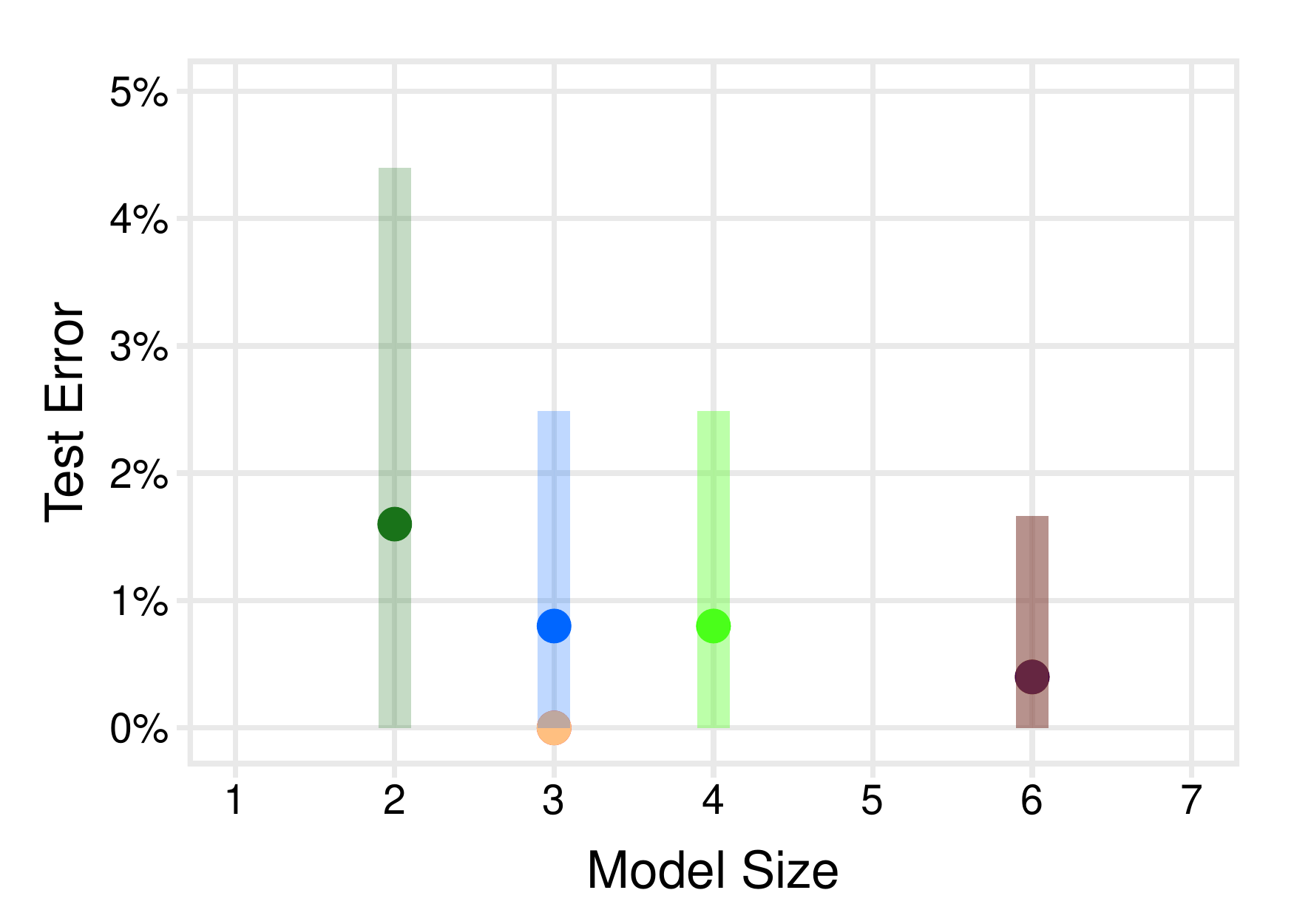}  & \includegraphics[width=\fwidth]{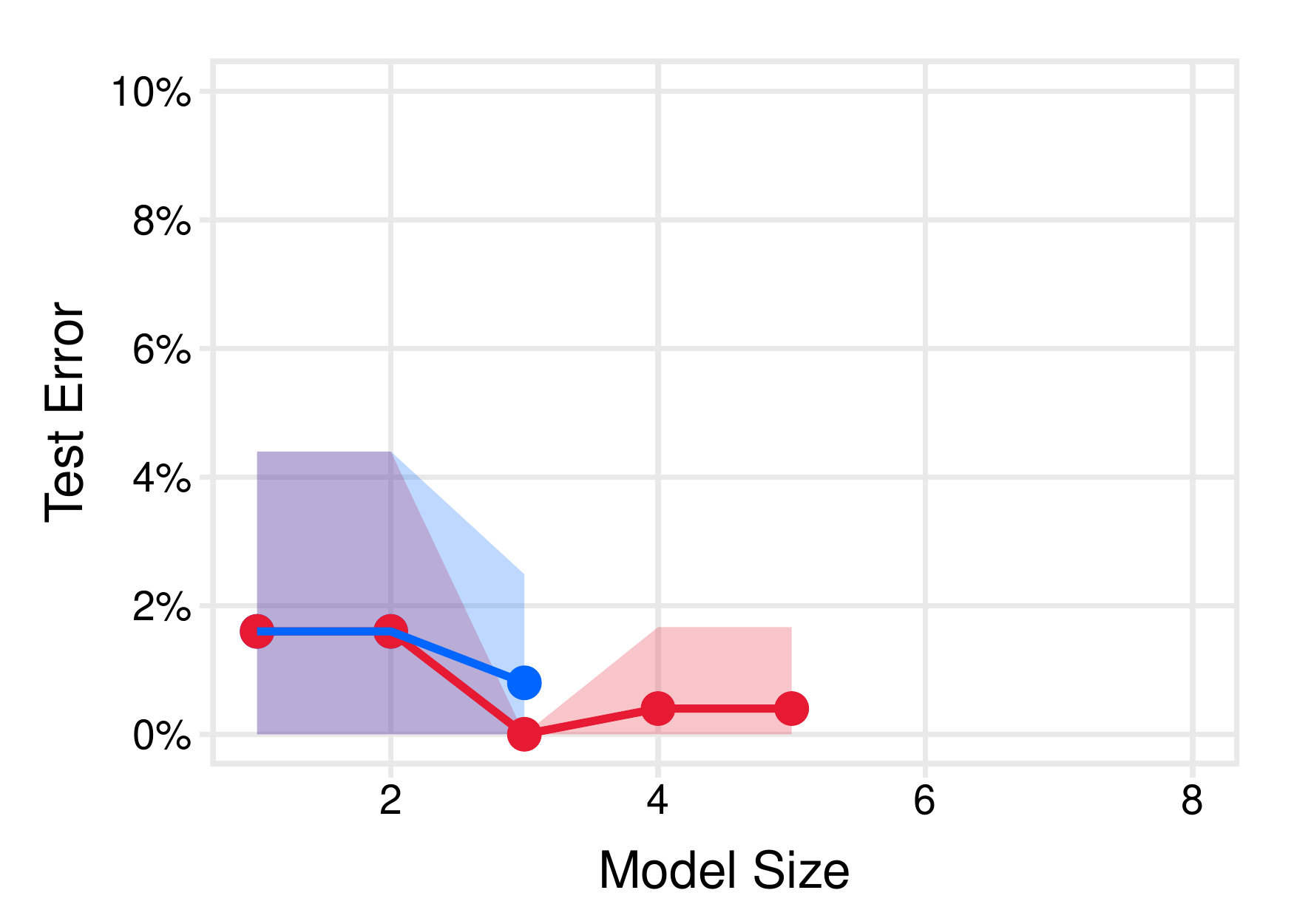} \end{tabular} \\ 
\tabdataname{haberman}           & \begin{tabular}{lr}\includegraphics[width=\fwidth]{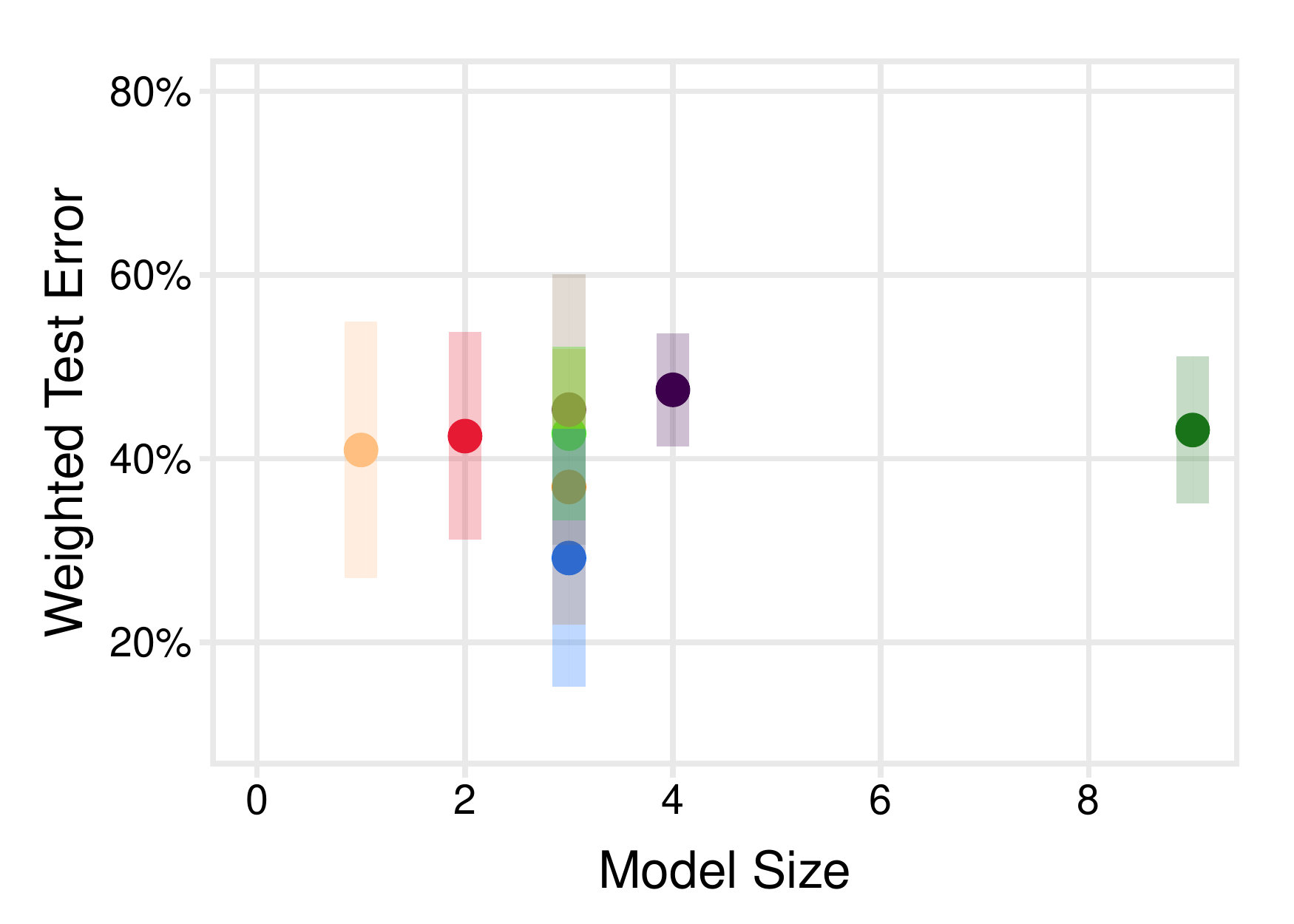}  & \includegraphics[width=\fwidth]{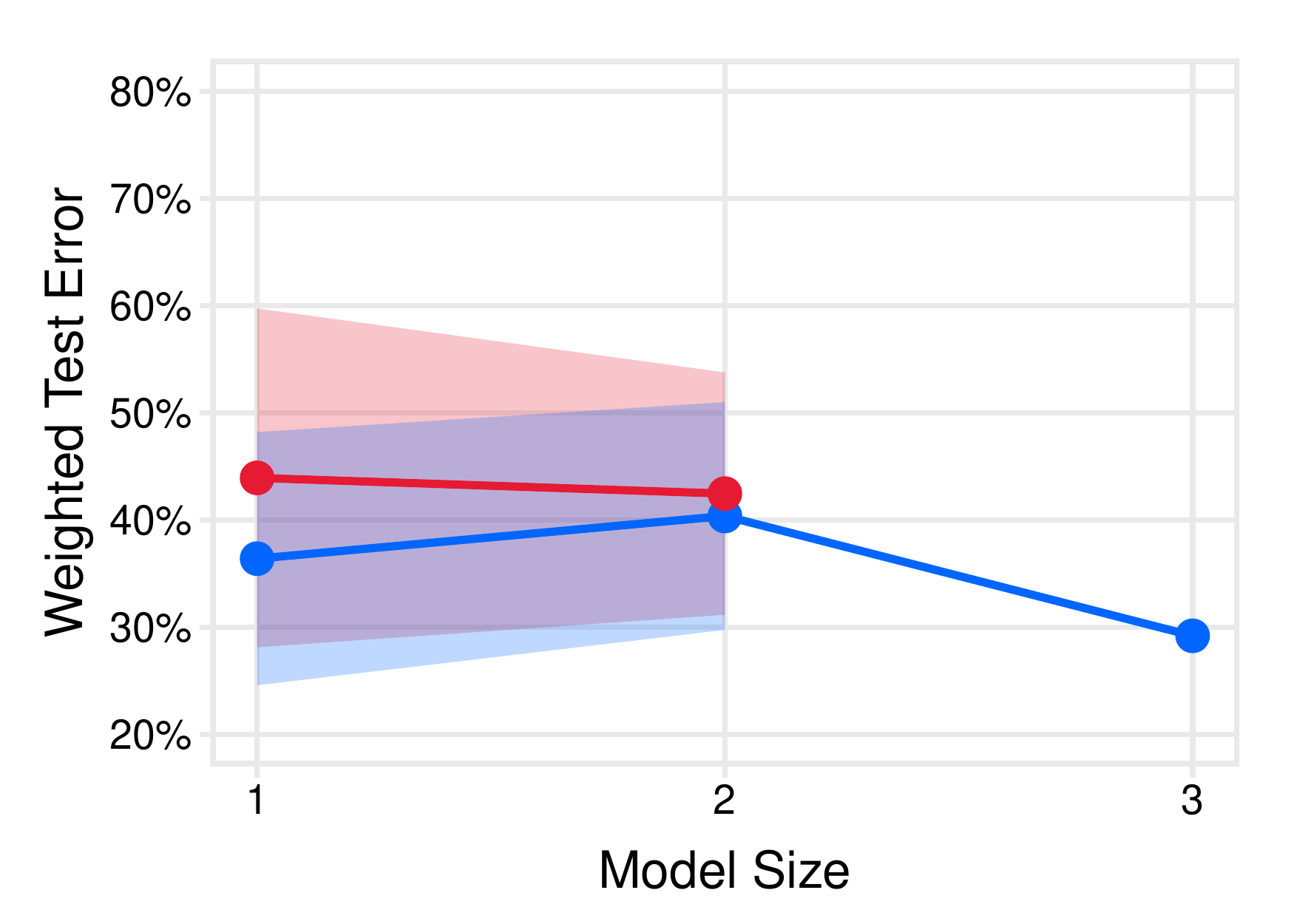} \end{tabular}
\end{tabular}
\caption{Accuracy and sparsity of all classification methods on all datasets. For each dataset, we plot the performance of models when free parameters are set to values that minimize the 10-CV mean test error (left), and plot the performance of SLIM and Lasso across the full $\lzero$-regularization path (right).}
\label{Fig::ExpPlots1}
\end{figure}
\begin{figure}[htbp]
\centering
\begin{tabular}{c}
\hspace{0.5in} \includegraphics[trim=1in 1in 1in 1in,clip,scale=0.5]{plot_legend} 
\end{tabular}
\begin{tabular}{>{\tiny}m{1cm}c} 
\tabdataname{mammo}     & \begin{tabular}{lr}\includegraphics[width=\fwidth]{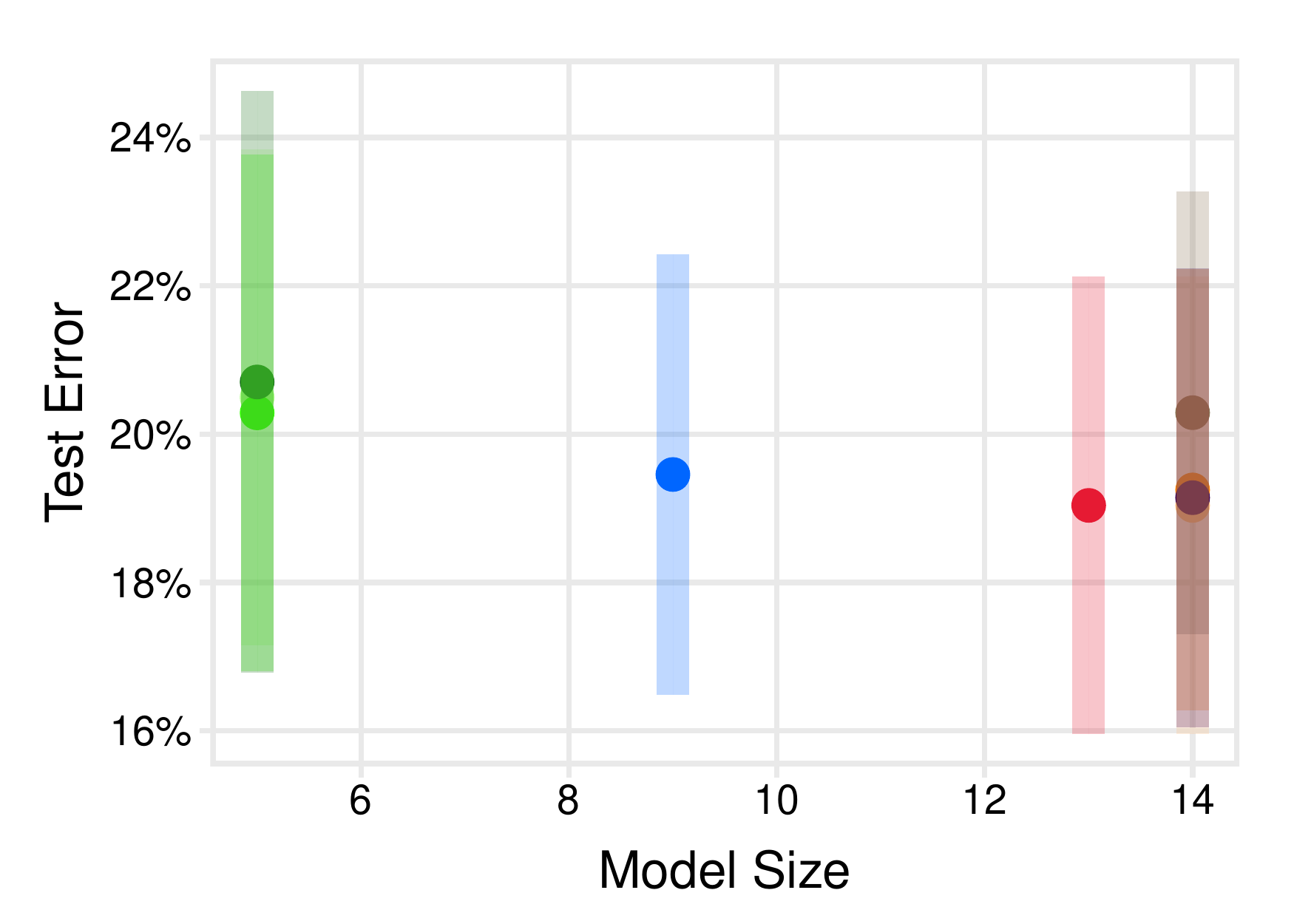}  & \includegraphics[width=\fwidth]{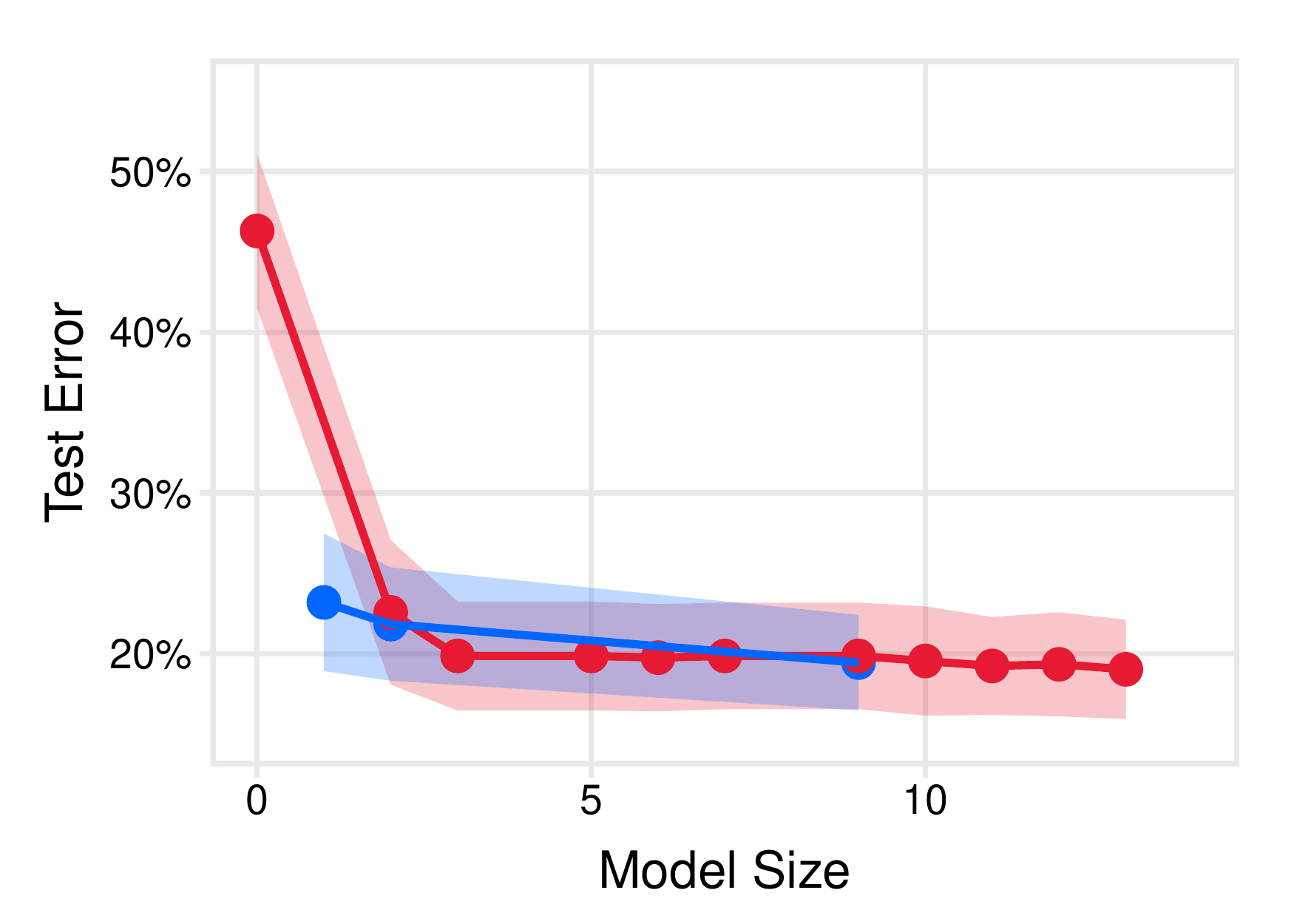} \end{tabular} \\ 
\tabdataname{heart}     & \begin{tabular}{lr}\includegraphics[width=\fwidth]{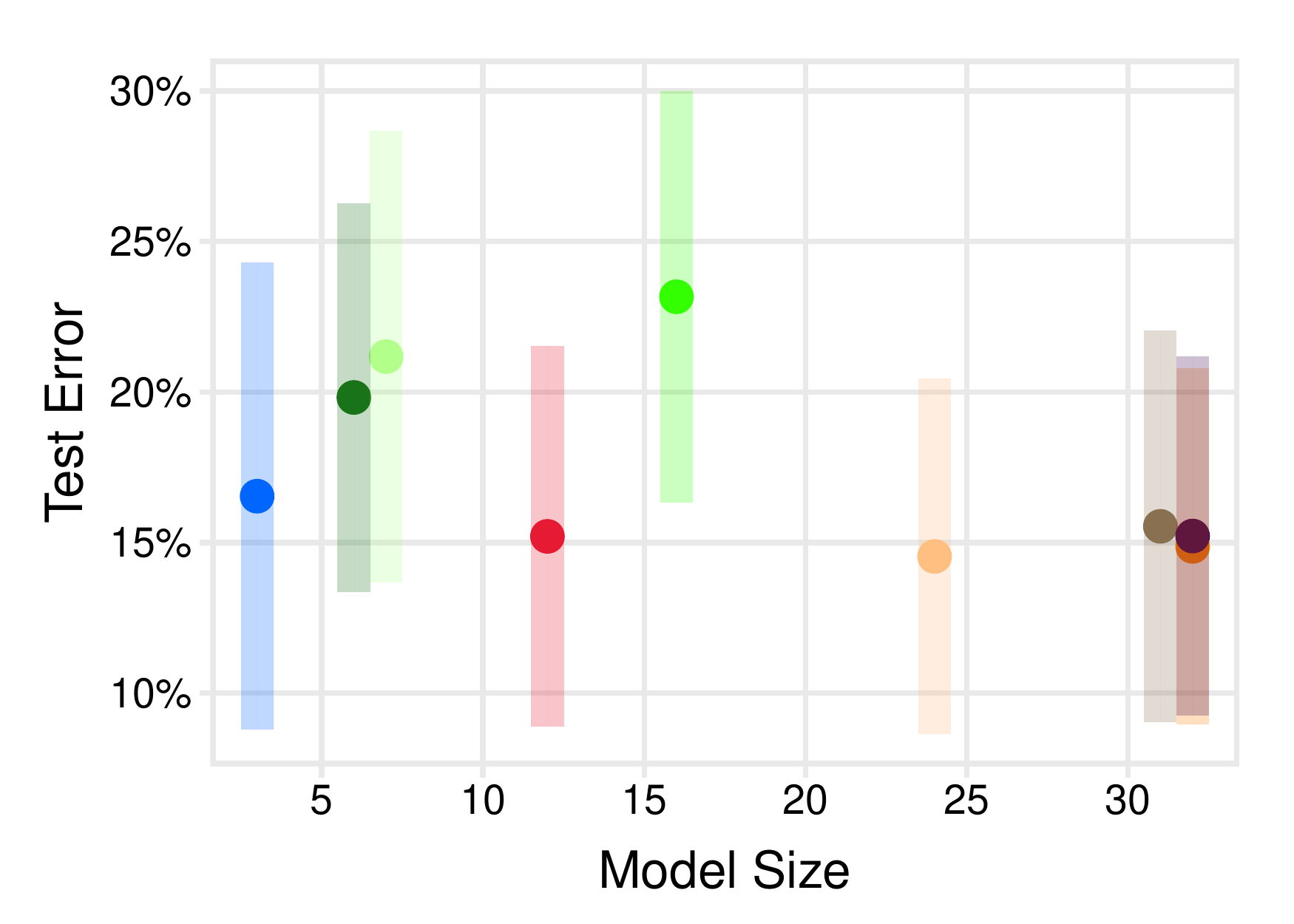}  & \includegraphics[width=\fwidth]{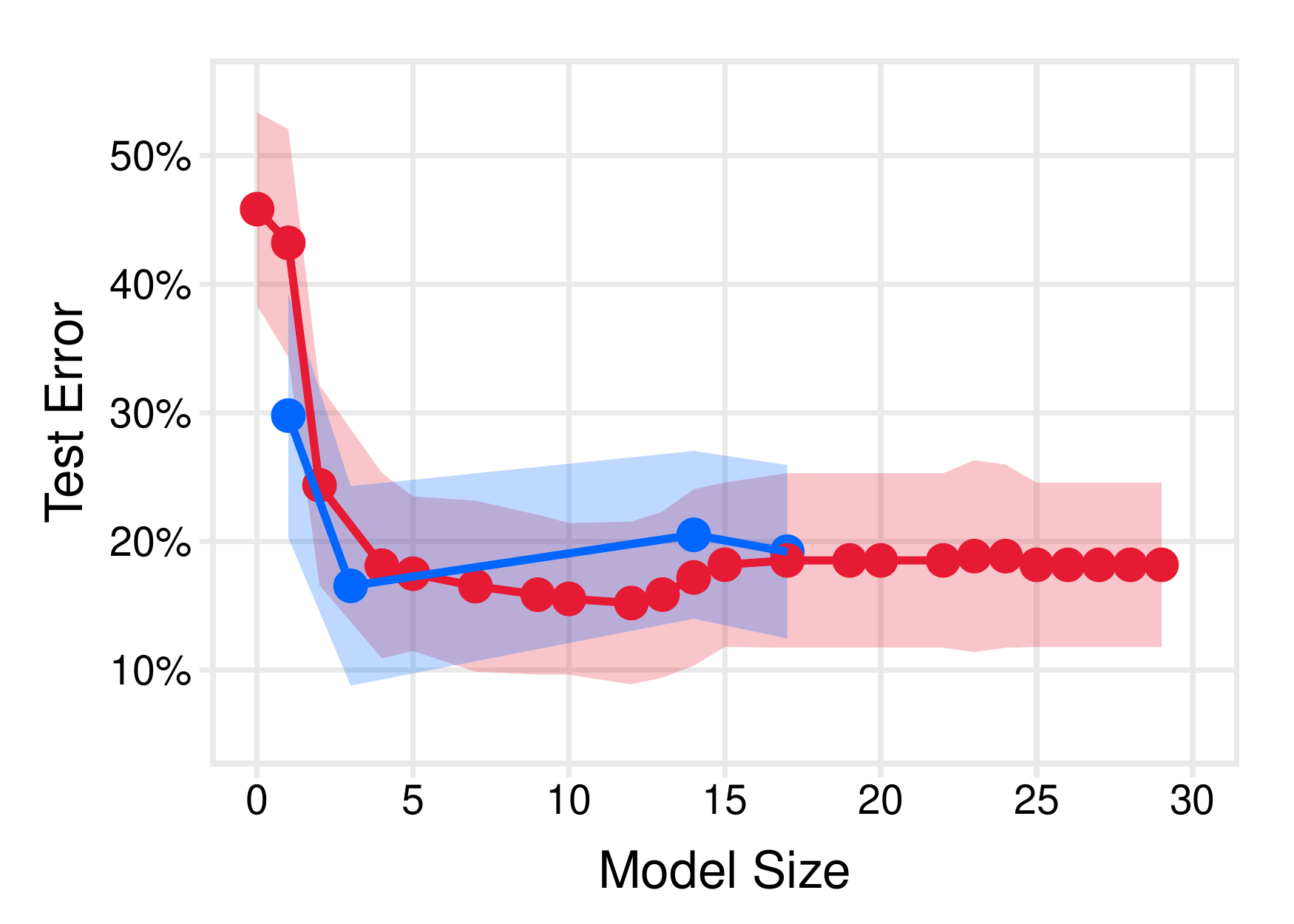} \end{tabular} \\ 
\tabdataname{mushroom}  & \begin{tabular}{lr}\includegraphics[width=\fwidth]{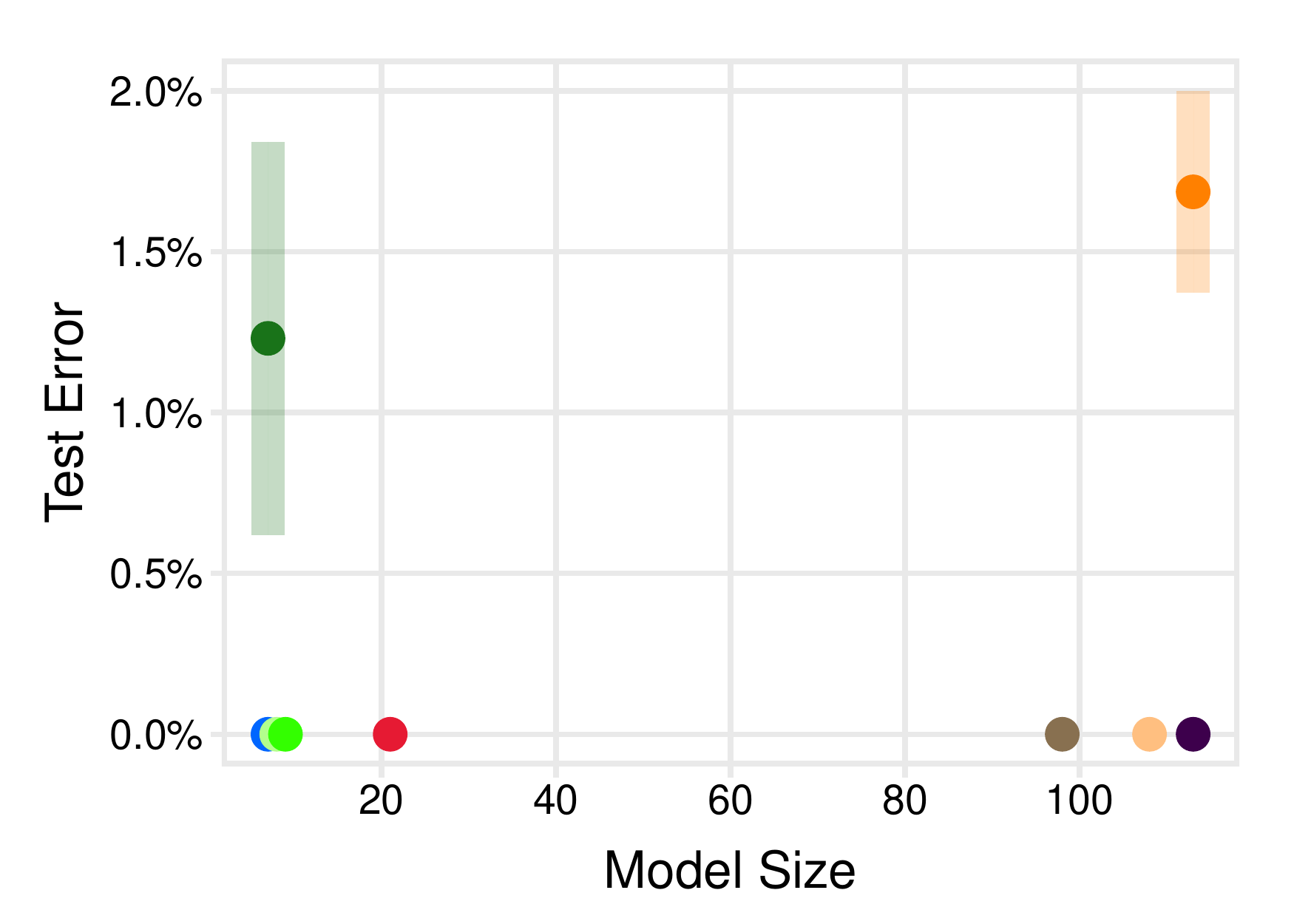}  & \includegraphics[width=\fwidth]{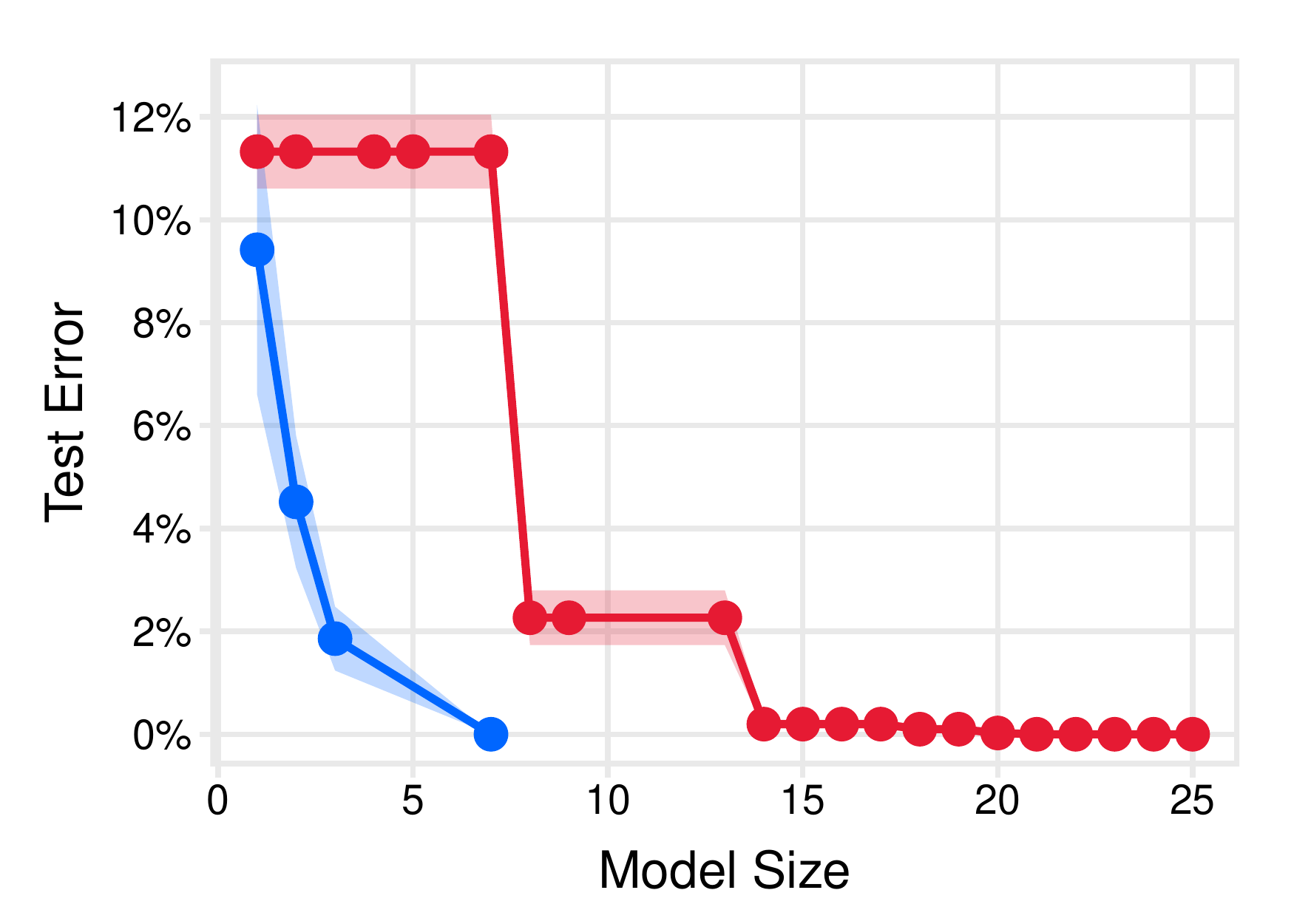} \end{tabular} \\ 
\tabdataname{spambase}  & \begin{tabular}{lr}\includegraphics[width=\fwidth]{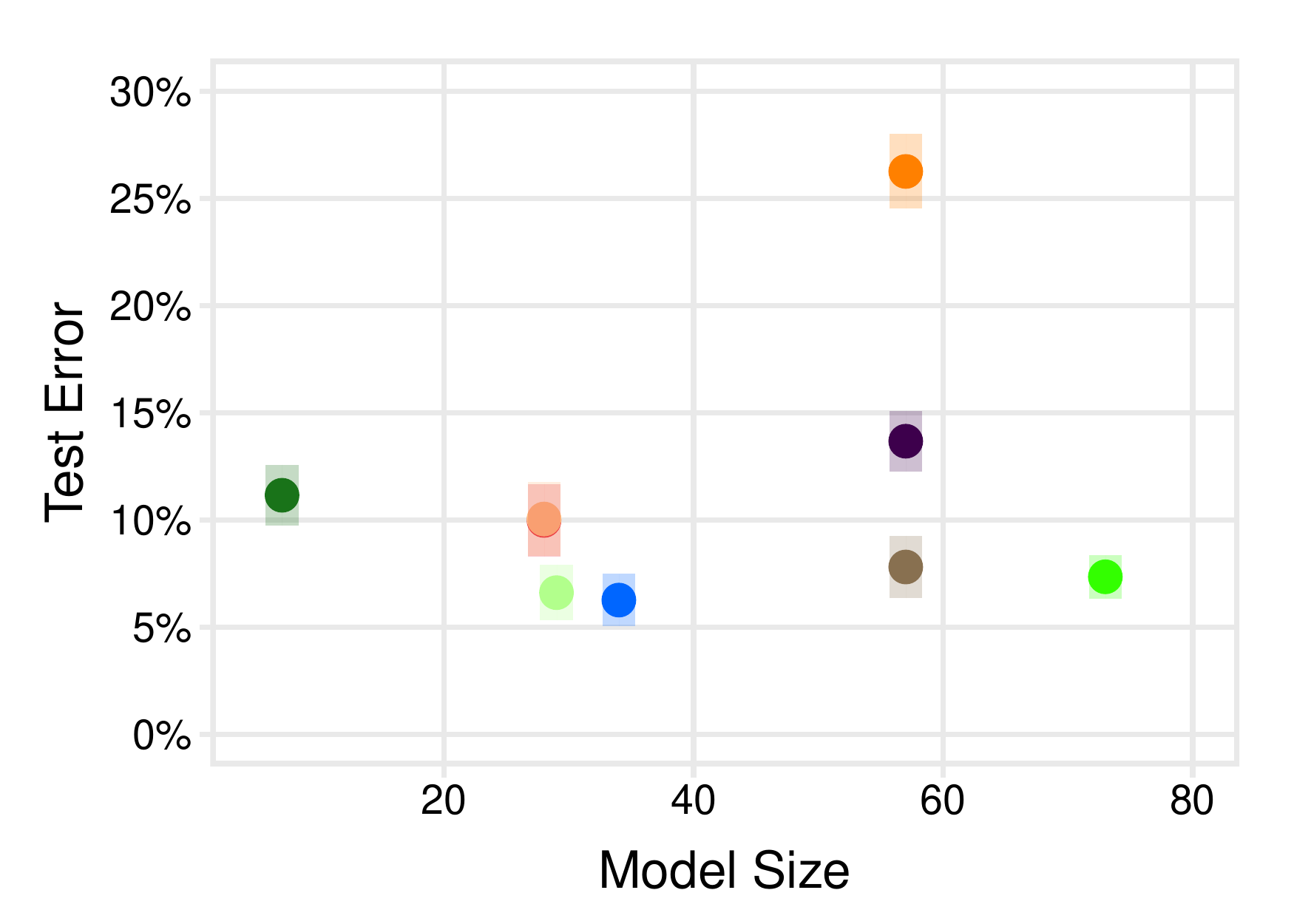}  & \includegraphics[width=\fwidth]{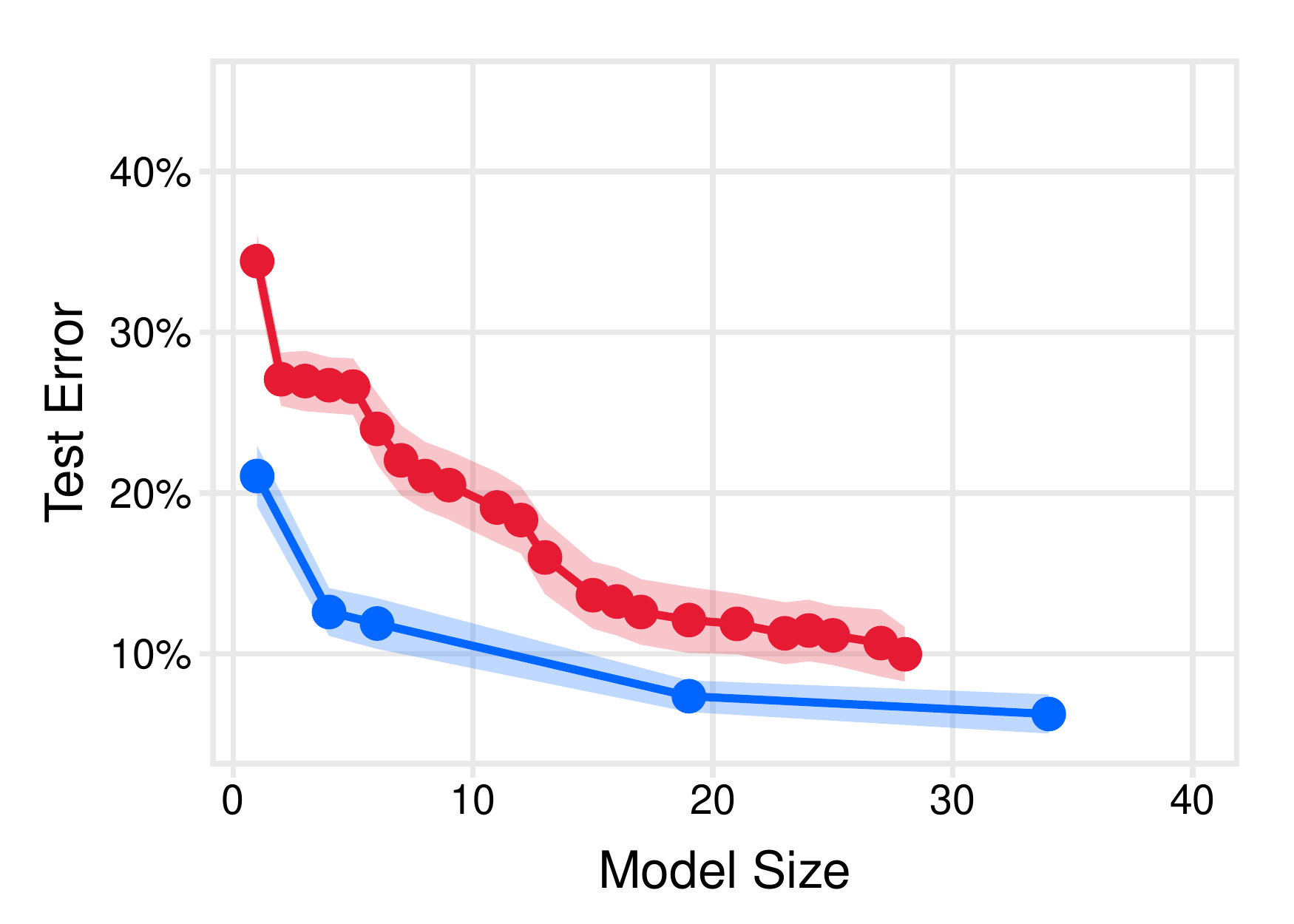} \end{tabular}
\end{tabular}
\caption{Accuracy and sparsity of all classification methods on all datasets. For each dataset, we plot the performance of models when free parameters are set to values that minimize the 10-CV mean test error (left)m and plot the performance of SLIM and Lasso across the full $\lzero$-regularization path (right).}
\label{Fig::ExpPlots2}
\end{figure}
\FloatBarrier
\section{Specialized Models}\label{Sec::Extensions}
In this section, we present three specialized models related to SLIM. These models are all special instances of the optimization problem in \eqref{Eq::InterpretabilityFrameworkLinear}.
\subsection{Personalized Models}\label{Sec::PILM}
A \textit{Personalized Integer Linear Model} (PILM) is a generalization of SLIM that provides soft control over the coefficients in a scoring system. To use this model, users define $R+1$ interpretability sets,
\begin{align*}
\Lset_r = \{ l_{r,1},\ldots,l_{r,K_r}\} \for r = 0,\ldots, R,
\end{align*}
as well as a ``personalized" interpretability penalty,
\begin{align*}
\IntPen_j(\lambda_j) &= 
  \begin{dcases*}
    C_0 & if $\lambda_j \in \Lset_0$ \vspace{-0.2cm}\\
    & $\tiny{\vdots}$ \vspace{-0.1cm} \\
    C_R & if $\lambda_j \in \Lset_R$.\vspace{-0.1cm}
  \end{dcases*}
\end{align*}
In order to penalize coefficients from less interpretable sets more heavily, we need that: (i) $\Lset_1,\ldots,\Lset_R$ are mutually exclusive; (ii) $\Lset_r$ is more interpretable than $\Lset_{r+1}$; (iii) the trade-off parameters are monotonically increasing in $r$, so that $C_0<\ldots<C_R$. The values of the parameters $C_r$ can be set as the minimum gain in training accuracy required for the optimal classifier to use a coefficient from $\Lset_r$. 

As an example, consider training a PILM scoring system with the penalty:
\begin{align*}
\IntPen_j(\lambda_j) &= 
  \begin{cases}
    C_0 = 0.00 & \text{if} \quad \lambda_j \in {0}\\
    C_1 = 0.01 & \text{if} \quad \lambda_j \in \pm \{1,\ldots,10\}\\
    C_2 = 0.05 & \text{if} \quad \lambda_j \in \pm \{11,\ldots,100\}.
  \end{cases}
\end{align*}
Here, the optimal classifier will use a coefficient from $\Lset_1$ if it yields at least a 1\% gain in training accuracy, and a coefficient from $\Lset_2$ if it yields at least a 5\% gain in training accuracy.

We can train a PILM scoring system by solving the following IP:
\begin{subequations}
\begin{equationarray}{crcl>{\hspace{0.1cm}}l>{\quad}r}
\min_{\lambdab,\bf{\psi},\bf{\Phi},\bf{u}} &\frac{1}{N}\sum_{i=1}^{N} \loss_i & + &  \sum_{j=1}^{P} \IntPen_j  \notag \\
\st & M_i\loss_i & \geq & \gamma -\sum_{j=0}^P y_i \lambda_j x_{i,j} & \mprange{i}{1}{N} & \mpdes{0--1 loss} \label{Con::PILMLoss} \\
& \IntPen_j & = & \sum_{r=0}^R \sum_{k=1}^{K_r} C_r u_{j,k,r} & \mprange{j}{1}{P} & \mpdes{int. penalty} \label{Con::PILMIntPenalty} \\
& \lambda_j & = & \sum_{r=0}^R \sum_{k=1}^{K_r} l_{r,k} u_{j,k,r} &\mprange{j}{0}{P} & \mpdes{coefficient values} \label{Con::PILMIntCon} \\
&  1  & = & \sum_{r=0}^R \sum_{k=1}^{K_r} u_{j,k,r} & \mprange{j}{0}{P} & \mpdes{1 int. set per coef.}  \label{Con::PILMOnly1} \\
& \loss_i & \in & \B &  \mprange{i}{1}{N} & \mpdes{loss variables} \notag  \\
& \IntPen_j  & \in & \R_+  & \mprange{j}{1}{P} & \mpdes{int. penalty variables} \notag \\
& u_{j,r,k} & \in & \B &  \mprange{j}{0}{P} \quad \mprange{r}{0}{R} \quad \mprange{k}{1}{K_r} & \mpdes{coef. value variables} \notag 
\end{equationarray}
\end{subequations}
Here, the loss constraints and Big-M parameters in \eqref{Con::PILMLoss} are identical to those from the SLIM IP formulation in Section \ref{Sec::Methodology}. The $u_{j,k,r}$ are binary indicator variables that are set to 1 if $\lambda_j$ is equal to $l_{k,r}$. Constraints \eqref{Con::PILMOnly1} ensure that each coefficient uses exactly one value from one interpretability set. Constraints \eqref{Con::PILMIntCon} ensure that each coefficient $\lambda_j$ is assigned a value from the appropriate interpretability set $\Lset_r$. Constraints \eqref{Con::PILMIntPenalty} ensure that each coefficient $\lambda_j$ is assigned the value specified by the personalized interpretability penalty.
\subsection{Rule-Based Models}
SLIM can be extended to produce specialized ``rule-based" models when the training data are composed of \textit{binary rules}. In general, any real-valued feature can be converted into a binary rule by setting a threshold (e.g., we can convert $age$ into the feature $age \geq 25 := \indic{age \geq 25}$). The values of the thresholds can be set using domain expertise, rule mining, or discretization techniques \citep{Liu:2002ty}.

In what follows, we assume that we train models with a binarized dataset that contains $T_j$ binary rules $\bm{h}_{j,t} \in \B^N$ for each feature $\xb_j \in \R^N$ in the original dataset. Thus, we consider models with the form:
\begin{align*}
\hat{y} = \sign{ \lambda_0 + \sum_{j=1}^P \sum_{t=1}^{T_j} \lambda_{j,t}h_{j,t}}. 
\end{align*}
We make the following assumptions about the binarization process. If $\xb_j$ is a binary variable, then it is left unchanged so that $T_j = 1$ and $\bm{h}_{j,T_j} := \xb_j$. If $\xb_j$ is a categorical variable $\xb_j \in \{1,\ldots,K\}$, the binarization yields a binary rule for each category so that $T_j=K$ and $\bm{h}_{j,t}:=\indic{\xb_j=k}$ for $t = 1,\ldots,K$. If $\xb_j$ is a real variable, then the binarization yields $T_j$ binary rules%
\footnote{While there exists an infinite number of thresholds for a real-valued feature, we only need consider at most $N-1$ thresholds (i.e. one threshold placed each pair of adjacent values, $x_{(i),j}<v_{j,t}<x_{(i+1),j}$). Using additional thresholds will produce the same set of binary rules and the same rule-based model.}
of the form $\bm{h}_{j,t} := \indic{\xb_j \geq v_{j,t}}$ where $v_{j,t}$ denotes the $t^\text{th}$ threshold for feature $j$.
\subsubsection{M-of-N Rule Tables}\label{Sec::MofNRuleTables}
\textit{M-of-N rule tables} are simple rule-based models that, given a set of N rules, predict $\hat{y} = +1$ if at least M of them are true (see e.g., Figure \ref{Fig::MNRuleExample}). These models have the major benefit that they do not require the user to compute a mathematical expression. M-of-N rule tables were originally proposed as auxiliary models that could be extracted from neural nets \citep{Towell:1993tx} but can also be trained as stand-alone discrete linear classification models as suggested by \citet{chevaleyre2013rounding}.

We can produce a fully optimized M-of-N rule table by solving an optimization problem of the form:
\begin{align*}
\hspace{5cm} \min_{\lambdab} & \qquad \sum_{i=1}^N\indic{y_i \hat{y}_i \leq 0}+ C_0 \vnorm{\lambdab}_0 & \\ 
\st & \qquad \lambda_0 \in \{-P,\ldots,0\} & \\ 
& \qquad \lambda_{j,t} \in \B &     \mprange{j}{1}{P} \quad \mprange{t}{1}{T_j}. 
\end{align*}
The coefficients from this optimization problem yield an M-of-N rule table with $M=\lambda_0 + 1$ and $N= \sum_{j=1}^P\sum_{t=1}^{T_j} \lambda_{j,t}$. Here, we can achieve exact $\lzero$-regularization using an $\lone$-penalty since $\vnorm{\lambda_{j,t}}_0 = \vnorm{\lambda_{j,t}}_1$ for $\lambda_{j,t} \in \B$. Since we use the 0--1 loss, the trade-off parameter $C_0$ can be set as minimum gain in training accuracy required to include a rule in the optimal table.
\begin{figure}[htbp]
\centering{
\small{
\begin{tabular}{c}
  \bfcell{c}{PREDICT TUMOR IS BENIGN\\IF AT LEAST 5 OF THE FOLLOWING 8 RULES ARE TRUE}\\ 
  \toprule
  $UniformityOfCellSize\geq3$ \\ 
  $UniformityOfCellShape\geq3$ \\ 
  $MarginalAdhesion\geq3$ \\ 
  $SingleEpithelialCellSize\geq3$ \\ 
  $BareNuclei\geq3$ \\ 
  $NormalNucleoli\geq3$ \\ 
  $BlandChromatin\geq3$ \\ 
  $Mitoses\geq3$
  \end{tabular}
}
}
\caption{M-of-N rule table for the \texttt{breastcancer} dataset for $C_0 = 0.9/NP$. This model has 8 rules and a 10-CV mean test error of $4.8 \pm 2.5\%$. We trained this model with binary rules $h_{i,j} := \indic{x_{i,j} \geq 3}$.}
\label{Fig::MNRuleExample}
\end{figure}

We can train an M-of-N rule table by solving the following IP:
\begin{subequations}
\begin{equationarray}{crcl>{\hspace{0.1cm}}l>{\quad}r}
\min_{\lambdab,\bf{\psi},\bf{\Phi}} &\frac{1}{N}\sum_{i=1}^{N} \loss_i & + & \sum_{j=1}^{P} \IntPen_j  \notag \\
\st          & M_i \loss_i                  & \geq & \gamma -\sum_{j=0}^P\sum_{t=1}^{T_j} y_i \lambda_{j,t} h_{i,j,t}       &\mprange{i}{1}{N} & \mpdes{0--1 loss} \label{Con::MNLoss} \\
& \IntPen_{j,t} & = & C_0 \lambda_{j,t} &\mprange{j}{1}{P} \quad \mprange{t}{1}{T_j} & \mpdes{int. penalty} \label{Con::MNIntPenalty} \\
& \lambda_0 & \in &  \{-P,\ldots,0\} &  & \mpdes{intercept value} \notag \\ 
& \lambda_{j,t} & \in & \{0,1\} & \mprange{j}{1}{P} \quad \mprange{t}{1}{T_j} & \mpdes{coefficient values} \notag \\
& \loss_i & \in & \B &\mprange{i}{1}{N} & \mpdes{0--1 loss indicators} \notag  \\
& \IntPen_{j,t}  & \in & \R_+  & \mprange{j}{1}{P} \quad \mprange{t}{1}{T_j} & \mpdes{int. penalty values} \notag 
\end{equationarray}
\end{subequations}
Here, the loss constraints and Big-M parameters in \eqref{Con::MNLoss} are identical to those from the SLIM IP formulation in Section \ref{Sec::Methodology}. Constraints \eqref{Con::MNIntPenalty} define the penalty variables $\IntPen_{j,t}$ as the value of the $\lzero$-penalty.
\subsubsection{Threshold-Rule Models}\label{Sec::TILM}
A \textit{Threshold-Rule Integer Linear Model} (TILM) is a scoring system where the input variables are thresholded versions of the original feature set (i.e. decision stumps). These models are well-suited to problems where the outcome has a non-linear relationship with real-valued features. As an example, consider the SAPS II scoring system of \citet{le1993new}, which assesses the mortality of patients in intensive care using thresholds on real-valued features such as $blood\_pressure>200$ and $heart\_rate<40$. TILM optimizes the binarization of real-valued features by using feature selection on a large (potentially exhaustive) pool of binary rules for each real-valued feature. \citet{carrizosa2010binarized,van2013risk} and \citet{goh2014box} take different but related approaches for constructing classifiers with binary threshold rules. 

We train TILM scoring systems using an optimization problem of the form:
\begin{align*}
\begin{split}
\min_{\lambdab} & \qquad \frac{1}{N}\sum_{i=1}^N\indic{y_i \hat{y}_i \leq 0} + C_f \cdot \text{Features} + C_t \cdot \text{Rules per Feature} + \epsilon \vnorm{\lambdab}_1 \\ 
\st & \qquad \lambdab \in \Lset, \\
    & \qquad \sum_{t=1}^{T_j} \indic{\lambda_{j,t}\neq 0} \leq R_{max} \text{ for } j = 1,\ldots,P,  \\ 
    & \qquad \sign{\lambda_{j,1}} = \ldots = \sign{\lambda_{j,T_j}} \text{ for } j = 1,\ldots,P.
\end{split}
\end{align*}
TILM uses an interpretability penalty that penalizes the number of rules used in the classifier as well as the number of features associated with these rules. The small $\lone$-penalty in the objective restricts coefficients to coprime values as in SLIM. Here, $C_f$ tunes the number of features used in the model, $C_t$ tunes the number of rules per feature, and $\epsilon$ is set to a small value to produce coprime coefficients. TILM includes additional hard constraints to limit the number of rules per feature to $R_{max}$ (e.g., $R_{max} = 3$), and to ensure that the coefficients for binary rules from a single feature agree in sign (this ensures that each feature maintains a strictly monotonically increasing or decreasing relationship with the outcome).

We can train a TILM scoring system by solving the following IP:
\begin{subequations}
\begin{equationarray}{crcl>{\hspace{0.1cm}}lr}
\min_{\lambdab,\bf{\psi},\bf{\Phi},\bf{\tau},\bf{\nu},\bf{\delta}} &\frac{1}{N}\sum_{i=1}^{N} \loss_i & + & \sum_{j=1}^{P} \IntPen_j  \notag \\
\st          & M_i \loss_i                  & \geq & \gamma -\sum_{j=0}^P\sum_{t=1}^{T_j} y_i \lambda_{j,t} h_{i,j,t} &\mprange{i}{1}{N} & \mpdes{0--1 loss} \label{Con::TILMLoss} \\
& \IntPen_j & = & C_f\nu_j + C_t \tau_j + \epsilon \sum_{t=1}^{T_j}{\beta_{j,t}} & \mprange{j}{1}{P} & \mpdes{int. penalty} \label{Con::TILMIntPenalty} \\
& T_j \nu_j           & = & \sum_{t=1}^{T_j} \alpha_{j,t}      &\mprange{j}{1}{P} & \mpdes{feature use} \label{Con::TILMFeatures}\\ 
& \tau_j              & = & \sum_{t=1}^{T_j} \alpha_{j,t}-1    &\mprange{j}{1}{P} & \mpdes{threshold/feature} \label{Con::TILMThresholdPerFeature} \\
& \tau_j              & \leq & R_{max} + 1  &\mprange{j}{1}{P} & \mpdes{max thresholds} \label{Con::TILMMaxThresholds} \\
&-\Lambda_j\alpha_{j,t}    & \leq & \lambda_{j,t} \leq \Lambda_j\alpha_{j,t}  &\mprange{j}{1}{P} \quad \mprange{t}{1}{T_j} & \mpdes{$\lzero$ norm} \notag \\ 
&-\beta_{j,t}              & \leq & \lambda_{j,t}  \leq \beta_{j,t}  &\mprange{j}{1}{P} \quad \mprange{t}{1}{T_j} & \mpdes{$\lone$ norm} \notag \\ 
&-\Lambda_j (1-\delta_{j})    & \leq & \lambda_{j,t} \leq \Lambda_j\delta_{j}  &\mprange{j}{1}{P} \quad \mprange{t}{1}{T_j} & \mpdes{agree in sign} \label{Con::TILMSigns} \\
& \lambda_{j,t} & \in & \Lset_j &  \mprange{j}{0}{P} \quad \mprange{t}{1}{T_j} & \mpdes{coefficient values} \notag \\ 
& \loss_i & \in & \B &  \mprange{i}{1}{N} & \mpdes{0--1 loss indicators} \notag  \\
& \IntPen_j  & \in & \R_+  & \mprange{j}{1}{P} & \mpdes{int. penalty variables} \notag \\
& \alpha_j  & \in & \B  & \mprange{j}{1}{P} & \mpdes{$\lzero$ variables} \notag \\
& \beta_j    & \in & \R_+ & \mprange{j}{1}{P} & \mpdes{$\lone$ variables} \notag \\
& \nu_j    & \in & \B & \mprange{j}{1}{P} & \mpdes{feature use indicators} \notag \\ 
& \tau_j    & \in & \Z_+ & \mprange{j}{1}{P} & \mpdes{threshold/feature variables} \notag \\ 
& \delta_j    & \in & \B & \mprange{j}{1}{P} & \mpdes{sign indicators} \notag
\end{equationarray}
\end{subequations}
Here, the loss constraints and Big-M parameters in \eqref{Con::TILMLoss} are identical to those from the SLIM IP formulation in Section \ref{Sec::Methodology}. Constraints \eqref{Con::TILMIntPenalty} set the interpretability penalty for each coefficient as $\IntPen_j = C_f\nu_j + C_t \tau_j + \epsilon \sum\beta_{j,t}$. The variables in the interpretability penalty include: $\nu_j$, which indicate that we use at least one threshold rule from feature $j$; $\tau_j$, which count the number of additional binary rules derived from feature $j$; and $\beta_{j,t}:= |\lambda_{j,t}|$. The values of $\nu_j$ and $\tau_j$ are set using the indicator variables $\alpha_{j,t} := \indic{\lambda_{j,t}\neq0}$ in constraints \eqref{Con::TILMFeatures} and \eqref{Con::TILMThresholdPerFeature}. Constraints \eqref{Con::TILMMaxThresholds} limit the number of binary rules from feature $j$ to $\R_{max}$. Constraints \eqref{Con::TILMSigns} ensure that the coefficients of binary rules derived from feature $j$ agree in sign; these constraints are encoded using the variables $\delta_j:=\indic{\lambda_{j,t} \geq 0}$. 
\clearpage\section{Conclusion}\label{Sec::Conclusion}
In this paper, we introduced a new method for creating data-driven medical scoring systems which we refer to as a Supersparse Linear Integer Model (SLIM). We showed how SLIM can produce scoring systems that are fully optimized for accuracy and sparsity, that can accomodate multiple operational constraints, and that can be trained without parameter tuning. 

The major benefits of our approach over existing methods come from the fact that we avoid approximations that are designed to achieve faster computation. Approximations such as surrogate loss functions and $\lone$-regularization hinder the accuracy and sparsity of models as well as the ability of practitioners to control these qualities. Such approximations are no longer needed for many datasets, since using current integer programming software, we can now train scoring systems for many real-world problems. Integer programming software also caters to practitioners in other ways, by allowing them to choose from a pool of models by mining feasible solutions and to seamlessly benefit from periodic computational improvements without revising their code.
\clearpage
\section*{Acknowledgments}
We thank the editors and reviewers for valuable comments that helped improve this paper.
In addition, we thank Dr. Matt Bianchi and Dr. Brandon Westover at the Massachusetts General Hospital Sleep Clinic for providing us with data used in Section 5.
We gratefully acknowledge support from Siemens and Wistron.
\bibliographystyle{plainnat}
\bibliography{SLIMforOptimizedMedicalScoringSystems}
\clearpage
\appendix
\section{Proofs of Theorems}\label{Appendix::Proofs}
%
\subsection*{Proof of Theorem \ref{Thm::MinMarginBound} (Minimum Margin Resolution Bound)}
\proof
%
%
%

We use normalized versions of the vectors, $\rhob/\vnorm{\rhob}_2$ and $\lambdab/\Lambda$ because the 0--1 loss is scale invariant:
\begin{align*}
 \sum_{i=1}^N \indic{y_i\lambdab^T\xb_i\leq 0} &= \sum_{i=1}^N \indic{y_i\frac{\lambdab^T\xb_i}{\Lambda}\leq 0}, \\
 \sum_{i=1}^N \indic{y_i \rhob^T \xb_i \leq 0} &=  \sum_{i=1}^N \indic{y_i\frac{\rhob^T\xb_i}{\vnorm{\rhob}_2}\leq 0}.
\end{align*}

We set $\Lambda > \frac{X_{\max}\sqrt{P}}{2 \gamma_{\min}}$ as in \eqref{Eq::MinMarginLambda}. Using $\Lambda$, we then define $\lambdab/\Lambda$ element-wise so that $\lambda_j/\Lambda$ is equal to $\rho_j/\vnorm{\rhob}_2$ rounded to the nearest $1/\Lambda$ for $j= 1,\ldots,P$.

We first show that our choice of $\Lambda$ and $\lambdab$ ensures that the difference between the margin of $\rhob/\vnorm{\rhob}_2$ and the margin of $\lambdab/\Lambda$ on all training examples is always less than the minimum margin of $\rhob/\vnorm{\rhob}_2$, defined as $\gamma_{\min} = \min_i \frac{|\rhob^T\xb_i|}{\vnorm{\rhob}_2}$. This statement follows from the fact that, for all $i$:
\begin{align}
\left| \frac{\lambdab^T \xb_i}{\Lambda}-\frac{\rhob^T\xb_i}{\vnorm{\rhob}_2} \right| ~ \leq ~ & \left\|\frac{\lambdab}{\Lambda} - \frac{\rhob}{\vnorm{\rhob}_2}\right\|_2\|\xb_i\|_2 \label{ThmPf::A} \\
= ~         & \left( \sum_{j=1}^P  \left| \frac{\lambda_j}{\Lambda} - \frac{\rho_j}{\vnorm{\rhob}_2} \right|^2 \right)^{1/2}\|\xb_i\|_2 \notag \\ 
\leq ~      & \left(\sum_{j=1}^P \frac{1}{(2\Lambda)^2} \right)^{1/2}\|\xb_i\|_2 \label{ThmPf::B} \\
= ~         & \frac{\sqrt{P}}{2\Lambda}X_{\max} \notag \\ 
< ~         & \frac{\sqrt{P}X_{\max}}{2\left( \frac{X_{\max}\sqrt{P}}{2\min_i  \frac{|\rhob^T\xb_i|}{\vnorm{\rhob}_2}}         \right) }  \label{ThmPf::C} \\ 
= ~ & \min_i \frac{|\rhob^T\xb_i|}{\vnorm{\rhob}_2}.  \label{ThmPf::D}
\end{align}
Here: the inequality in \eqref{ThmPf::A} uses the Cauchy-Schwarz inequality; the inequality in \eqref{ThmPf::B} is due to the fact that the distance between $\rho_j/\vnorm{\rhob}_2$ and $\lambda_j/\Lambda$ is at most $1/2\Lambda$; and the inequality in \eqref{ThmPf::C} is due to our choice of $\Lambda$.

Next, we show that our choice of $\Lambda$ and $\lambdab$ ensures that $\rhob/\vnorm{\rhob}_2$ and $\lambdab/\Lambda$ classify each point in the same way. We consider three cases: first, the case where $\xb_i$ lies on the margin; second, the case where $\rhob$ has a positive margin on $\xb_i$; and third, the case where $\rhob$ has a negative margin on $\xb_i$. 
For the case when $\xb_i$ lies on the margin, $\min_i|\rhob^T\xb_i|=0$ and the theorem holds trivially.
For the case where $\rhob$ has positive margin, $\rhob^T\xb_i>0$, the following calculation using \eqref{ThmPf::D} is relevant:
\begin{eqnarray*}
\frac{\rhob^T\xb_i}{\vnorm{\rhob}_2} - \frac{\lambdab^T\xb_i}{\Lambda}  \leq \left| \frac{\lambdab^T\xb_i}{\Lambda}-\frac{\rhob^T\xb_i}{\vnorm{\rhob}_2}   \right| 
< \min_i \frac{|\rhob^T\xb_i|}{\vnorm{\rhob}_2}.
\end{eqnarray*}
We will use the fact that for any $i^{'}$, by definition of the minimum: 
\begin{align*}
0 \leq \frac{|\rhob^T\xb_{i^{'}}|}{\vnorm{\rhob}_2}- \min_i \frac{|\rhob^T\xb_i |}{\vnorm{\rhob}_2},
\end{align*}
and combine this with a rearrangement of the previous expression to obtain:
\begin{eqnarray*}
0\leq \frac{|\rhob^T\xb_i|}{\vnorm{\rhob}_2}  -  \min_i \frac{|\rhob^T\xb_i |}{\vnorm{\rhob}_2} = \frac{\rhob^T\xb_i}{\vnorm{\rhob}_2}  -  \min_i \frac{|\rhob^T\xb_i |}{\vnorm{\rhob}_2} < \frac{\lambdab^T\xb_i}{\Lambda}.
\end{eqnarray*}
Thus, we have shown that $\lambdab^T\xb_i>0$ whenever $\rhob^T\xb_i>0$.

For the case where $\rhob$ has a negative margin on $\xb_i$, $\rhob^T\xb_i<0$, we perform an analogous calculation:
\begin{eqnarray*}
\frac{\lambdab^T\xb_i}{\|\lambdab\|_2} - \frac{\rhob^T\xb_i}{\vnorm{\rhob}_2}  \leq \left| \frac{\lambdab^T\xb_i}{\Lambda}-\frac{\rhob^T\xb_i}{\vnorm{\rhob}_2}   \right| 
< \min_i \frac{|\rhob^T\xb_i|}{\vnorm{\rhob}_2}.
\end{eqnarray*}
and then using that $\rhob^T\xb_i<0$,
\begin{eqnarray*}
0\leq \frac{|\rhob^T\xb_i|}{\vnorm{\rhob}_2}  -  \min_i \frac{|\rhob^T\xb_i |}{\vnorm{\rhob}_2} = \frac{-\rhob^T\xb_i}{\vnorm{\rhob}_2}  -  \min_i \frac{|\rhob^T\xb_i |}{\vnorm{\rhob}_2} < -\frac{\lambdab^T\xb_i}{\Lambda}.
\end{eqnarray*}
Thus, we have shown $\lambdab^T\xb_i<0$ whenever $\rhob^T\xb_i<0$.

Putting both the positive margin and negative margin cases together, we find that for all $i$,
\[\indic{y_i\rhob^T\xb_i\leq 0} = \indic{y_i\lambdab^T\xb_i\leq 0}.\] 
Summing over $i$ yields the statement of the theorem. 
\qed 
\endproof

\subsection*{Proof of Theorem \ref{Thm::L0Bound} (Generalization of Sparse Discrete Linear Classifiers)}
\proof
Let $Z(\lambdab;\data_N) = \frac{1}{N} \sum_{i=1}^N \indic{y_i \lambdab^T \xb_i \leq 0} + C_0\vnorm{\lambdab}_0.$ Note that $\lambdab = 0$  is a feasible solution since we assume that $0 \in \Lset$. Since $\lambdab = 0$ achieves an objective value of $Z(0;\data_N) = 1$, any optimal solution, $\lambdab \in \argmin_{\lambda \in \Lset} Z(\lambdab;\data_N)$, must attain an objective value $Z(\lambdab;\data_N) \leq 1$. This implies
\begin{eqnarray*}
Z(\lambdab;\data_N) \leq 1,\\
 C_0\vnorm{\lambdab}_0 \leq \frac{1}{N}\sum_{i=1}^N \indic{y_i \lambdab^T \xb_i \leq 0}  + C_0\vnorm{\lambdab}_0  \leq 1,\\ 
\vnorm{\lambdab}_0 \leq \frac{1}{C_0},\\
\vnorm{\lambdab}_0 \leq \left\lfloor \frac{1}{C_0} \right\rfloor. 
\end{eqnarray*}
The last line uses that $\|\lambdab\|_0$ is an integer. 

Thus, $\mathcal{H}_{P,C_0}$ is large enough to contain all minimizers of $Z(\cdot;\data_N)$ for any $\data_N$. The statement of the theorem follows from applying Theorem \ref{Thm::NormalBound}.
\qed
\endproof

\subsection*{Proof of Theorem \ref{Thm::DataReduction} (Equivalence of the Reduced Data)}
\proof
Let us denote the set of classifiers whose objective value is less or equal to $\tilde{Z}(\tilde{f}^*;\data_N)$ as
\begin{align*}
\tilde{\F}^{\varepsilon} = \left \{ f \in \tilde{\F} \; \Big | \; \tilde{Z}(f;\data_N) \leq \tilde{Z}(\tilde{f};\data_N) + \varepsilon \right\}.
\end{align*}
In addition, let us denote the set of points that have been removed by the data reduction algorithm
\begin{align*}
\mathcal{S} = \data_N \setminus \data_M.
\end{align*}
By definition, data reduction only removes an example if its sign is fixed. This means that $\sign{f(\xb_i)}=\sign{\tilde{f}(\xb_i)}$ for all $i \in \mathcal{S}$ and $f\in\tilde{\F}^{\varepsilon}.$ Thus, we can see that for all classifiers $f \in \tilde{\F}^{\varepsilon}$, 
\begin{align}
\label{Eq:ReductionConstant}
Z(f;\data_N) = Z(f;\data_M)+\sum_{i\in\mathcal{S}} \indic{y_if(\xb_i)\leq 0}= Z(f;\data_M)+\sum_{i\in\mathcal{S}} \indic{y_i\tilde{f}(\xb_i)\leq 0} =Z(f;\data_M)+C.
\end{align}

We now proceed to prove the statement in \eqref{Eq::ReductionWTS}. When $\mathcal{S}=\emptyset$, then $\data_N = \data_M$, and \eqref{Eq::ReductionWTS} follows trivially. When, $\mathcal{S}\neq\emptyset$, we note that
\begin{align}
\F^*=\argmin_{f\in\F} Z(f;\data_N) &= \argmin_{f\in\F} Z(f;\data_M \cup \mathcal{S}), \nonumber \\
&= \argmin_{f\in\F} Z(f;\data_M) + Z(f; \mathcal{S}), \nonumber \\
&= \argmin_{f\in\F} Z(f;\data_M) + C, \label{Eq:ReductionFinal} \\ 
&= \argmin_{f\in\F} Z(f;\data_M). \nonumber
\end{align}
Here, the statement in \eqref{Eq:ReductionFinal} follows directly from \eqref{Eq:ReductionConstant}.
\qed 
\endproof
\subsection*{Proof of Theorem \ref{Thm::ReductionSufficientConditions} (Sufficient Conditions to Satisfy the Level Set Condition)}
\proof
We assume that we have found a surrogate function, $\psi$, that satisfies conditions I--IV and choose $C_{\psi} > 2\varepsilon$. 

Our proof uses the following result: if 
$\| \lmip - \lcvx \| > C_{\lambdab}$
then $\lmip$ cannot be a minimizer of $\Zmip{\lambdab}$ because 
this would lead to a contradiction with the definition of $\lmip$. 
To see that this result holds, we use condition III with $\lambdab=\lmip$ to see that $\|\lmip - \lcvx\|>C_{\lambdab}$ implies $\Zcvx{\lmip} - \Zcvx{\lcvx} >C_{\psi}$. Thus,
\begin{align} 
\Zcvx{\lcvx} + C_{\psi} &< \Zcvx{\lmip} \notag \\ 
\Zcvx{\lcvx} + C_{\psi} &< \Zmip{\lmip} + \varepsilon \label{lemmaeq1} \\ 
\Zcvx{\lcvx} + C_{\psi} - \varepsilon &< \Zmip{\lmip} \notag \\ 
\Zcvx{\lcvx} + C_{\psi} - \varepsilon &< \Zcvx{\lmip} \label{lemmaeq2} \\
\Zcvx{\lcvx} + \varepsilon &< \Zcvx{\lmip}.\label{lemmaeq3}
\end{align}
Here the inequality in \eqref{lemmaeq1} follows from condition IV, the inequality in \eqref{lemmaeq2} follows from condition I, and the inequality in \eqref{lemmaeq3} follows from our choice that $C_{\psi} > 2\varepsilon$.

We proceed by looking at the LHS and RHS of \eqref{lemmaeq3} separately. Using condition I on the LHS of \eqref{lemmaeq3} we get that:
\begin{align}
\Zmip{\lcvx} + \varepsilon \leq \Zcvx{\lcvx} + \varepsilon. \label{lemmaeq4}
\end{align}
Using condition IV on the RHS of \eqref{lemmaeq3} we get that:
\begin{align}
\Zcvx{\lmip} \leq \Zmip{\lmip} + \varepsilon. \label{lemmaeq5}
\end{align}
Combining the inequalities in \eqref{lemmaeq3}, \eqref{lemmaeq4} and  \eqref{lemmaeq5}, we get that:
\begin{align}
\Zmip{\lcvx} < \Zmip{\lmip}. \label{lemmaeq6}
\end{align}
The statement in \eqref{lemmaeq6} is a contradiction of the definition of $\lmip$. Thus, we know that our assumption was incorrect and thus $\| \lmip - \lcvx \| \leq C_{\lambdab}$. 
We plug this into the Lipschitz condition II as follows:
\begin{align} 
\Zcvx{\lmip} - \Zcvx{\lcvx} & \leq L \| \lmip - \lcvx \| < L C_{\lambdab},  \notag\\ 
\Zcvx{\lmip} & < L C_{\lambdab} + \Zcvx{\lcvx}. \notag
\end{align}
Thus, we have satisfied the level set condition with $\varepsilon = LC_{\lambdab}$.
\qed
\endproof

\end{document}